\documentclass{article}
\pdfoutput=1

\usepackage{ijcai23}

\usepackage[colorlinks,hyperindex,breaklinks]{hyperref}
\usepackage{times}
\usepackage{float}
\usepackage{soul}
\usepackage{url}
\usepackage[utf8]{inputenc}
\usepackage[small]{caption}
\usepackage{graphicx}
\usepackage{amsmath}
\usepackage{amsthm}
\usepackage{booktabs}
\usepackage{algorithm}
\usepackage{algorithmic}
\usepackage[switch]{lineno}
\usepackage{algorithm}
\usepackage{subfigure}
\usepackage{amsfonts}
\usepackage{amsmath}
\usepackage{verbatim}
\usepackage{amssymb}
\usepackage{newfloat}
\usepackage{listings}
\usepackage{xcolor}
\usepackage {multirow}
\usepackage{appendix}

\title{ALL-E: Aesthetics-guided Low-light Image Enhancement}

\author{
    Author Name
    \affiliations
    Affiliation
    \emails
    email@example.com
}

\author{
Ling Li\and
Dong Liang\and
Yuanhang Gao\and
Sheng-Jun Huang\and
Songcan Chen
\affiliations
Nanjing University of Aeronautics and Astronautics\\
MIIT Key Laboratory of Pattern Analysis and Machine Intelligence\\
Collaborative Innovation Center of Novel Software Technology and Industrialization
\emails
\{liling, liangdong, gaoyuanhang, huangsj, s.chen\}@nuaa.edu.cn\\
\vspace{.2cm}
{\color{magenta}\url{https://dongl-group.github.io/project_pages/ALLE.html}}
}

\begin{document}

\maketitle

\begin{abstract}
    Evaluating the performance of low-light image enhancement (LLE) is highly subjective, thus making integrating human preferences into image enhancement a necessity. Existing methods fail to consider this and present a series of potentially valid heuristic criteria for training enhancement models. In this paper, we propose a new paradigm, \emph{i.e.}, aesthetics-guided low-light image enhancement (ALL-E), which introduces aesthetic preferences to LLE and motivates training in a reinforcement learning framework with an aesthetic reward. Each pixel, functioning as an agent, refines itself by recursive actions, \emph{i.e.}, its corresponding adjustment curve is estimated sequentially. Extensive experiments show that integrating aesthetic assessment improves both subjective experience and objective evaluation. Our results on various benchmarks demonstrate the superiority of ALL-E over state-of-the-art methods. 
\end{abstract}

\section{Introduction}
\begin{figure}[t]
\centering
\subfigure[Normal images with an average Aesthetic Score 7.081]{
\includegraphics[width=3.72cm]{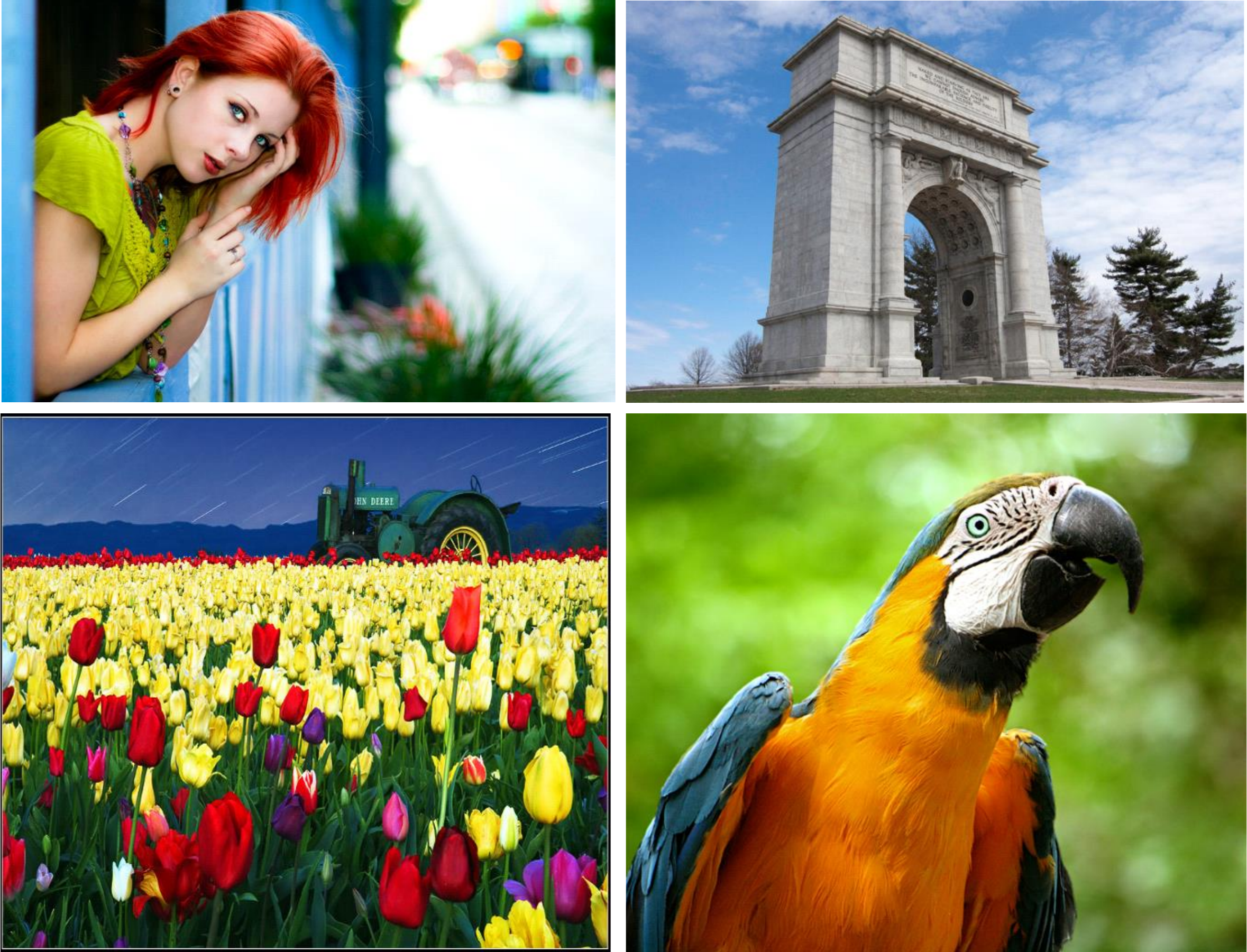}
}
%\hspace{-0.3cm}
\centering
\subfigure[Low-light images with an average Aesthetic Score 5.694]{
\includegraphics[width=3.75cm]{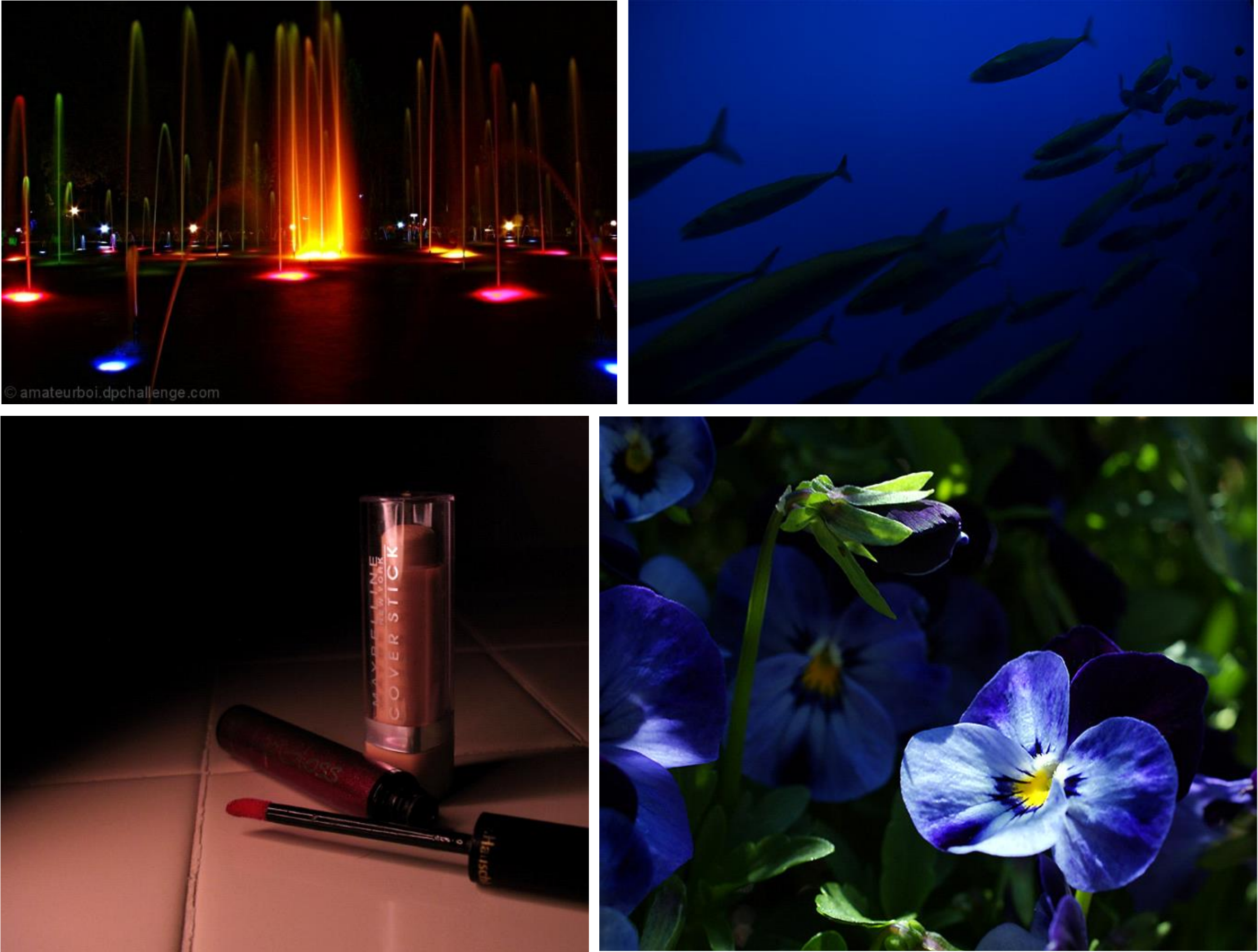}
}
\vspace{-0.4cm}
\caption{Sample images on DPChallenge protect\protect\footnotemark[1]. Each image receives dozens to hundreds of user ratings, ranging from 1 to 10. Higher scores indicate higher aesthetic quality.}
\label{ava}
\end{figure}

Due to the limitations inherent in optical devices and the variability of external imaging conditions, images are often captured with poor lighting, under-saturation, and a narrow dynamic range. These degradation factors significantly impair the visual aesthetics of the images, and lead to detrimental effects on a wide range of downstream computer vision and multimedia tasks \cite{Yu_2021_ICCV,cho2020,Guo_2020_CVPR}.
Manually editing and enhancing low-light images is time-consuming, even for a professional. Therefore, many learning-based strategies have been proposed, introducing a series of potentially valid heuristic constraints for training image enhancement models \cite{lore2017llnet,Chen2018Retinex,zhang2019kindling,Xu_2020_CVPR,8692732,liang2022semantically}. 
However, the aforementioned methods fail to account for the importance of subjective human evaluation in low-light image enhancement (LLE) tasks. On an online professional photography contest -- DPChallenge\footnotemark[1], the subjective human aesthetic scores of normal images are much higher than those of low-light images, and the former provides a significantly superior visual experience compared to the latter, as shown in Fig.~\ref{ava}. 
\footnotetext[1]{https://www.dpchallenge.com/}
In light of this observation, we propose a novel LLE paradigm incorporating aesthetic assessment to effectively model human subjective preferences and improve both subjective experience and objective evaluation of the enhanced images.

However, embedding aesthetic assessment into LLE is non-trivial, as aesthetics are highly subjective and personalized evaluations that vary between individuals. These personalized user preferences can bring confusion to model learning. In addition, the manual aesthetic retouching of photographs is a highly causal and progressive process, which is difficult to replicate using existing LLE methods. To tackle these challenges, we propose the following solutions.
First, we introduce a well-trained `aesthetic oracle network' to construct an Aesthetic Assessment Module that can produce general aesthetic preferences, reducing penalization and increasing the method's versatility and aesthetic appeal. Second, we apply reinforcement learning to interact with the environment (\emph{i.e.}, the Aesthetic Assessment Module) to calculate rewards; we treat LLE as a Markov decision process, decomposing the augmented mapping relationship into a series of iterations through an Aesthetic Policy Generation Module, thus realizing progressive LLE adjustment. Third, we develop a group of complementary rewards, including aesthetics quality, feature preservation, and exposure control, to preserve better subjective visual experience and objective evaluation. All the rewards are non-reference, indicating that they do not require paired training images.

The main contributions of this paper are three-fold:

\begin{itemize}
    \item [$\bullet$] We propose a new paradigm for LLE by integrating aesthetic assessment, leveraging aesthetic scores to mimic human subjective evaluations as a reward to guide LLE.
    To our knowledge, this is the first attempt to solve low-light image enhancement using aesthetics.
    \item [$\bullet$] We devise aesthetics-guided LLE (ALL-E) through reinforcement learning and treat LLE as a Markov decision process, which divides the LLE process into two phases: aesthetic policy generation and aesthetic assessment.
    \item [$\bullet$] Our method is evaluated against state-of-the-art competitors through comprehensive experiments in terms of visual quality, no- and full-referenced image quality assessment, and human subjective surveys. All results consistently demonstrate the superiority of ALL-E.
\end{itemize}

\section{Related Work}
\subsubsection{Low-Light Image Enhancement}

\textbf{Traditional Approaches.} Early efforts commonly presented heuristic priors with empirical observations to address the LLE problems \cite{Pizer1990ContrastlimitedAH,land1977retinex,xu2014novel,guo2016lime}. 
Histogram equalization \cite{Pizer1990ContrastlimitedAH} used a cumulative distribution function to regularize the pixel values to achieve a uniform distribution of overall intensity levels in the image. The Retinex model \cite{land1977retinex} and its multi-scale version \cite{Jobson1997A} decomposed the brightness into illumination and reflections, which were then processed separately. Guo~\emph{et al.} 
 \cite{guo2016lime} introduced a structural prior before refining the initially obtained illumination map and synthesizing the enhanced image according to the Retinex theory. However, these constraints/priors must be more self-adaptive to recover image details and color, avoiding the washing out details, local under/over-saturation, uneven exposure, or halo artifacts.
 \\
\textbf{Deep Learning Approaches.} Lore~\emph{et al.} 
 \cite{lore2017llnet} proposed a variant of the stacked sparse denoising autoencoder to enhance the degraded images. RetinexNet \cite{Chen2018Retinex,zhang2019kindling} leveraged a deep architecture based on Retinex to enhance low-light images. 
%Zhang~$et \ al.$~\cite{} developed three subnetworks for layer decomposition, reflectance restoration, and illumination adjustment based on Retinex. 
RUAS \cite{liu2021retinex} constructed the overall LLE network architecture by unfolding its optimization process. EnlightenGAN \cite{jiang2021enlightengan} introduced GAN-based unsupervised training on unpaired normal-light images for LLE. Zero-DCE \cite{Guo_2020_CVPR} transformed the LLE task into an image-specific curve estimation problem. SCL-LLE \cite{liang2022semantically} cast the image enhancement task as multi-task contrast learning with unpaired positive and negative images and enabled interaction with scene semantics. 
However, all the above methods ignore the human subjective preferences of LLE.
\subsubsection{Aesthetic Quality Assessment}

Although there is no aesthetics-guided LLE method, and the focus of LLE is not on assessing the {aesthetic quality} of a given image, our work relates to this research domain in the sense that LLE aims at improving image quality.

Image aesthetics has become a widely researched topic in current computer vision. Image aesthetic quality assessment aims to simulate human cognition and perception of beauty to automatically predict how beautiful an image looks to a human observer. Previous attempts have trained convolutional neural networks for binary classification of image quality \cite{lu2014rapid,2017Lamp} or aesthetic score regression \cite{kong2016photo}. Assessing visual aesthetics has practical applications in areas such as image retrieval \cite{yu2004exploratory}, image recommendation \cite{yu2021novel}, and color correction \cite{deng2018aesthetic}. 

Since the highly differentiated aesthetic preference, image aesthetics assessment can be divided into two categories: generic and personalized image aesthetics assessment (GIAA and PIAA)\cite{PAM}. \cite{PAM} proposed the first PIAA database named FLICKR-AES. They addressed this problem by leveraging GIAA knowledge on user-related data so that that model can capture aesthetic "offset". Later, research work attempted to learn PIAA from various perspectives, such as multi-modal collaborative learning \cite{CFAN}, meta-learning \cite{BLG-PIAA}, multi-task learning \cite{PA_IAA} etc. Due to the high subjectivity of PIAA tasks, which pay more attention to the differences in personal emotions and personality factors, we aim to introduce "general aesthetic preferences"/generic image aesthetics assessment (GIAA) to LLE.
To our knowledge, the proposed scheme in this paper is the first attempt to solve the LLE problem using aesthetics to preserve better both subjective visual experience and objective evaluation.

\subsubsection{Reinforcement Learning for Image Processing}
After deep Q-network achieved human-level performance on Atari games, there has been a surge of interest in deep reinforcement learning (DRL).
For image-processing tasks, Yu~\emph{et al.} \cite{yu2018crafting} proposed RL-Restore to learn a policy of selecting appropriate tools from a predefined toolbox to restore the quality of corrupted images gradually. Park~\emph{et al.} \cite{park2018distort} presented a DRL-based method for color enhancement and a distortion-recovery training scheme that only requires high-quality reference images for training. 

While these methods focus on global image restoration, Furuta~\emph{et al.} \cite{furuta2019fully} proposed pixelRL to enable pixel-wise image restoration, which extended DRL to pixel-level reinforcement learning, making it more flexible in dealing with image problems. Similarly, Zhang~\emph{et al.} \cite{zhang2021rellie} proposed a novel DRL-based method for achieving LLE at the pixel level. In contrast, our DRL network learns with an image aesthetic reward to obtain LLE results that try to satisfy universal users.

\begin{figure*}[!htb]
\centering
\includegraphics[width=17cm]{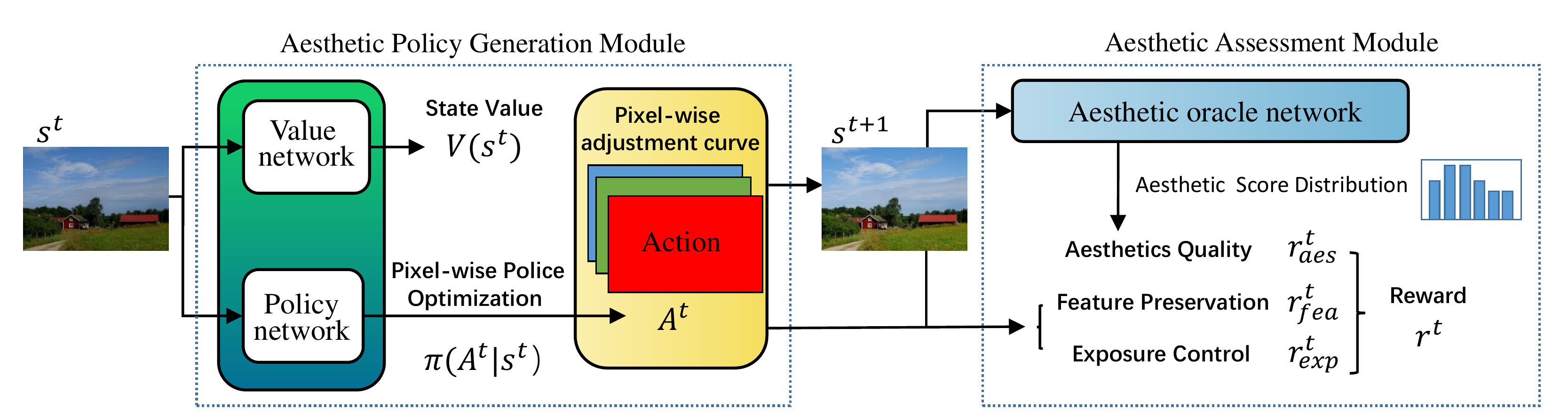}
\caption{Overall architecture of the proposed ALL-E. It includes an aesthetic policy generation module and an aesthetic assessment module.
}
\label{Net}
\vspace{-0.3cm}
\end{figure*}

\section{Methodology}

\subsection{From Aesthetic Annotation to LLE}
Many factors can affect the beauty of images, including the richness of color, correct exposure, depth of view, resolution, high-level semantics, and so on. Previous work  \cite{lu2014rapid,2017Lamp,kong2016photo,yu2004exploratory} have preliminarily analyzed these factors to generate the aesthetic rating of an image, facilitating manual aesthetic annotation for downstream tasks. 

As aesthetics is a highly subjective and personalized evaluation with individual differences, \cite{kang2020eva} modeled an aesthetic mean opinion score (AMOS) as a weighted sum of its four relatively independent attributes to generate more reliable aesthetic annotation,
% \begin{equation}
% \begin{aligned}
%     S_I = \sum_{j = 1}^{4} a^j \cdot g_{I}^{j} 
% \end{aligned}
% \end{equation}
\begin{equation}
\begin{aligned}
  AMOS &\cong 0.288 f_{1} + 0.288 f_{2}+ 0.082  f_{3} + 0.342  f_{4}
\end{aligned}
\label{amos}
\end{equation}
where $f$ is the scores of attribute index $\{1,2,3,4\}$ of an image.  $1$ is the light/color attribute, $2$ is the composition and depth attribute, $3$ is the imaging quality attribute, and $4$ is the semantic attribute. To derive the specific importance (the weight values) in AMOS, \cite{kang2020eva} directly elicited from more than 30 observers the importance of each attribute in forming their overall aesthetic opinion. The observers had to indicate which factor(s) influenced their overall aesthetic score among the rated attributes. 
From Eq.(\ref{amos}), we observe that light/color is one of the three most important attributes, contributing approximately 30\%. Another work \cite{PIAA} utilizes a Pearson correlation coefficient between the light condition of an image and its aesthetic rating, archiving a high correlation coefficient of 0.67.

Motivated by the above findings, we propose an aesthetics-guided LLE.
The first step is to select an aesthetic `oracle' that can provide general aesthetic preferences with versatility and reflect popular aesthetics. In our implementation, we employ a lightweight aesthetic network \cite{TalebiM18} with a VGG-16 backbone as the aesthetic `oracle', trained on the AVA dataset \cite{murray2012ava}. Though NIMA  \cite{TalebiM18} (81.5$\%$) stands at the middle level in terms of accuracy ({DMA-Net \cite{DMA-Net} 75.4$\%$}; {MTCNN \cite{MTCNN} 79.1$\%$}; {Pool-3FC \cite{Pool-3FC} 81.7$\%$}; {ReLIC \cite{ReLIC} 82.35$\%$}), NIMA is with less computational complexity compared with other available models (ReLIC with a MobileNetV2 backbone has 2.7 times more parameters than NIMA with a MobileNetV1 backbone, while MobileNetV2 has fewer parameters than MobileNetV1).
AVA is a database with 250,000 photos evaluated for aesthetic quality, with crowd-sourcing voting on the aesthetics for each image (from 78 to 549 people for each image). The aesthetic oracle network pre-trained on this vast dataset yields trustworthy preferences for broad aesthetics.

Since human aesthetic image retouching is a dynamically and explicitly progressive process that causally interacts with the current state of the image, we treat LLE as a Markov decision process, decomposing it into a series of iterations. To mimic this process, we design a reinforcement-learning-based Aesthetic Policy Generation Module, which interacts with the environment -- Aesthetic Assessment Module to obtain rewards and provide optimized actions to realize progressive LLE adjustment. 
As shown in Fig.~\ref{Net}, our ALL-E consists of an Aesthetic Policy Generation Module based on an asynchronous advantage actor-critic network (A3C) \cite{mnih2016asynchronous} to generate the action $A^{t}$, and an Aesthetic Assessment Module based on an aesthetic oracle network \cite{TalebiM18} and a series of loss function to generate the reward $r^{t}$. At the $t$-th step, given an image $s^{t}$, the Aesthetic Policy Generation Module generates an enhanced image $s^{t+1}$ via $A^{t}$, which is then fed to the Aesthetic Assessment Module to generate $r^{t}$, then progressively complete image enhancement until $n$ steps. 

\subsection{Aesthetic Policy Generation} 
The reason to use A3C \cite{mnih2016asynchronous} is that it reports pixel-wise action performance with efficient training. 
A3C is an actor-critic method consisting of two sub-networks: a value network and a policy network denoted as $\theta_{v}$ and $\theta_{p}$, respectively. 
 %$s^t$ is the state of step $t$, $r^t$ is the reward of step $t$. 
 Both networks use the current state image $s^{t}$ as the input at the $t$-th step. The value network outputs the value $V(s^{t})$, which represents the expected total discounted rewards from state $s^{t}$ to $s^{n}$, and indicates how good the current state is:
 \begin{equation}
\begin{aligned}
   % \ld{ V(s^{(t)}) = \mathbb{E}\left[r^{(t)}+\gamma r^{(t + 1)}+\ldots +\gamma^{n}r^{(t+n-1)}\mid s^{(t)}\right]}
    V(s^{t}) = \mathbb{E}\left[ R^{t}\mid s^{t}\right]
\end{aligned}
\end{equation}
where $R^{t}$ is the total discounted reward:
\begin{equation}
\begin{aligned}
    % \ld{R^{(t)}} &= r^{(t)} + \gamma r^{(t+1)} + \gamma^{2}r^{(t+2)}+...\\
    %     &+\gamma^{n-1}r^{(t+n-1)}+\gamma^{n}V(s^{(t+n)})
    R^{t} =\sum_{i=0}^{n-t - 1} \gamma^{i} r^{t+i} +\gamma^{n- t}V(s^{n})
\end{aligned}
\end{equation}
where $\gamma^{i}$ is the $i$-th power of the discount factor $\gamma$, $r^{t}$ is the immediate reward (will be introduced in Section 3.4) at $t$-th step. As some actions can affect the reward after many steps by influencing the environment, reinforcement learning aims to maximize the total discounted reward $R^{t}$ rather than the immediate reward $r^{t}$.

 The gradient for $\theta_{v}$ is computed as follows:
\begin{equation}
\begin{aligned}
    d\theta_{v} = \triangledown _{\theta_{v}}(R^{t}-V(s^{t}))^{2}
\end{aligned}
\end{equation}
The policy network outputs the probability of taking action $A^{t} \in A.S.$ (the action space, will be introduced in the following subsection) through softmax, denoted as $\pi(A^{t}|s^{t})$.
The output dimension of the policy network is equal to $|A|$. 

To measure the rationality of selecting a specific action $A^{t}$ in a state $s^{t}$, we define the advantage function as:
\begin{equation}
\begin{aligned}
    G(A^{t}, s^{t}) = R^{t} - V(s^{t})
\end{aligned}
\end{equation}
It directly gives the difference between the performance of action $A^{t}$ and the mean value of the performance of all possible actions. If this difference (\emph{i.e.} the advantage of the chosen action) is greater than 0, then it indicates that action $A^{t}$ is better than the average and is a reasonable choice; if the difference is less than 0, then it implies that action $A^{t}$ is inferior to the average and should not be selected.

The gradient for $\theta_{p}$ is computed as follows:
\begin{equation}
\begin{aligned}
    d \theta_{p} = -\triangledown_{\theta_{p}} \log \pi(A^{t}|s^{t})G(A^{t}, s^{t})
\end{aligned}
\end{equation}
 More processes can be found in the supplementary material.

\subsection{Action Space Setting} 
Human experts often manually retouch photographs through curve adjustments in retouching software, where the curve parameters depend only on the input image. Usually, curves for challenging low-light images are of a high order. \cite{Guo_2020_CVPR} suggests that this procedure can be realized by recurrently applying the low-order curves. 
In this work, we apply a second-order pixel-wise adjustment curve (PAC) at each step $t$.
We enhance an input image $s^{t}$ by iteratively applying a PAC-based action $A^{t}(x)$. At the $t$-th step, the enhanced output is:
\begin{equation}
\begin{aligned}
    s^{t + 1}(x) = s^{t}(x)+A^{t}(x)s^{t}(x)(1-s^{t}(x))
\end{aligned}
\label{aa}
\end{equation}
where $x$ denotes the pixel coordinates. Eq.(\ref{aa}) models the enhancement of a low-light image as a sequential action-making problem by finding the optimal pixel-wise parameter map $A^{t}(x)$ for light adjustment curves at each step $t$. 
Therefore, our optimization goal is to find an optimal light adjustment action sequence. 
To achieve this, we need a metric (the reward) to measure the light aesthetic of an image $s^{t}$. The exact calculation is described in the following subsection.

Fundamentally, low-light image enhancement can be regarded as the search for a mapping function ${F}$, such that $s_H = {F}(s_L)$ is the desired image, which is enhanced from the input image $s_L$.
In our design, the mapping function $F$ is represented as the ultimate form of multiple submappings $\left\{A^{1}, A^{2}, \ldots, A^{n}\right\}$ continuously and iteratively augmented, 
where $A^{t}$ is constrained in a predetermined range of the action space ($A.S.$).
The range of $A.S.$ is crucial to the performance of our method, as too narrow a range will result in insufficient improvements, and too extensive a range will result in excessive search space. 

Here, we empirically set the range of $A.S.\in[-0.5, 1]$ with a graduation interval $1/18$. This setting ensures that:

1) The brightness of each pixel is in the normalized range $[0, 1]$, according to Eq.(7); 

2) As the brightness of some pixels may need to be reduced, a negative range $[-0.5, 0]$ is also set;

3) PAC is monotonous while also alleviating the cost of searching for suitable PAC for low-light image enhancement. 

\subsection{Reward Function}
This section introduces three complementary rewards, including aesthetics quality, feature preservation, and exposure control, to preserve better subjective visual experience and objective evaluation. Note that all the rewards are non-reference, requiring no paired training images.

\textbf{Aesthetics Quality Reward.}
As mentioned above, the aesthetic quality score of an image is closely correlated with several factors.
In this work, we focus on dynamically adjusting and improving the brightness by the aesthetic score of an image. Therefore, utilizing the aesthetic score as a direct reward function would be an inadequate representation of the desired outcome.  
Instead, the difference in aesthetic scores between the original and enhanced images is employed as the reward for the currently selected action. The image aesthetics quality reward $r^t_{aes}$ is:

\begin{equation}
\begin{aligned}
   r^t_{aes} = \sum_{k=1}^{K} k(P_{k}(s^{t+1}) - P_{k}(s^{t}) )
\end{aligned}
\end{equation}
where $K$ denotes the range of ratings for the aesthetic scores of the images, ~\emph{i.e.}, [1, 10], and $P$ denotes the probability of each rating. $s^{t}$ denotes the state of the image at $t$-th step, $s^{t+1}$ denotes the state of the image at $t + 1$-th step.

\textbf{Feature Preservation Reward.}
Since color naturalness is a concern in low-light image enhancement, we introduce a color constancy term incorporating an illumination smoothness penalty term as the feature preservation reward. 
It is based on the gray-world color constancy hypothesis \cite{buchsbaum1980spatial,Guo_2020_CVPR}, which posits that the average pixel values of the three channels tend to be of the same value. $r^t_{fea}$ constrains the ratio of the three channels to prevent potential color deviations in the enhanced image. In addition, to avoid aggressive and sharp changes between neighboring pixels, an illumination smoothness penalty term is also embedded in $r^t_{fea}$:
\begin{equation}
\begin{aligned}
    r^t_{fea} &=\sum_{\forall (p,q)\in \xi }(J^p-J^q)^2 \\
            &+ \lambda \frac{1}{n}\sum_{t=1}^{n}\sum_{p\in \xi }(\left |{\triangledown _x{(A^{t})}^{p}} \right|+\left |{\triangledown _y{(A^{t})}^{p}} \right|)
\end{aligned}
\end{equation}
where $\xi=\left \{ R,G,B \right \}$, $J^p$ denotes the average intensity value of $p$ channel in an image, $(p,q)$ represents a pair of channels, $n$ is the number of the steps, and $\triangledown _x$ and $\triangledown _y$ denote the horizontal and vertical gradient steps, respectively. We set $\lambda$ to 100 in our experiments to achieve the best results.

\begin{figure*}[h]
\centering
\subfigure[Input]{
\includegraphics[width=3.2cm]{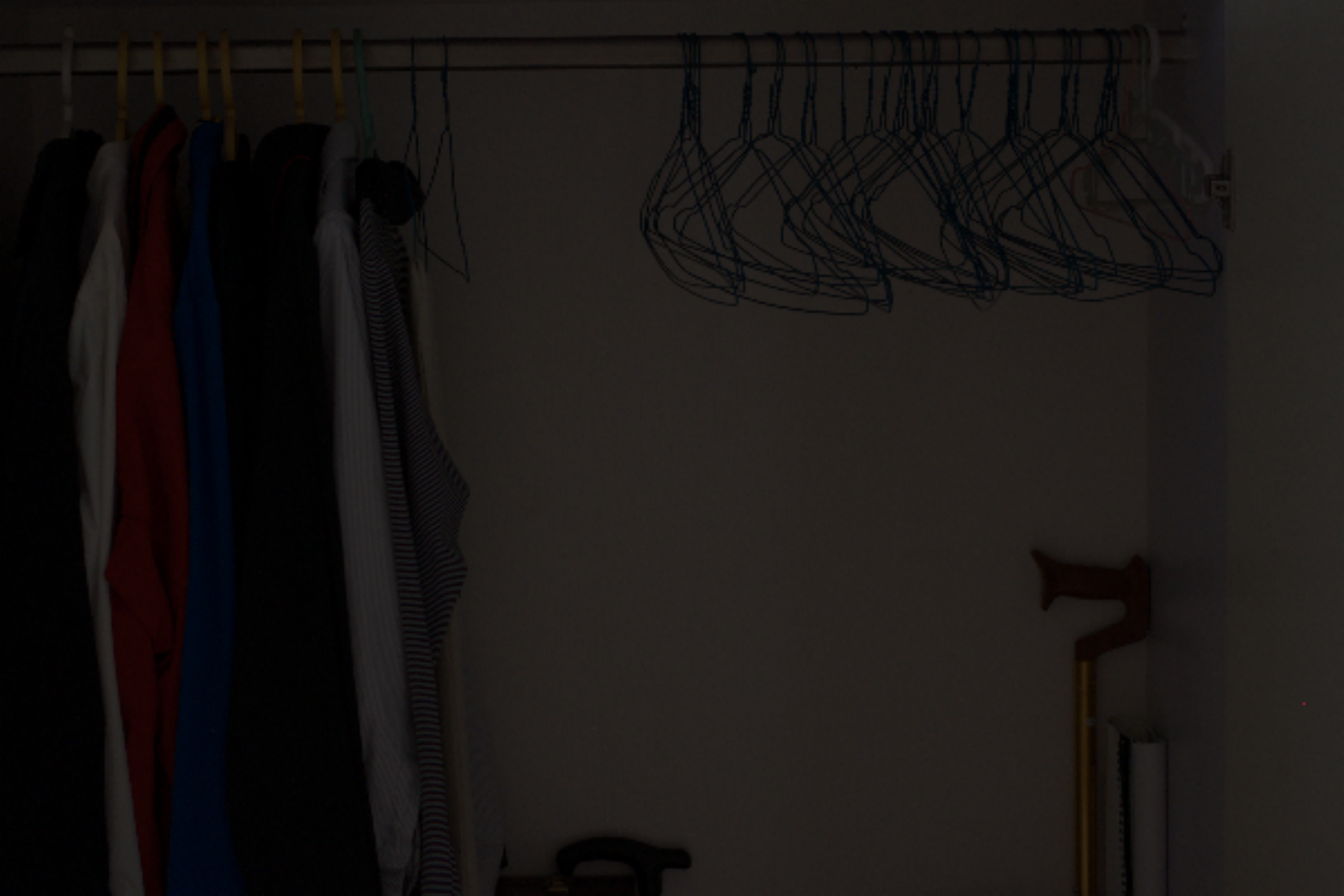}
}
\subfigure[LIME]{
\includegraphics[width=3.2cm]{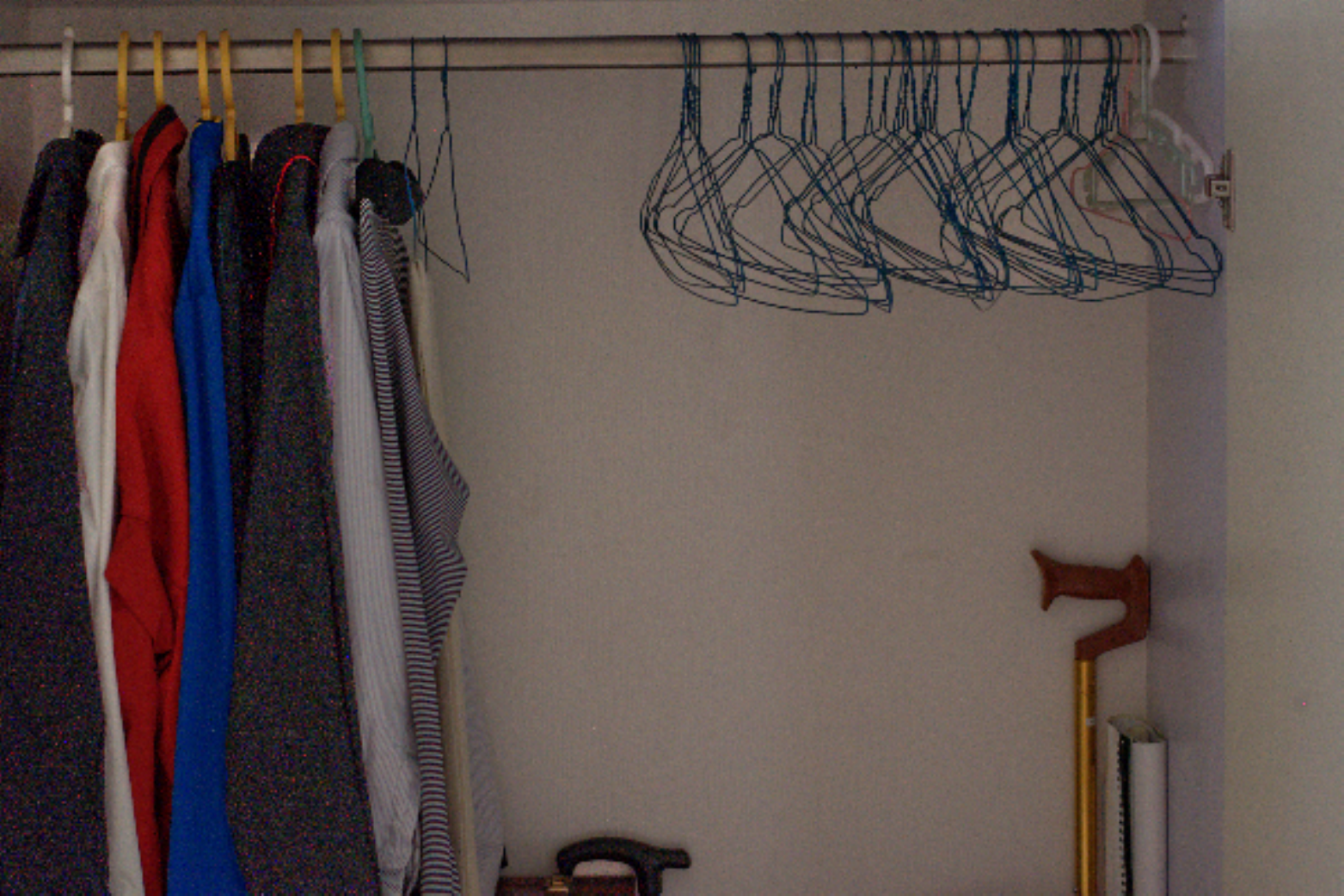}
}
\subfigure[Retinex-Net]{
\includegraphics[width=3.2cm]{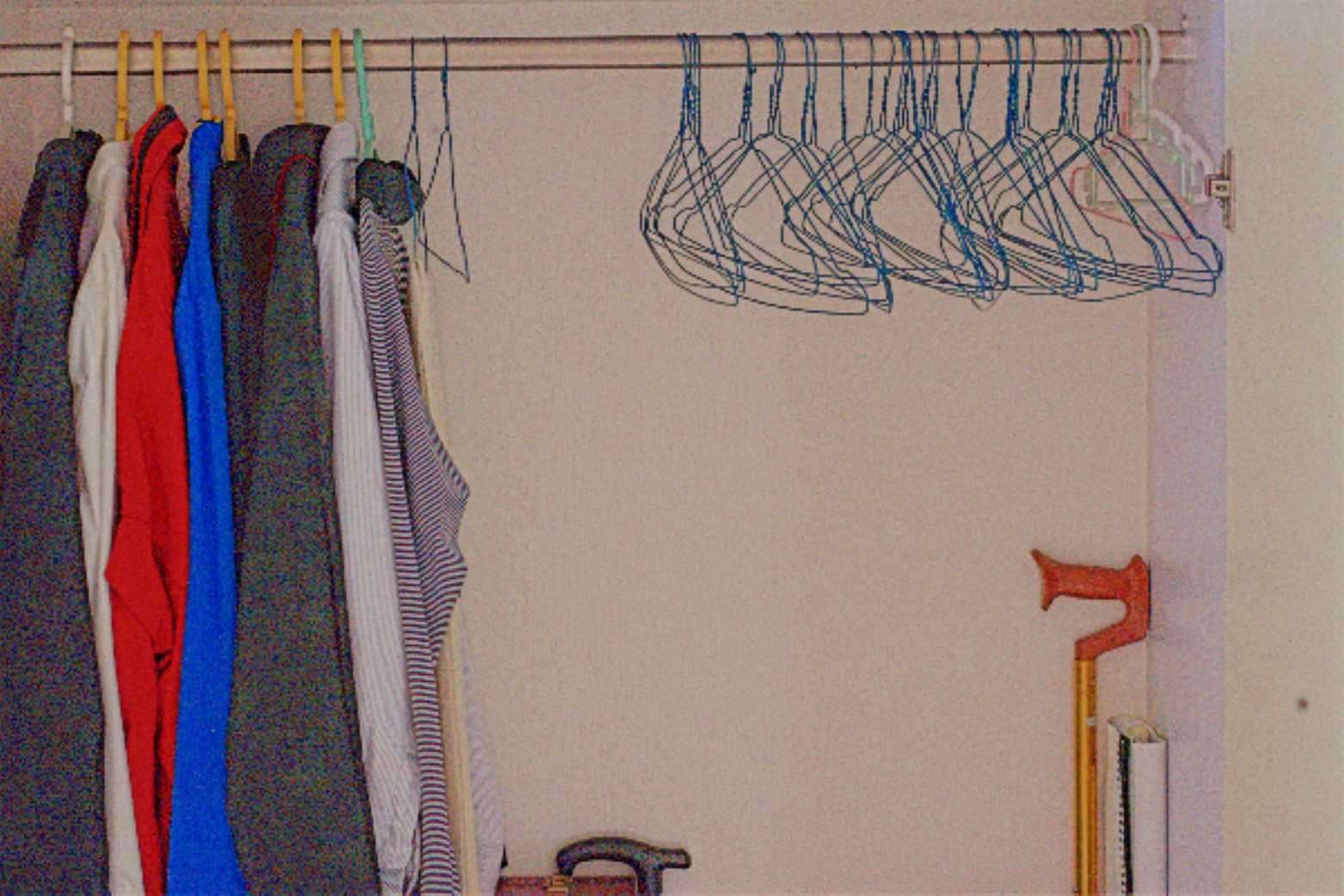}
}
\subfigure[ISSR]{
\includegraphics[width=3.2cm]{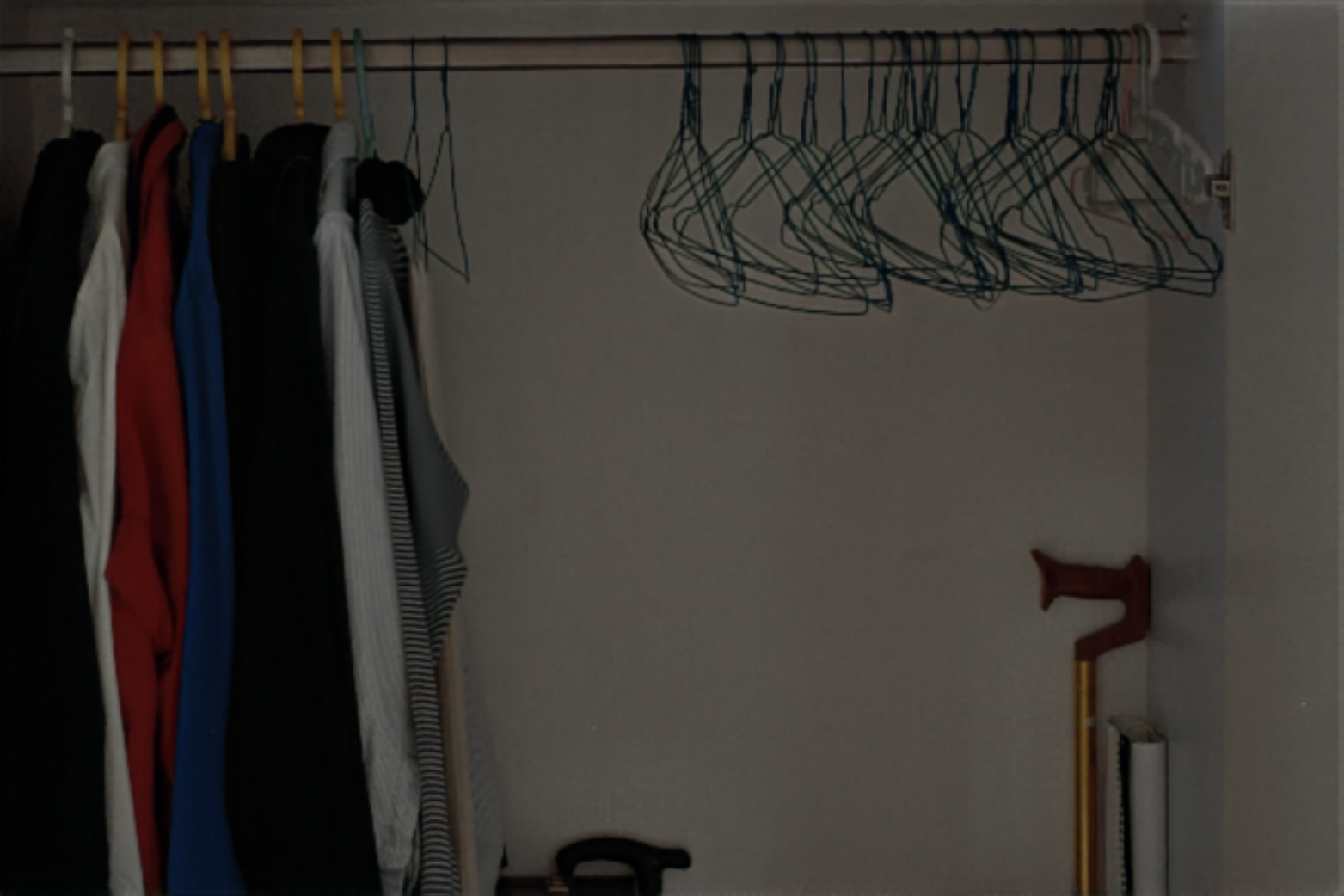}
}
\subfigure[Zero-DCE]{
\includegraphics[width=3.2cm]{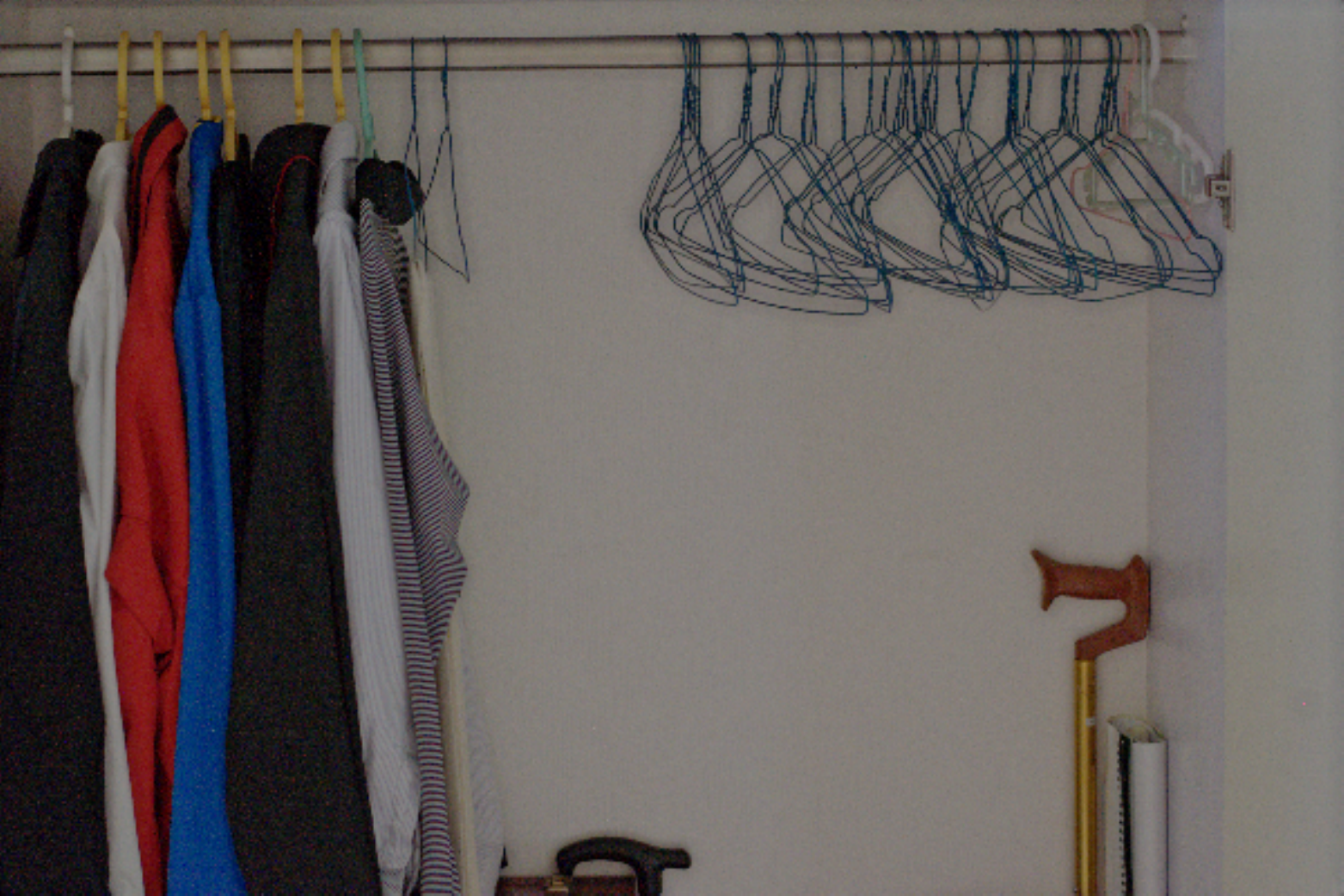}
}
\quad
\subfigure[EnlightenGAN]{
\includegraphics[width=3.2cm]{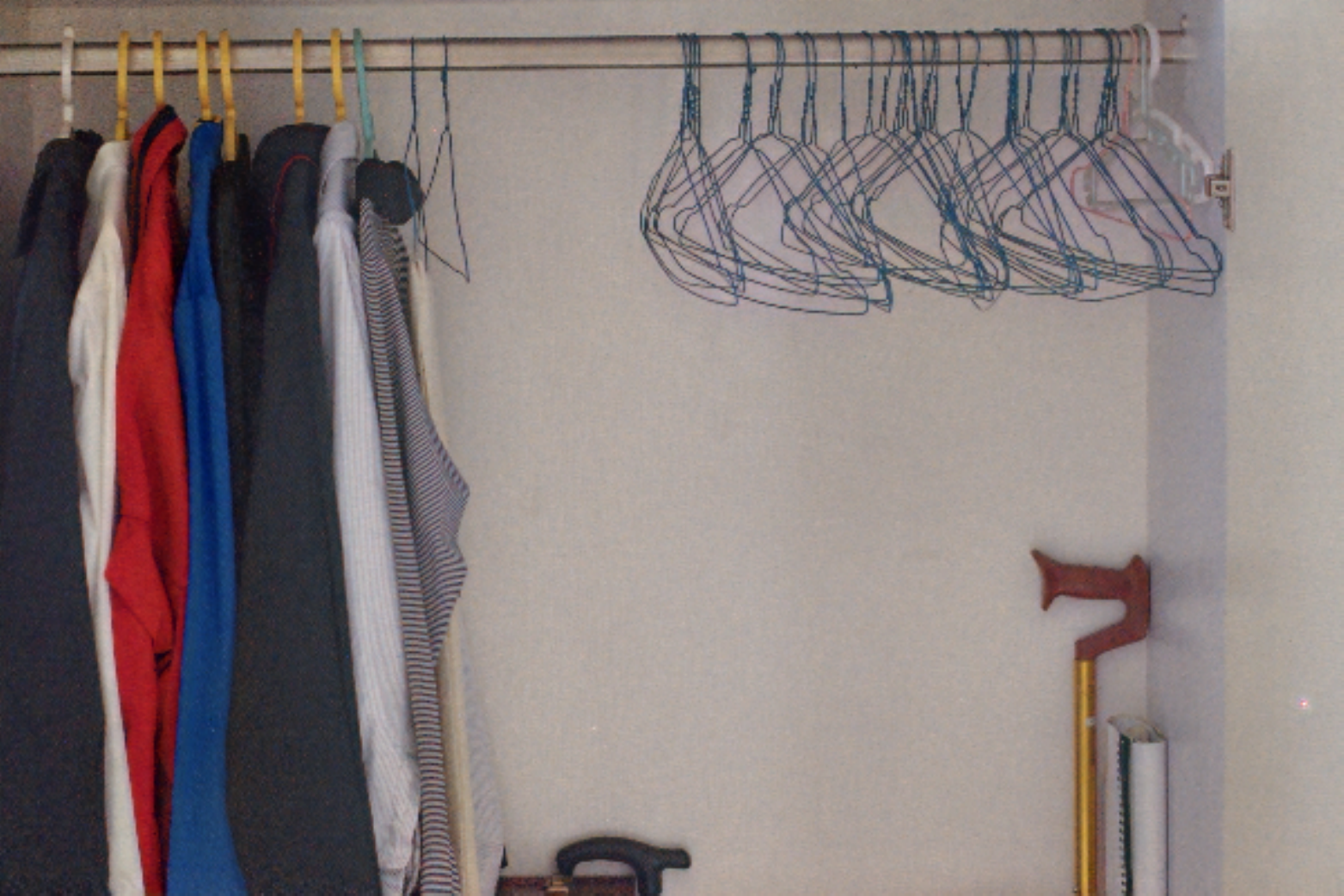}
}
\subfigure[RUAS]{
\includegraphics[width=3.2cm]{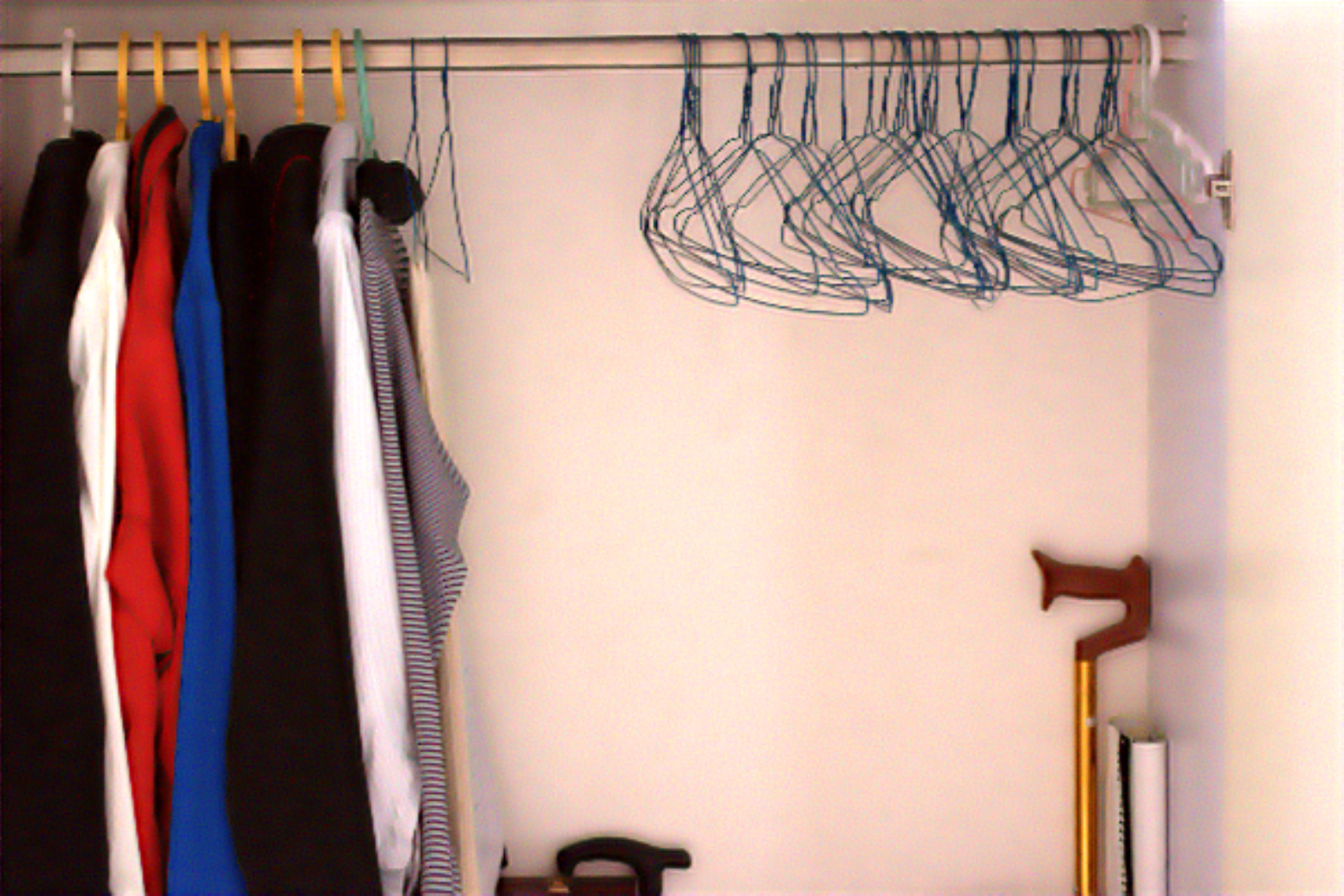}
}
\subfigure[ReLLIE]{
\includegraphics[width=3.2cm]{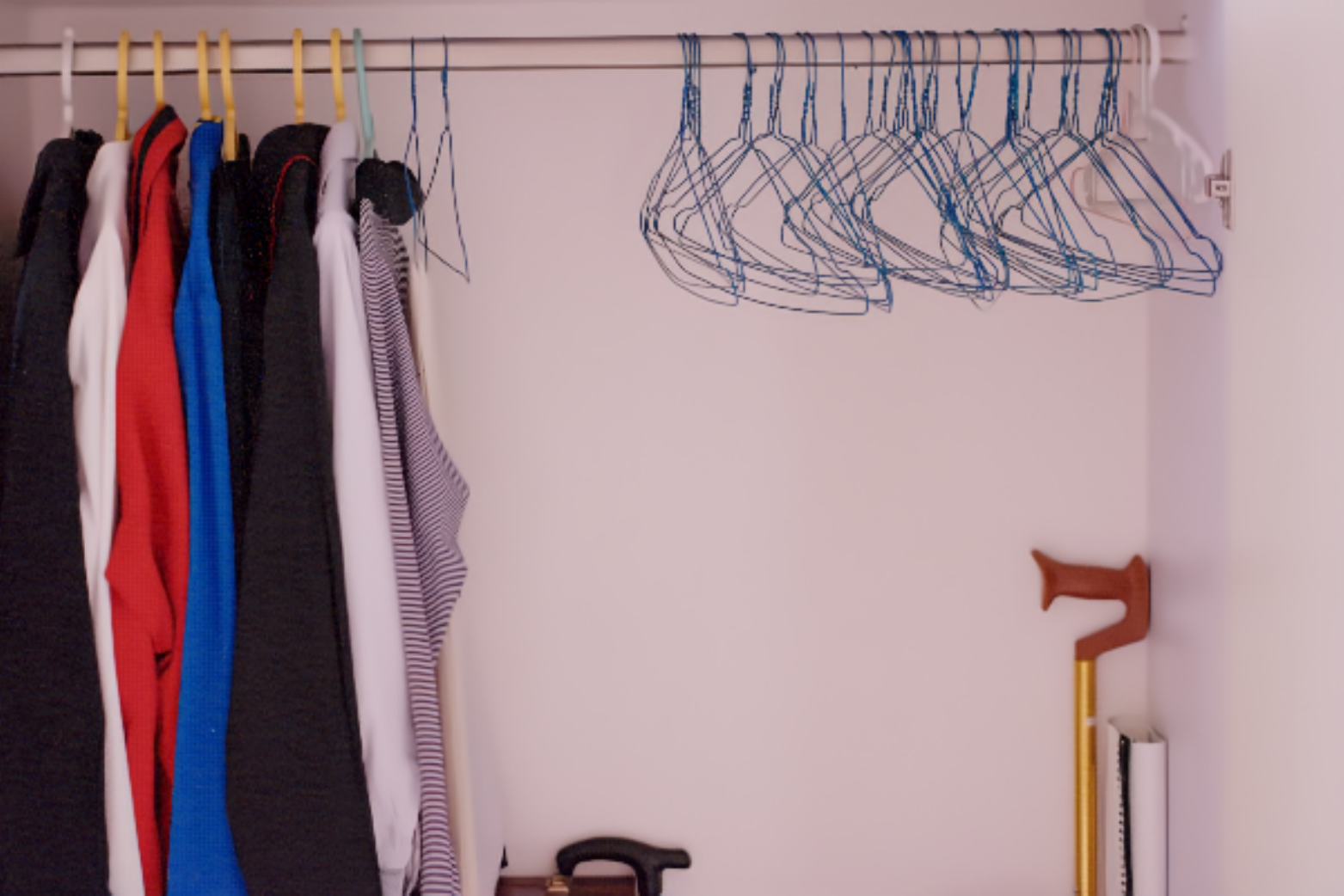}
}
\subfigure[SCL-LLE]{
\includegraphics[width=3.2cm]{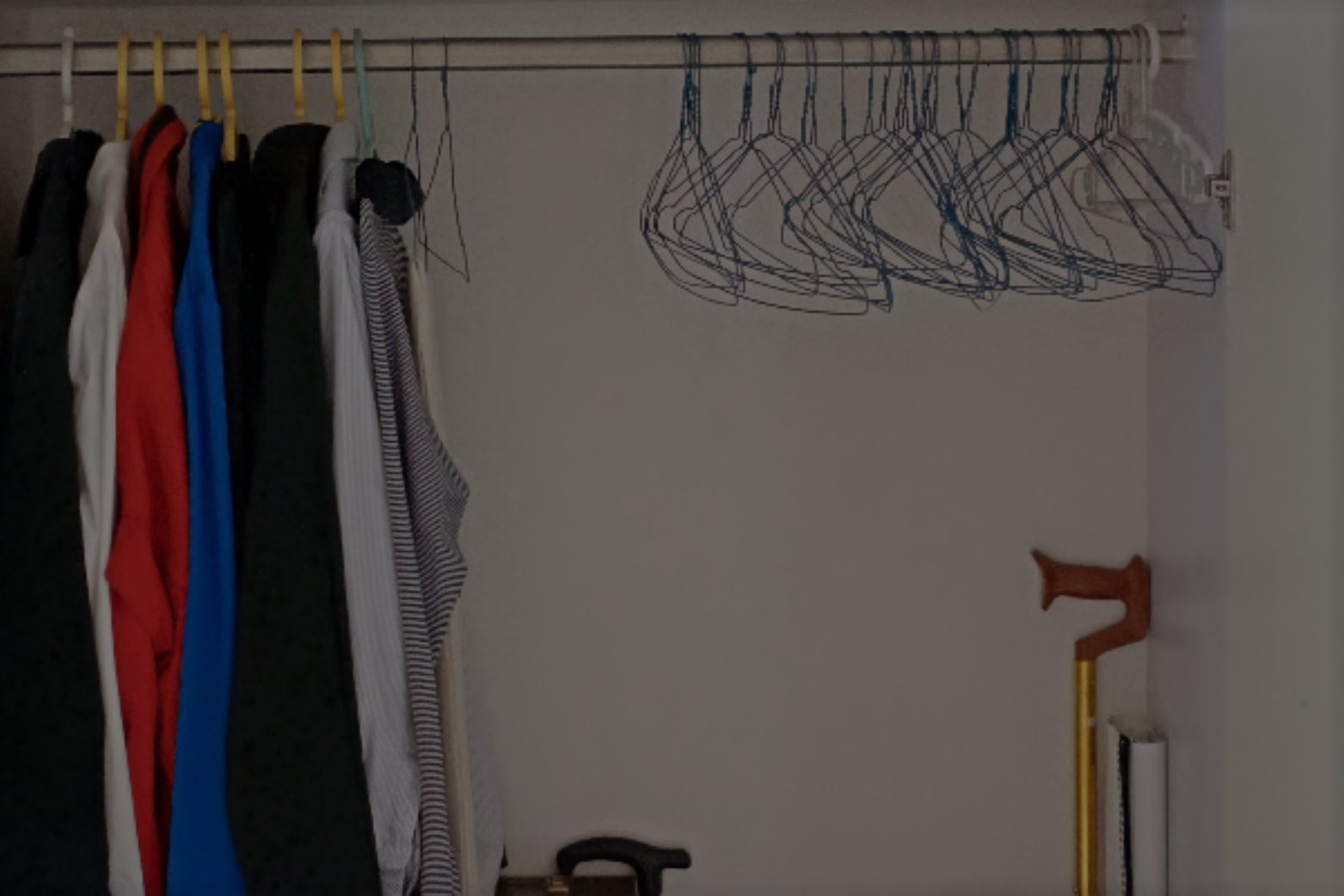}
}
\subfigure[Ours]{
\includegraphics[width=3.2cm]{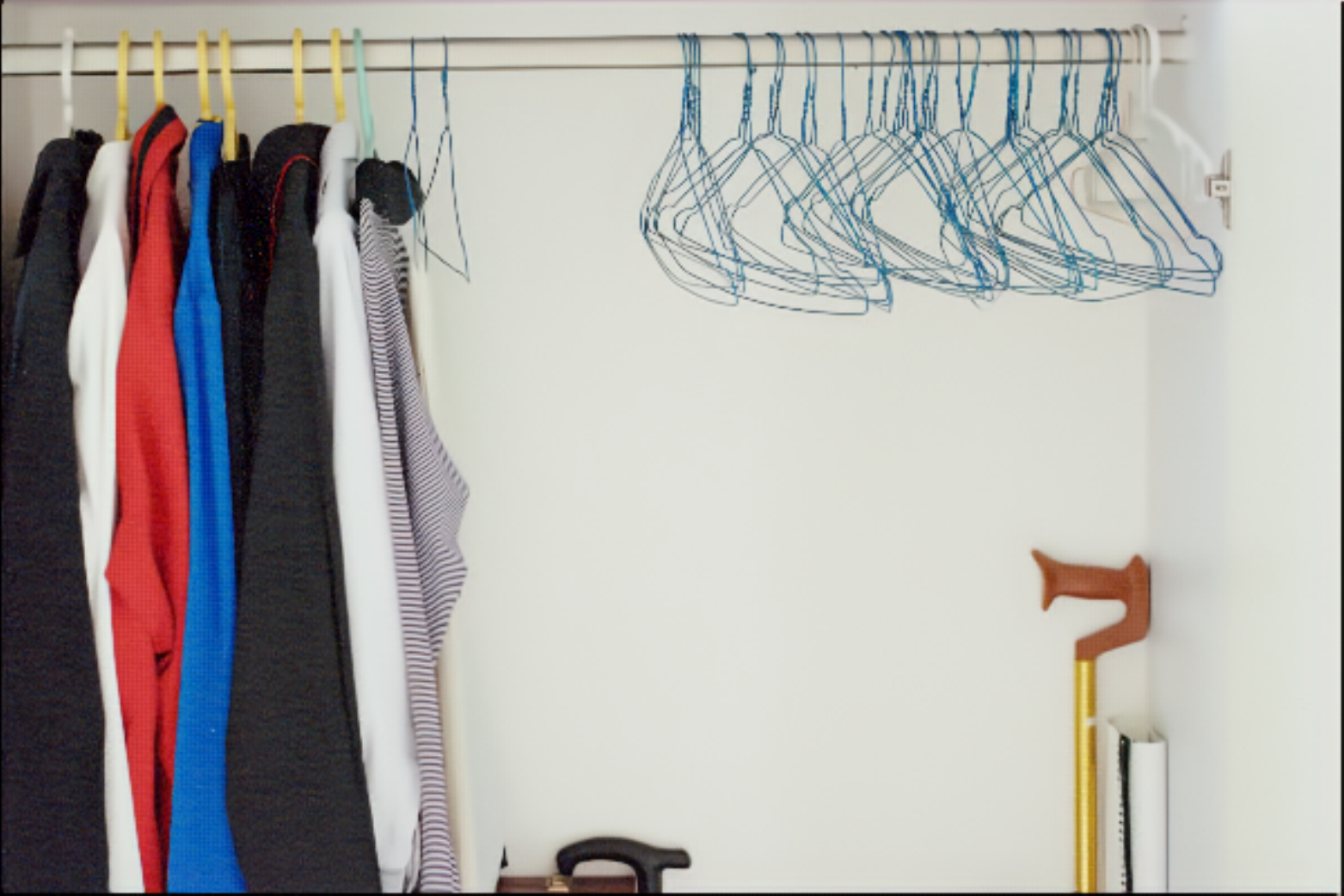}
}
\vspace{-.21cm}
\caption{Examples of enhancement results on LOL test dataset.}
\label{v1}
\end{figure*}

\begin{table*}[!htb]
\centering
\small
%\scalebox{2.1\columnwidth}{!}
{
\begin{tabular}{c|ccccc}
\hline
Methods & NIQE $\downarrow$ & UNIQUE $\uparrow$  &PSNR $\uparrow$ &SSIM $\uparrow $ &User study$\downarrow$ \\ \hline
Input &6.749 &-0.144  &7.773 &0.194 &4.333\\
(TIP'17) LIME &8.058 &0.333  &14.221 &0.521 &3.855\\ 
(BMVC'18) Retinex-Net &8.879 &-0.026 &16.774 &0.424 &3.277\\
(ACMMM'20) ISSR &3.872 &0.739 &12.469 &0.525 &3.950\\
(CVPR'20) Zero-DCE &7.767 &0.335 &14.860 &0.562 &3.286\\
(TIP'21) EnlightenGAN &5.807 &0.546  &17.654 &0.666 &3.156\\
(CVPR'21) RUAS &6.340 &0.427 &16.405 &0.503 &3.431\\    
(ACMMM'21) ReLLIE &4.535& 1.133& \textbf{19.454}& 0.756 &2.677\\
(AAAI'22) SCL-LLE &4.571 &0.544 &12.354 &0.591 &3.270\\\hline    
Ours &\textbf{3.774} &\textbf{1.227} &18.216 &\textbf{0.763} &\textbf{2.450}\\ 
\hline
\end{tabular}}
\caption{NIQE $\downarrow$, UNIQUE $\uparrow$, PSNR $\uparrow$, SSIM $\uparrow$ and User study $\downarrow$ scores on LOL test dataset. 
}
\label{tab1}
\end{table*}

\begin{table*}[!htb]
\small
\centering
%\resizebox{2.1\columnwidth}{!}
{
\begin{tabular}{c|cc|cc|cc|cc|cc|cc}
\hline
\multirow{2}*{Methods}  &\multicolumn{2}{c|}{DICM} &\multicolumn{2}{c|}{LIME} &\multicolumn{2}{c|}{MEF} &\multicolumn{2}{c|}{VV} &\multicolumn{2}{c|}{NPE} &\multicolumn{2}{c}{Average} \\
\cline{2-13}
 & NIQE & UN.&NIQE &UN. &NIQE &UN. &NIQE &UN. &NIQE &UN. &NIQE &UN.\\ \hline
Input &4.26 &0.72  &4.36 &0.70 &4.26 &0.72 &3.52 &\textbf{0.74} &4.32 &\textbf{1.17}  &4.13 &0.75  \\
(TIP'17) LIME &3.75 &0.78 &3.85 &0.53& 3.65 &0.65 &\textbf{2.54} &0.44 &4.44 &0.93 &3.55 &0.69 \\ 
(BMVC'18) Retinex-Net &4.47 &0.75 &4.60 &0.52&4.41 &0.97 &{2.70} &0.36 &4.60 &0.81 &4.13 &0.69\\
(ACMMM'20) ISSR &4.14 &0.59 &4.17 & 0.83 & 4.22 &0.87 & 3.57 &0.62 & 4.02 &0.99  &4.03 &0.68\\
(CVPR'20) Zero-DCE &3.56 &0.82 &{3.77} &0.73 & {3.28} &1.22 &3.21 &0.48 &{3.93} &1.07 &3.50 &0.81 \\
(TIP'21) EnlightenGAN &{3.55} &0.63 &\textbf{3.70} &0.49 & \textbf{3.16} &1.03 & 3.25 &0.58 & 3.95 &1.07 &{3.47} &0.69\\
(CVPR'21) RUAS &5.21 &-0.17 &4.26 &0.34& 3.83 &0.73 &4.29 &-0.04 &5.53 &0.13 &4.78 &0.04  \\    
(AAAI'22) SCL-LLE &3.51 &{0.87} & 3.78 &0.76& 3.31 &{1.25}  &3.16 &0.49  &{3.88} &1.08 &{3.46} &{0.85}\\
(CVPR'22)Uretinex-net &3.95 &0.85  &4.34 &\textbf{0.93} &3.79 &1.18 &3.01 & 0.51&4.69 & 0.99 & 3.83& 0.84  \\
(ICCV'21)Zhao \emph{et al.} &3.68 &\textbf{0.91} &4.16 &0.79 &3.83 &0.97 &3.01 &0.57 &\textbf{3.69} &1.06  &3.61&0.84  \\
(CVPR'22)Ma \emph{et al.} &4.11 &0.11  &4.21 &0.35 &3.63&1.04 &2.92 &0.05 &4.47 &0.21  &3.87 &0.35  \\
\hline    
Ours &\textbf{3.49} & 0.88 & 3.78 &0.80& 3.32 &\textbf{1.27}  &3.08 &0.49  &3.85 &1.10 &\textbf{3.45} &\textbf{0.88} \\ 
\hline
\end{tabular}}
\caption{NIQE $\downarrow$ and UNIQUE (UN.) $\uparrow$ 
scores on DICM, LIME, MEF, VV, and NPE datasets. 
}
\label{tab2}
\vspace{-0.2cm}
\end{table*}

\textbf{Exposure Control Reward.} 
The exposure control reward $r^t_{exp}$, which is a loss function widely used in the recent literature \cite{Guo_2020_CVPR,zhang2021rellie}, measures the deviation of the average intensity value of a local region from a predefined well-exposedness level $E$, $i.e.$, the gray level in RGB color space:
\begin{equation}
\begin{aligned}
    r^t_{exp} = \frac{1}{B} \sum_{b=1}^{B} |Y_{b}-E|
\end{aligned}
\end{equation}
where $B$ represents the number of non-overlapping local regions of size 16×16, $Y_{b}$ is the average intensity value of a local region $b$ in $s^{t + 1}$. According to \cite{Guo_2020_CVPR,zhang2021rellie}, $E$ is set to 0.6.

Hence, for a given enhanced image, the immediate reward $r^{t}$ at a current state $s^{t}$ is:
\begin{equation}
\begin{aligned}
    r^{t} = w_1 r^t_{aes} - w_2 r^t_{fea} - w_3 r^t_{exp}
\end{aligned}
\label{r}
\end{equation}
where $w_1$, $w_2$ and $w_3$ are tunable hyperparameters. 
As introduced in Section 3.2, the goal of reinforcement learning is to maximize the total discounted reward $R^{t}$ in Eq.(3) with this immediate reward $r^{t}$.

\subsection{Efficient Training Details}
We use 485 low-light images of the LOL dataset \cite{Chen2018Retinex} to train the proposed framework.
We resize the training images to the size of 244×244. The maximum number of training epochs was set to 1000, with a batch size of 2.
%As for the numerical parameters, we set the maximum epoch as 3000 and the batch size as 2. 
We train our framework end-to-end while fixing the weights of the aesthetic oracle network. Our framework is implemented in PyTorch on an NVIDIA 1080Ti GPU. 
The model is optimized using the Adam optimizer with a learning rate of $1e^{-4}$.
The total number of steps in the training phase is set to $n=6$. In accordance with the training phase, the total number of steps is also set to $n=6$ in the testing phase. Under these settings, training 1000 epochs costs about one day.
%More training details are in the supplementary materials.
\section{Experiments}
\subsection{Benchmark Evaluations}
We compare our method with several state-of-the-art methods: LIME \cite{guo2016lime},  Retinex-Net \cite{Chen2018Retinex},  ISSR \cite{FanWY020},  Zero-DCE \cite{Guo_2020_CVPR},  EnlightenGAN \cite{jiang2021enlightengan},  RUAS \cite{liu2021retinex},  ReLLIE \cite{zhang2021rellie},  SCL-LLE \cite{liang2022semantically}, Uretinex-net\cite{wu2022uretinex}, Zhao et al.\cite{zhao2021deep}, Ma et al.\cite{ma2022toward}.
The results of the above methods are reproduced by the publicly available models provided with the recommended test settings. To thoroughly evaluate the proposed method, a comprehensive set of experiments were conducted, including a visual quality comparison, image quality assessment, and human subjective survey, which are discussed in the following sections.
\begin{figure*}[!htb]
\centering
\subfigure[Input]{
\includegraphics[width=3.3cm]{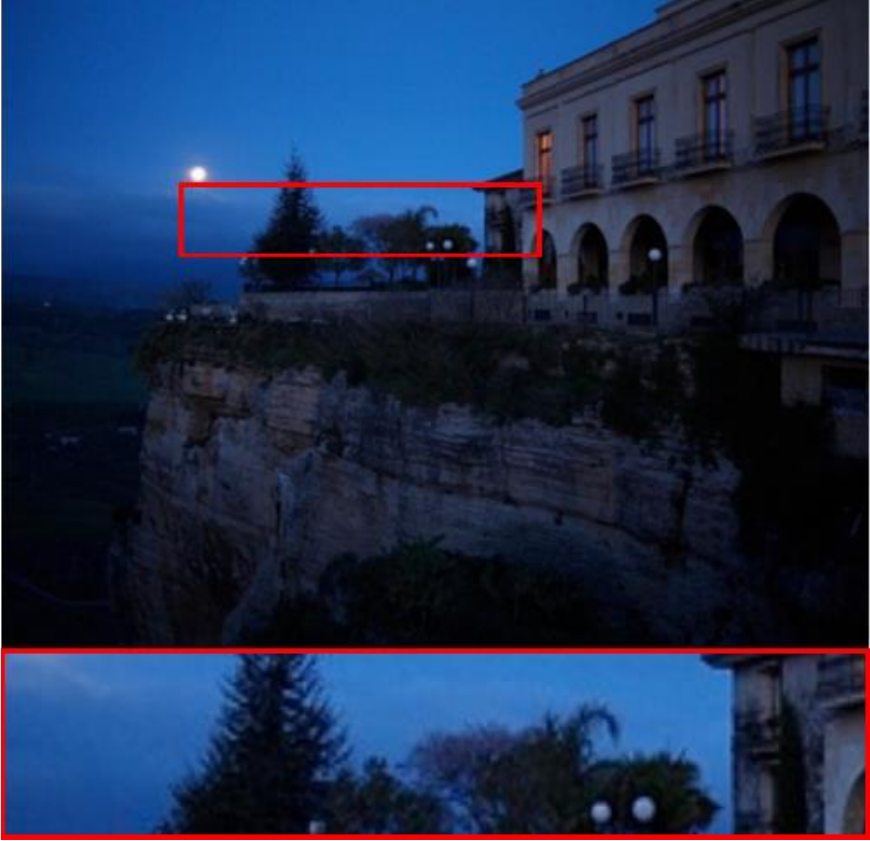}
%\caption{fig1}
}
\subfigure[LIME]{
\includegraphics[width=3.3cm]{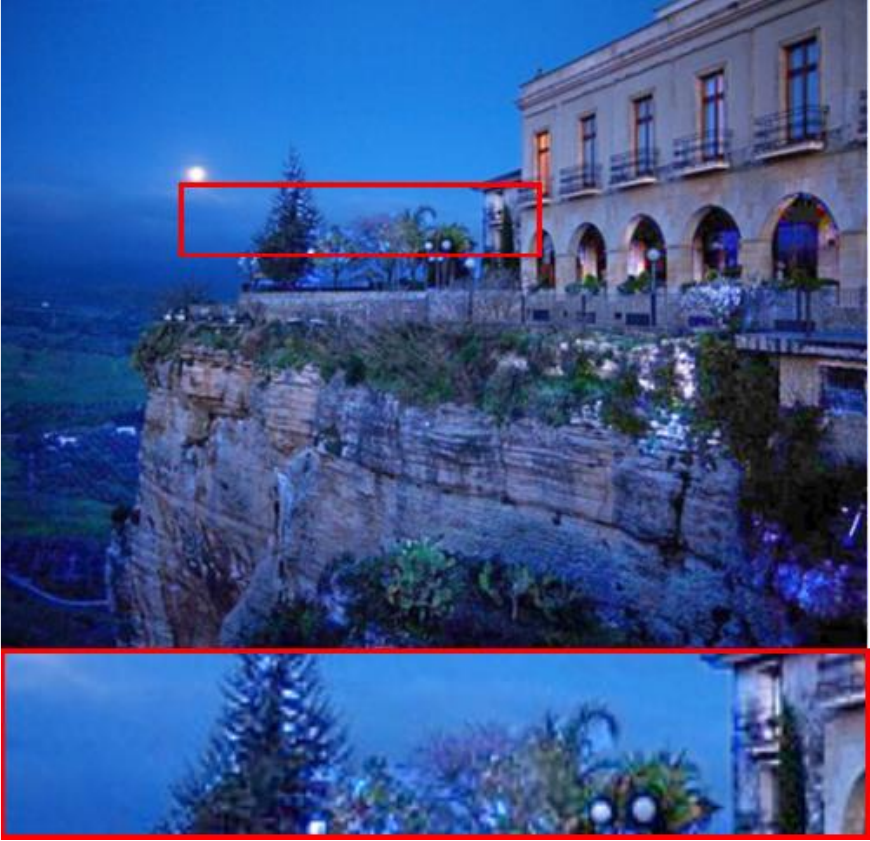}
}
\subfigure[Retinex-Net]{
\includegraphics[width=3.3cm]{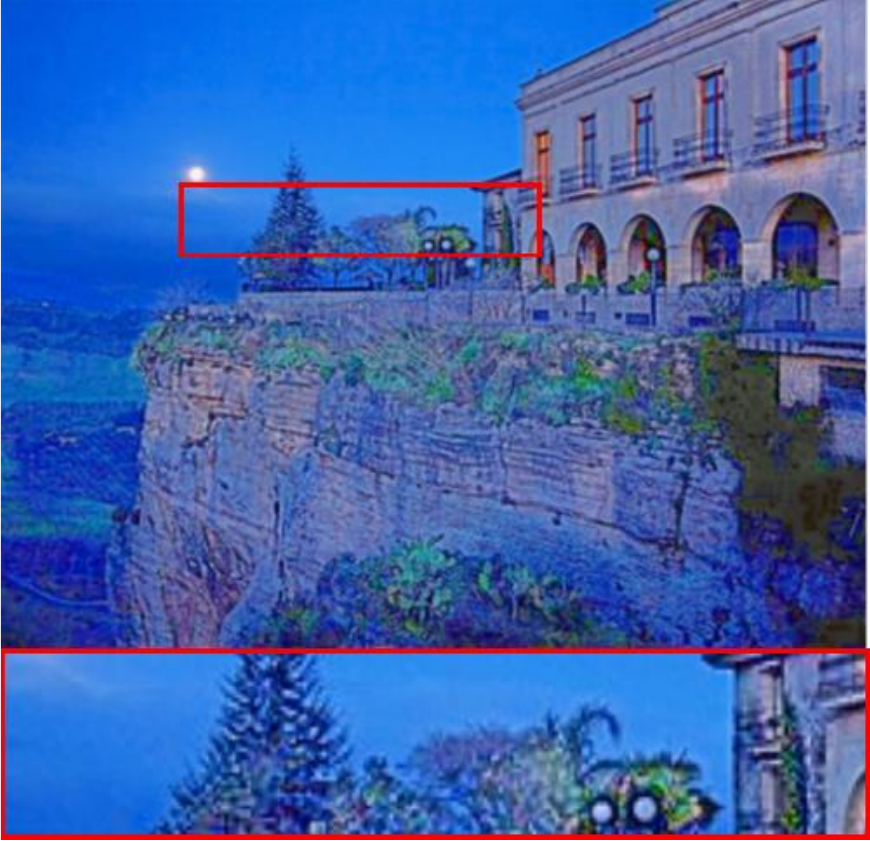}
}
\subfigure[ISSR]{
\includegraphics[width=3.3cm]{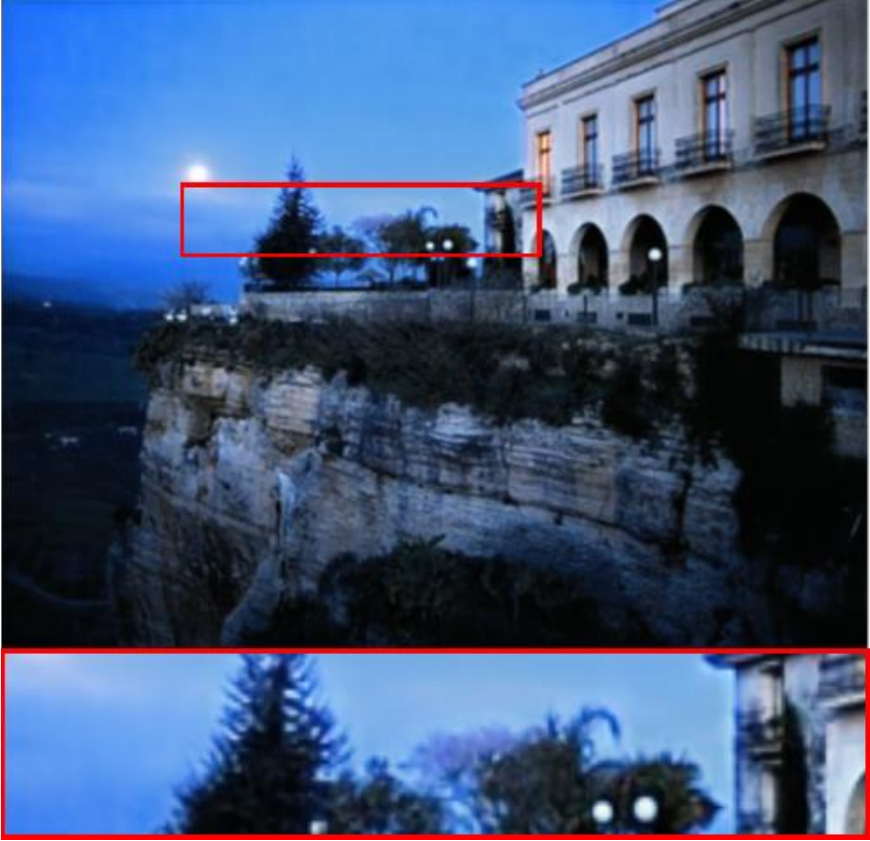}
}
\subfigure[Zero-DCE]{
\includegraphics[width=3.3cm]{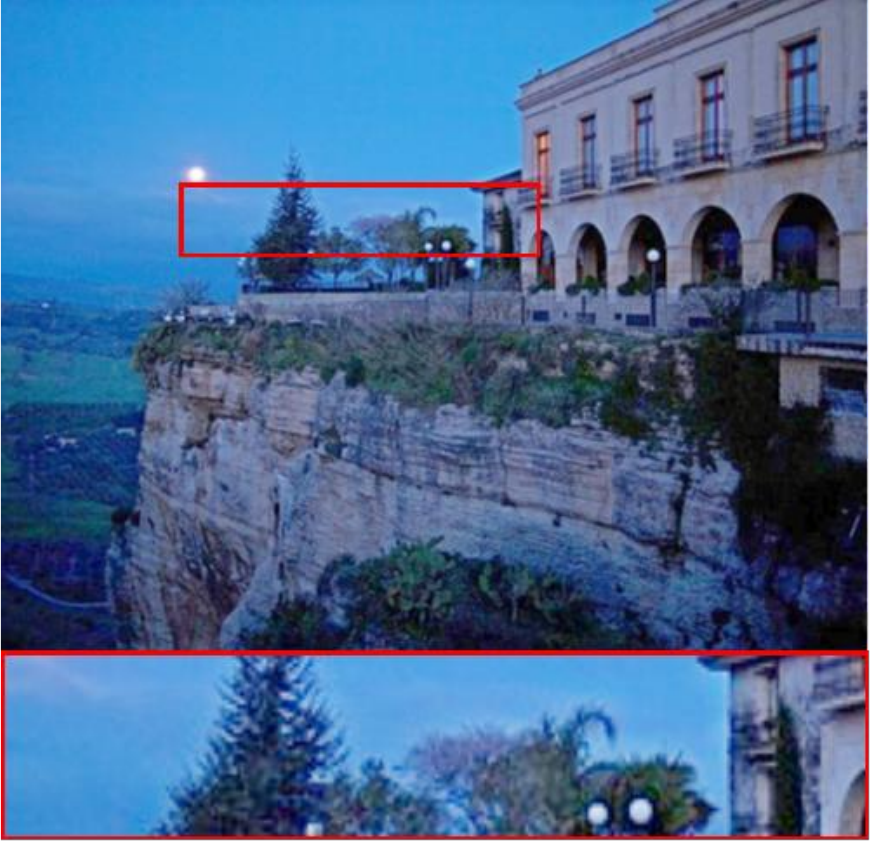}
}
\quad
\subfigure[EnlightenGAN]{
\includegraphics[width=3.3cm]{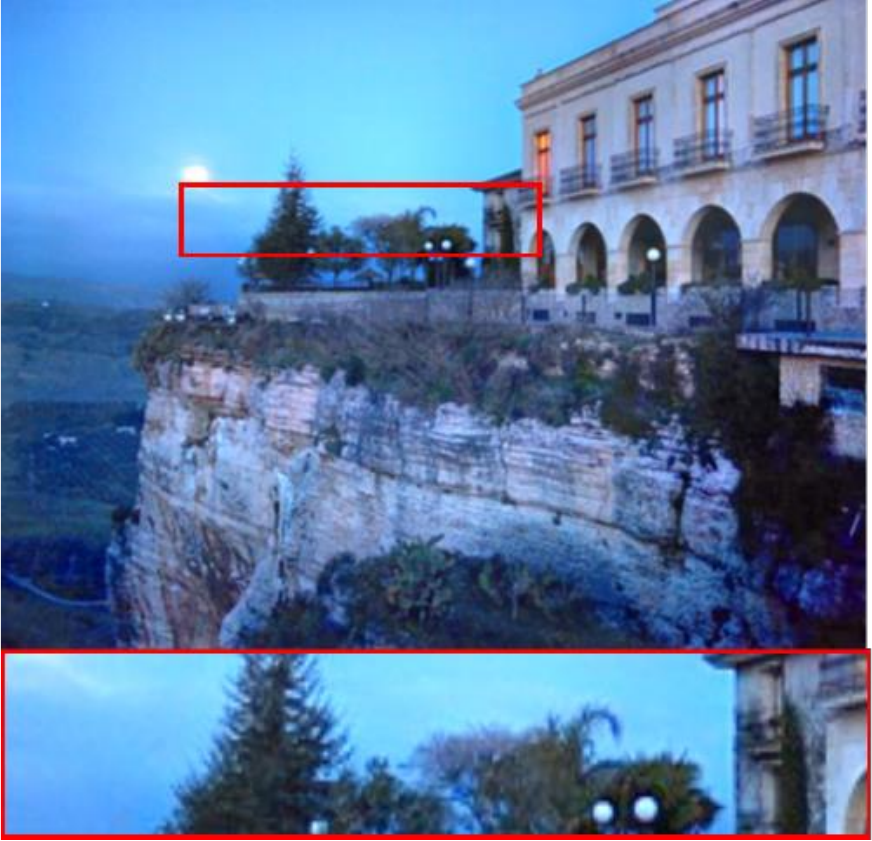}
}
\subfigure[RUAS]{
\includegraphics[width=3.3cm]{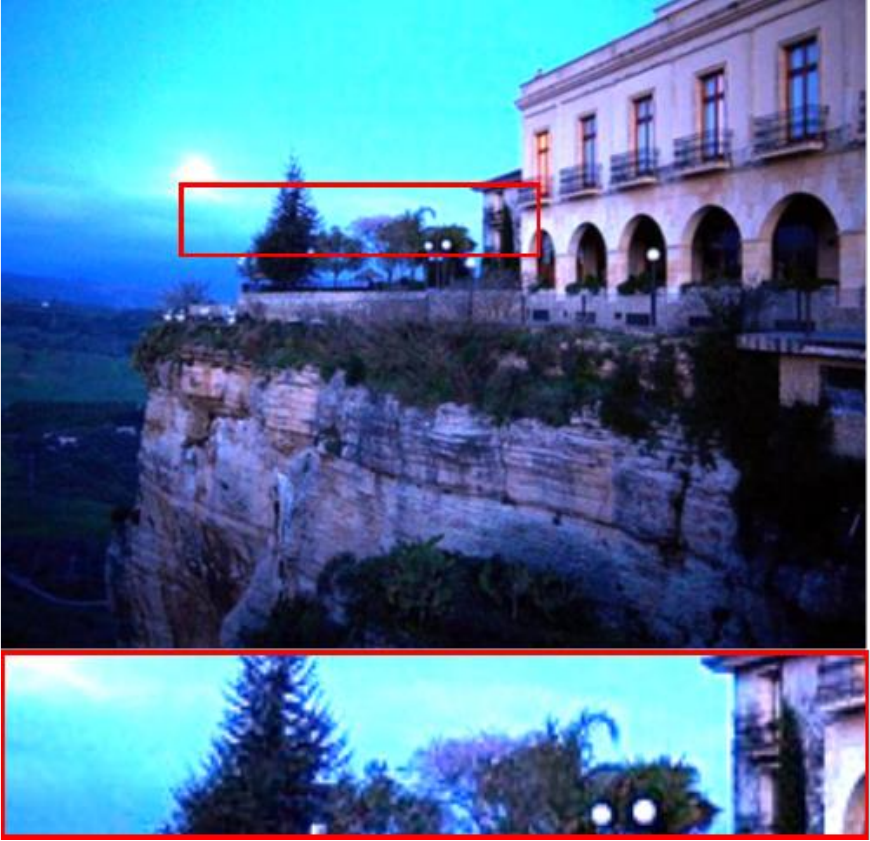}
}
\subfigure[ReLLIE]{
\includegraphics[width=3.3cm]{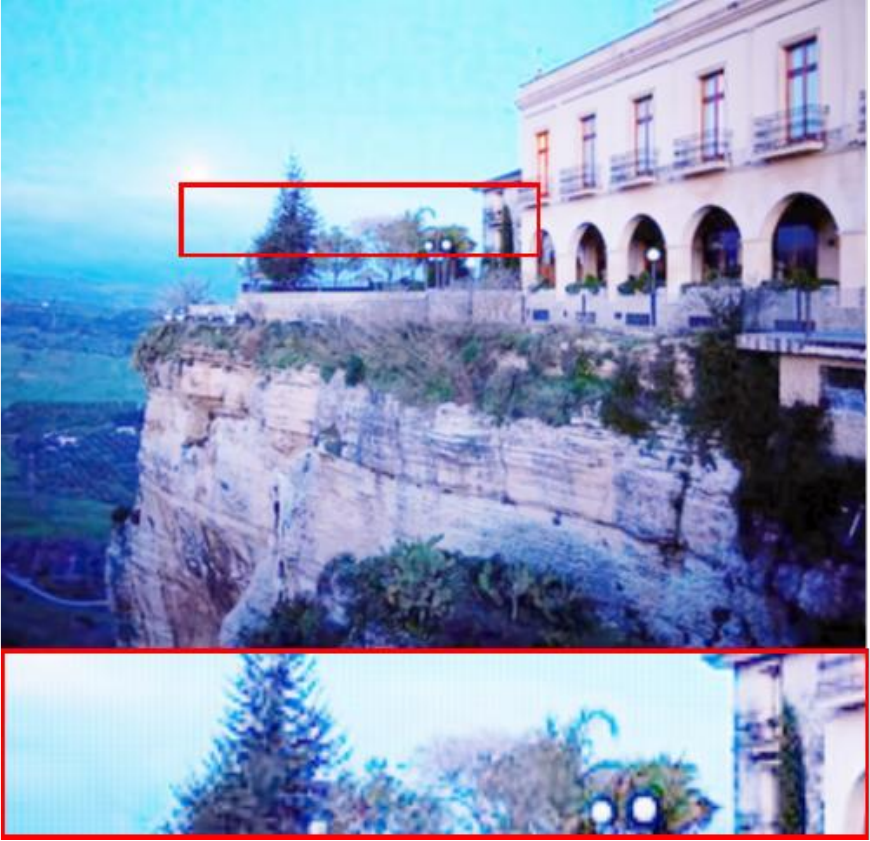}
}
\subfigure[SCL-LLE]{
\includegraphics[width=3.3cm]{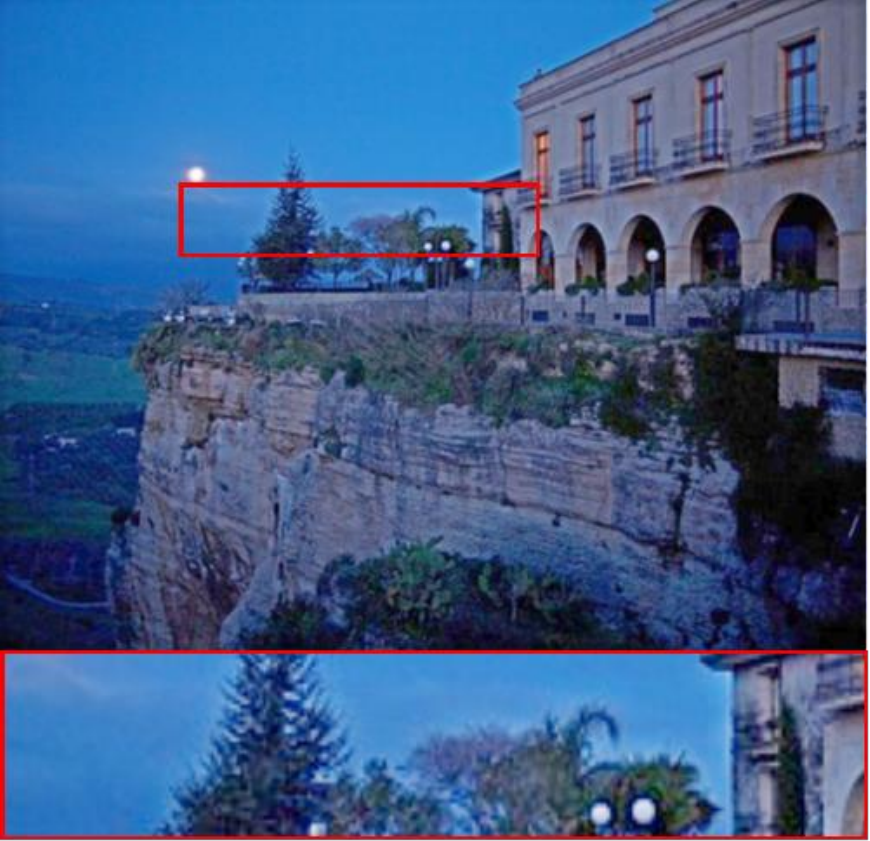}
}
\subfigure[Ours]{
\includegraphics[width=3.3cm]{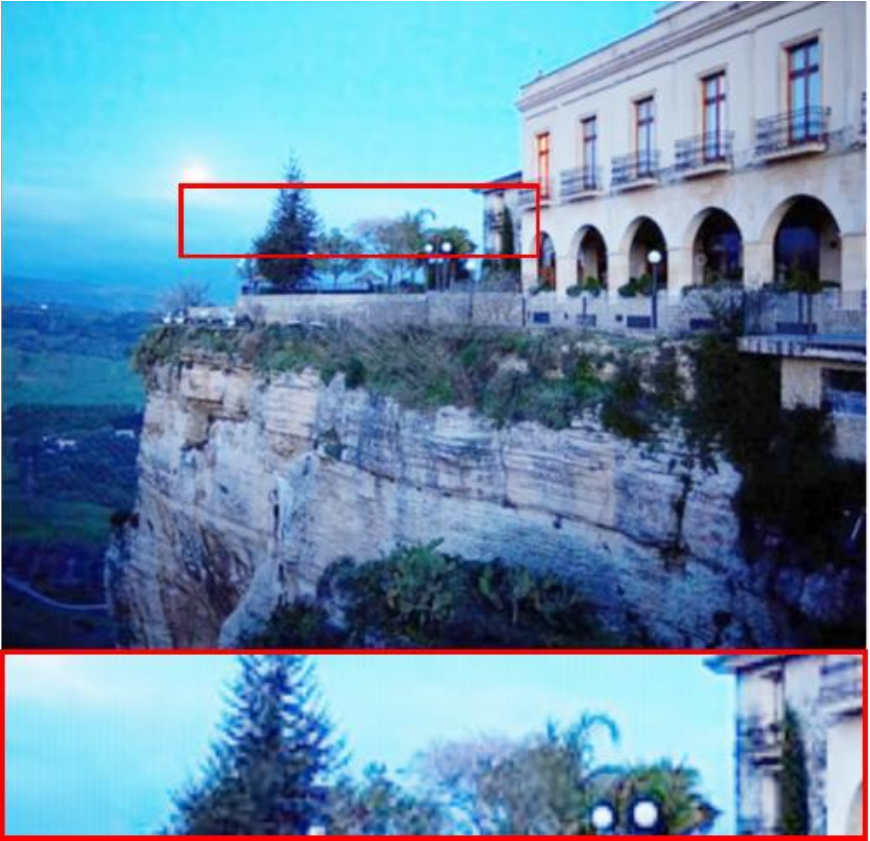}
}
\vspace{-0.1cm}
\caption{Comparison of our method and the state-of-the-art methods over LIME dataset with zoom-in regions. Our method enables the enhanced images to look more realistic and recovers better details in both foreground and background. }
\label{v2}
\vspace{-0.1cm}
\end{figure*}

\subsubsection{Visual Quality Comparison}

\begin{figure*}[!htb]
\centering
\subfigure[Input]{
\includegraphics[width=4.2cm]{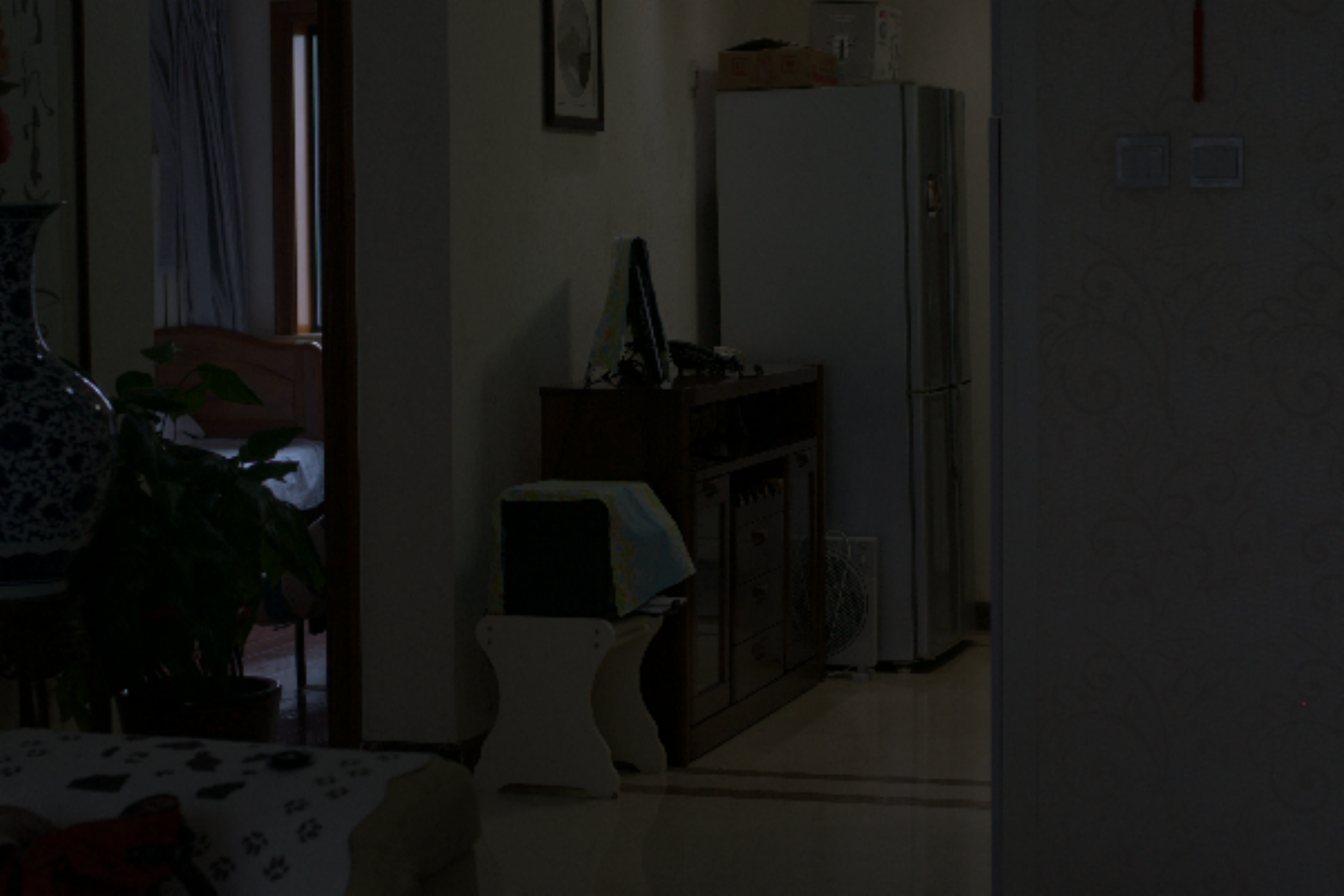}
}
\subfigure[$t = 1$]{
\includegraphics[width=4.2cm]{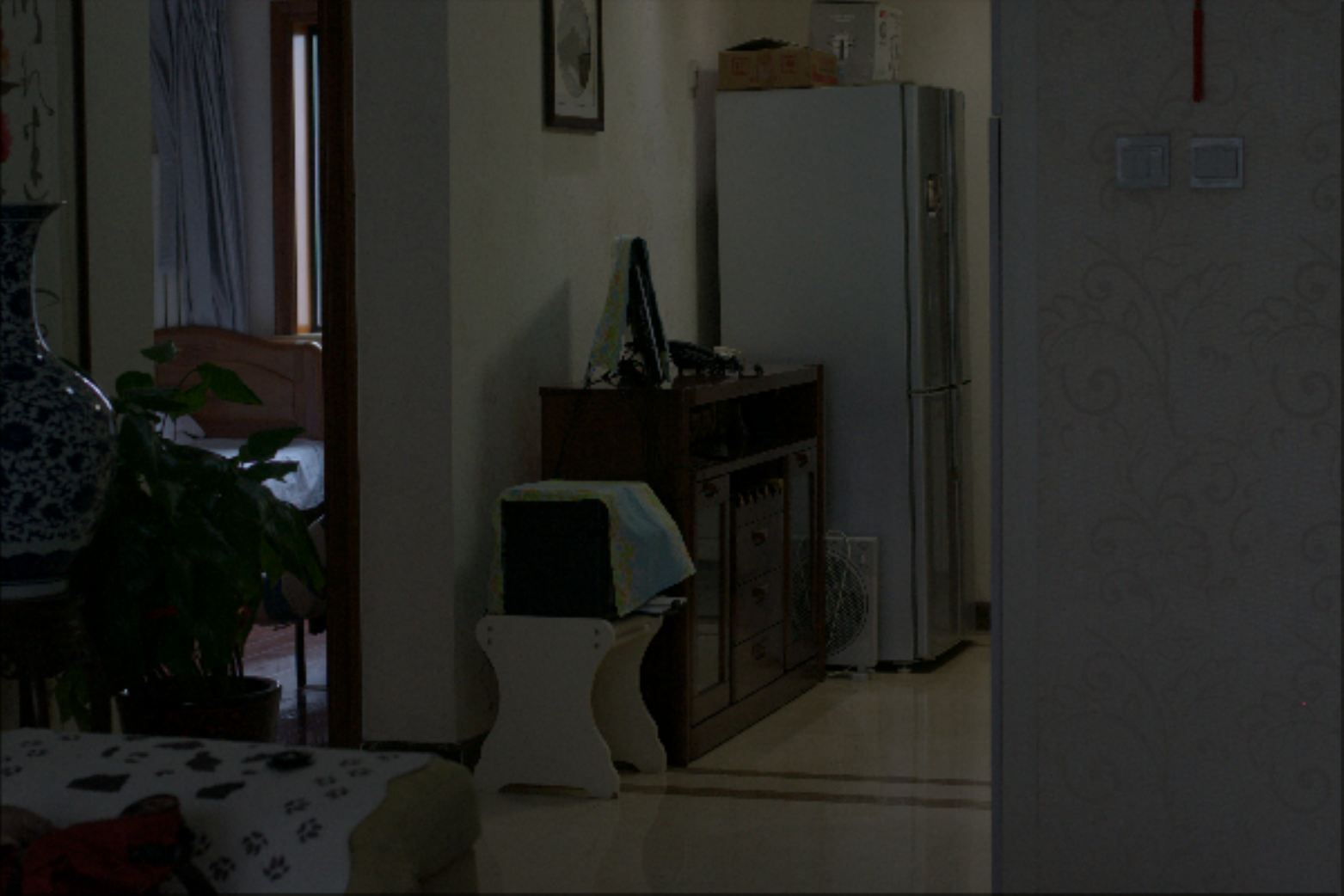}
}
\subfigure[$t = 2$]{
\includegraphics[width=4.2cm]{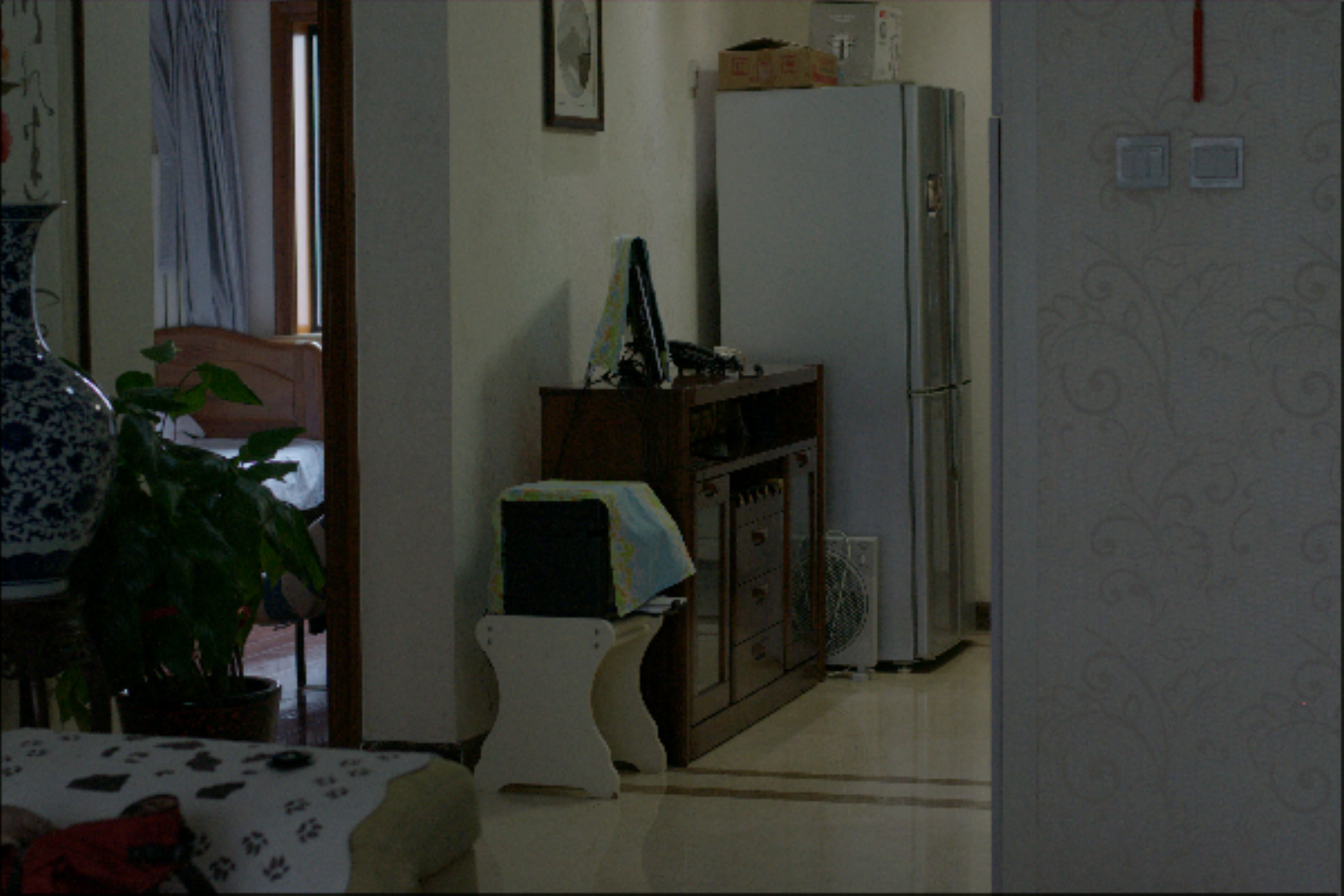}
}
\subfigure[$t = 3$]{
\includegraphics[width=4.2cm]{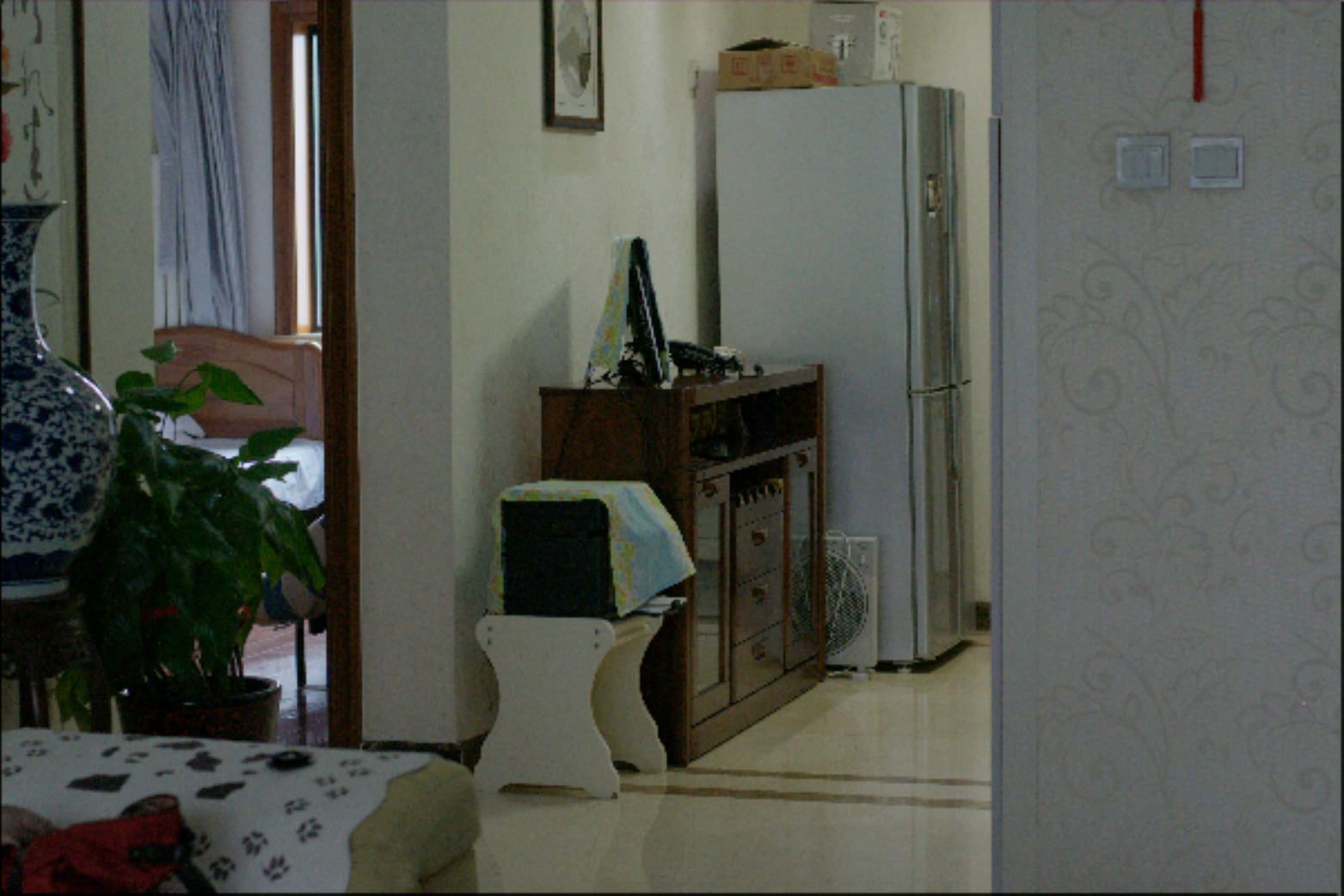}
}
\quad
\subfigure[$t = 4$]{
\includegraphics[width=4.2cm]{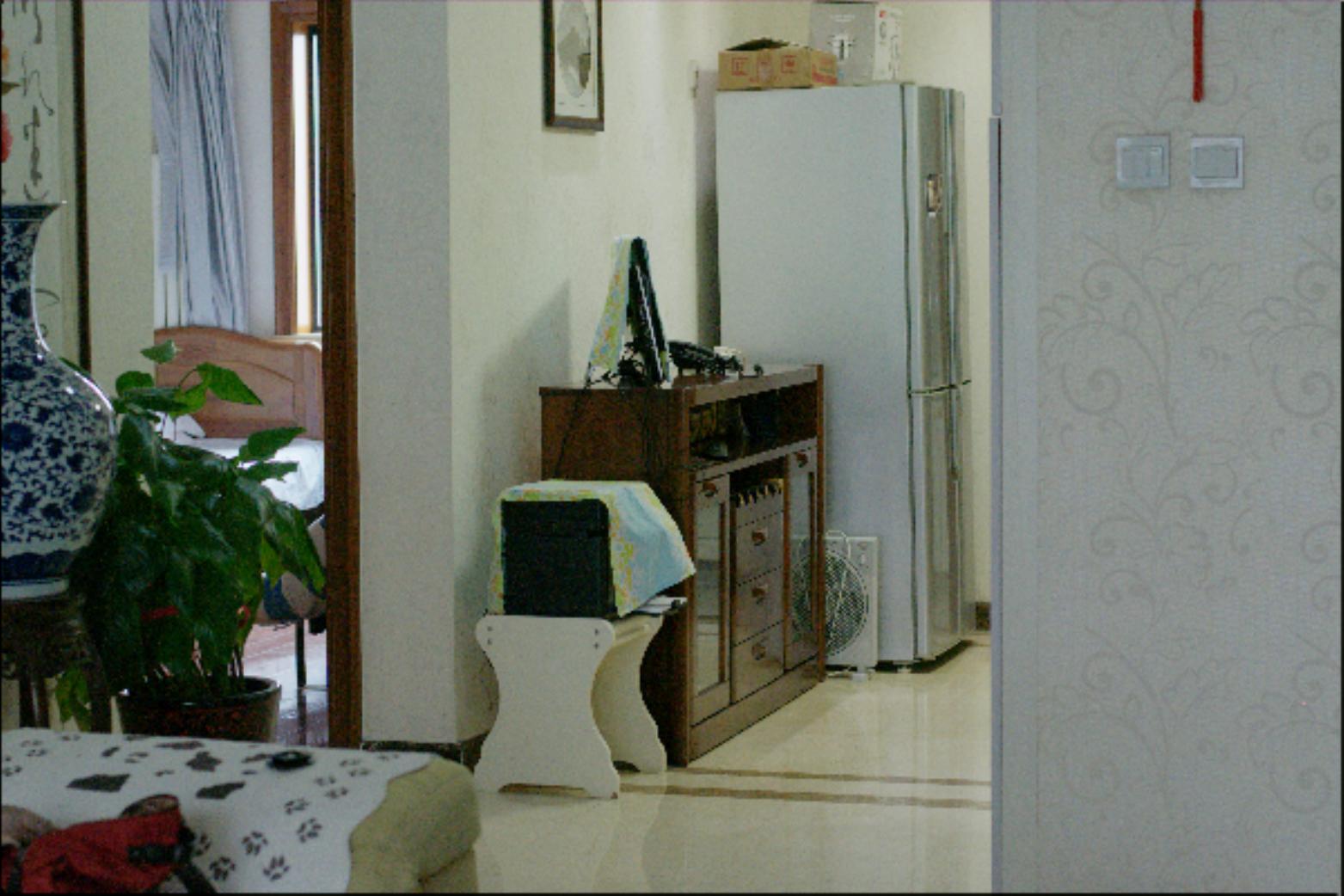}
}
\subfigure[$t = 5$]{
\includegraphics[width=4.2cm]{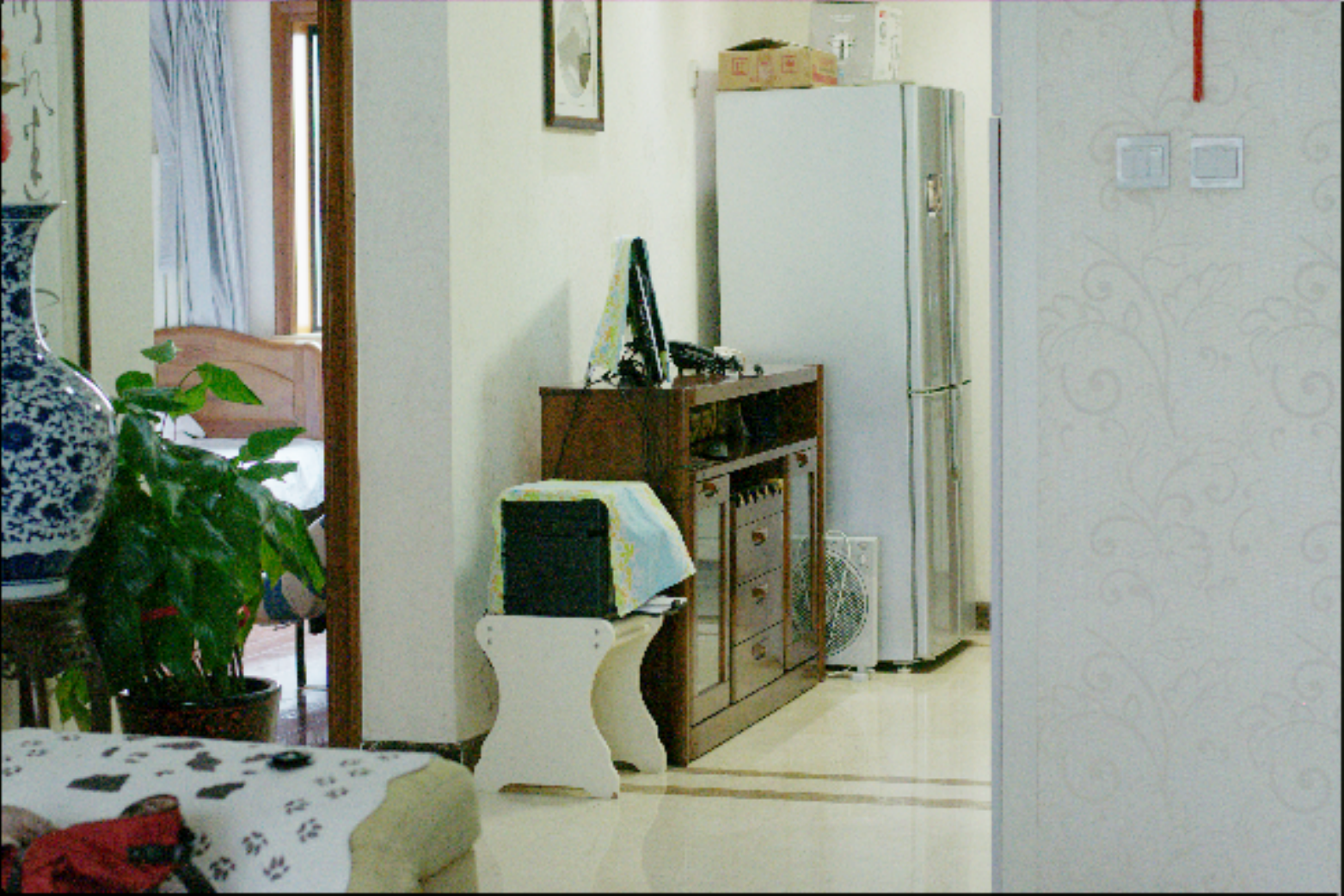}
}
\subfigure[$t = 6$]{
\includegraphics[width=4.2cm]{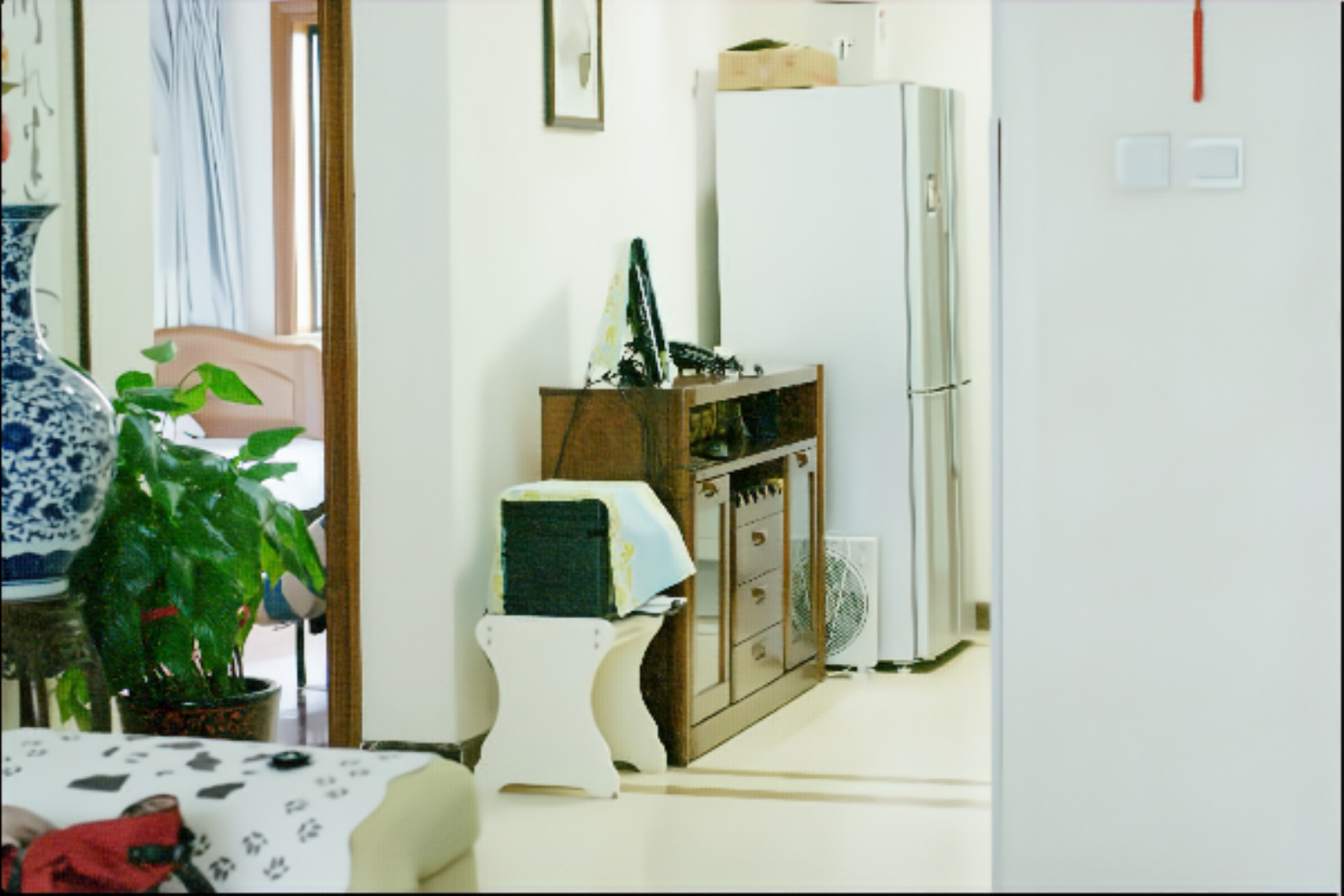}
}
\subfigure[$t = 7$]{
\includegraphics[width=4.2cm]{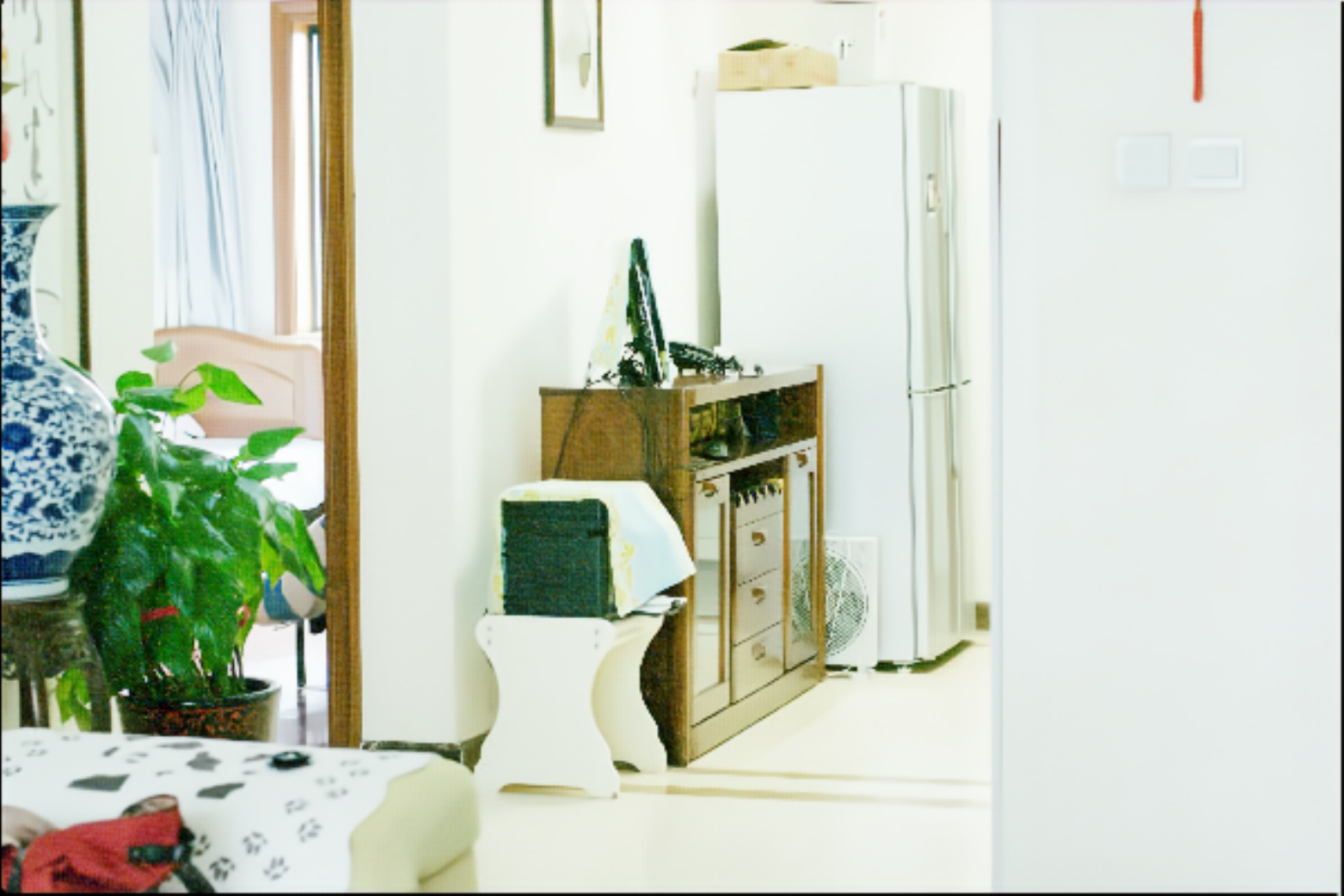}
}
\vspace{-0.2cm}
\caption{An example of the proposed method with different enhancement steps.}
\label{v3}
\vspace{-0.2cm}
\end{figure*}
We present the visual comparisons on typical low-light images in LOL test dataset \cite{Chen2018Retinex} and LIME \cite{guo2016lime} dataset.
We first investigate whether the proposed method achieves visually pleasing results in terms of brightness, color, contrast, and naturalness. We observe from Fig.~\ref{v1} and Fig.~\ref{v2} that the images enhanced by our method are aesthetically acceptable and do not cause any discernible noise and artifacts. Specifically, LIME causes color artifacts at strong local edges; Retinex-Net and EnlightenGAN cause local color distortion and lack of detail; ISSR and RUAS produce severe global and local over/under-exposure; Zero-DCE and SCL-LLE under-enhance extremely dark images, while ReLLIE over-enhances. More qualitative results can be found in the supplementary material.

Fig.~\ref{v3} demonstrates the visualization of the enhancing procedure of the proposed method. As the enhancement steps $t$ progress, the image's brightness increases. In our test experiments, step $t=6$ yields the best visual performance, which is rational as the total number of steps is set to $n=6$ in the training phase. When continuing to enhance the image with an additional step ($t=7$), the image may tend to be over-enhanced, as shown in Fig.~\ref{v3} (h). After many experiments, we found that $t=6$ is the global best in most cases. The optimal enhancement step for a specific image may be different. The image sequences of all steps can be listed for people to choose their preferred one.

\subsubsection{Image Quality Assessment (IQA)}
\begin{figure*}[!htp]
\centering
\subfigure[Input]{
\includegraphics[width=3.3cm]{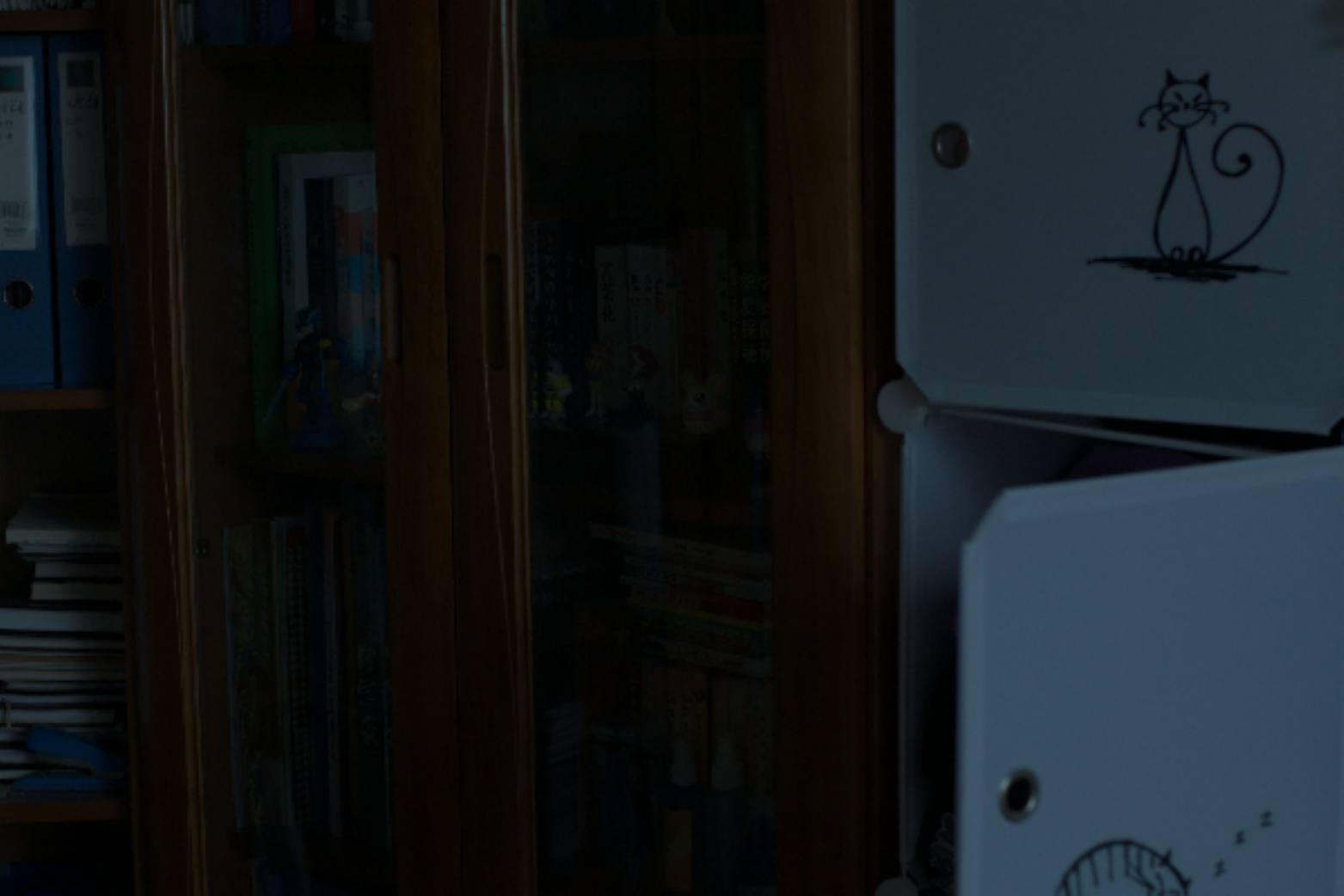}
}
\subfigure[w/o $L_{aes}$]{
\includegraphics[width=3.3cm]{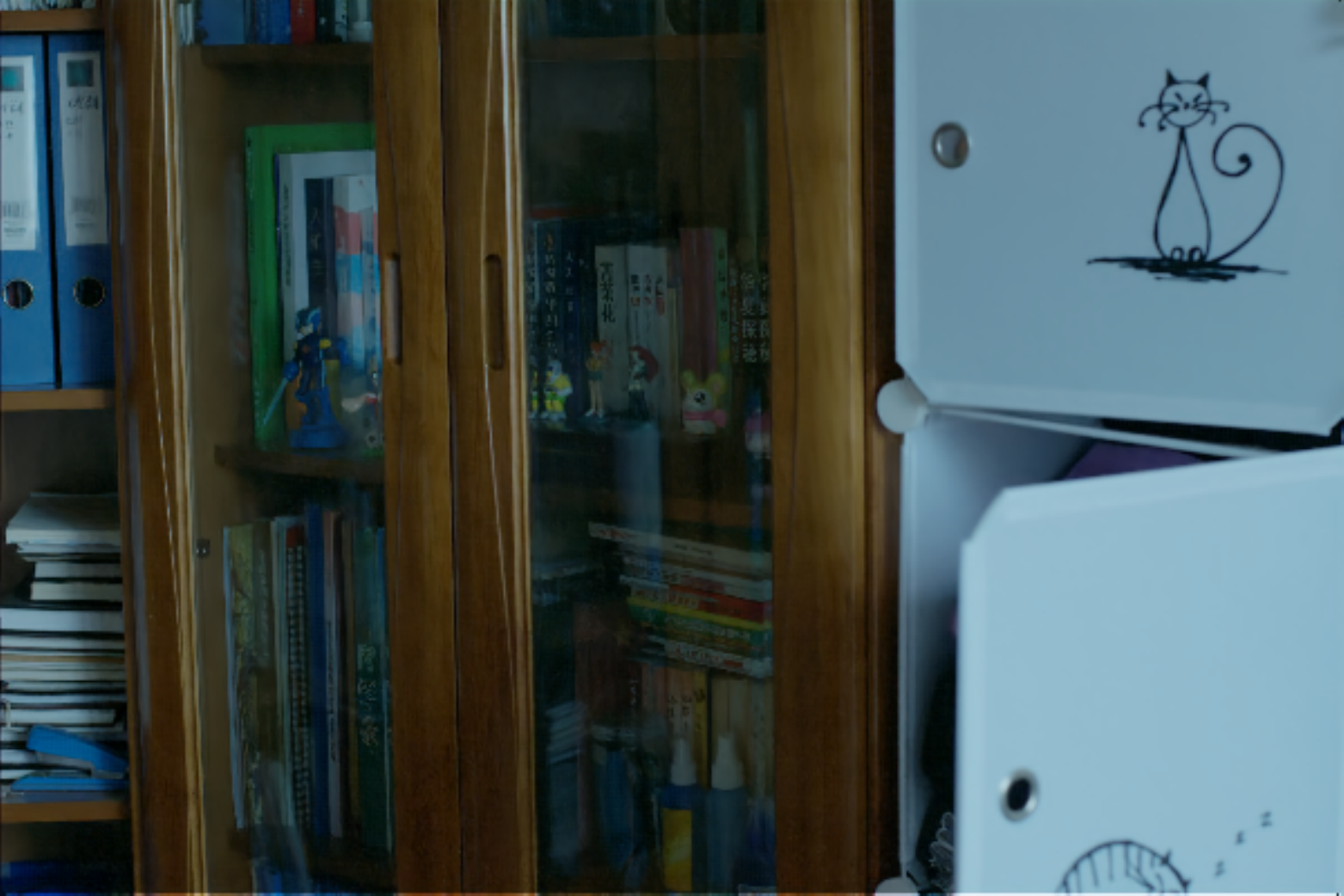}
}
\subfigure[Baseline $A.S.$ settings]{
\includegraphics[width=3.3cm]{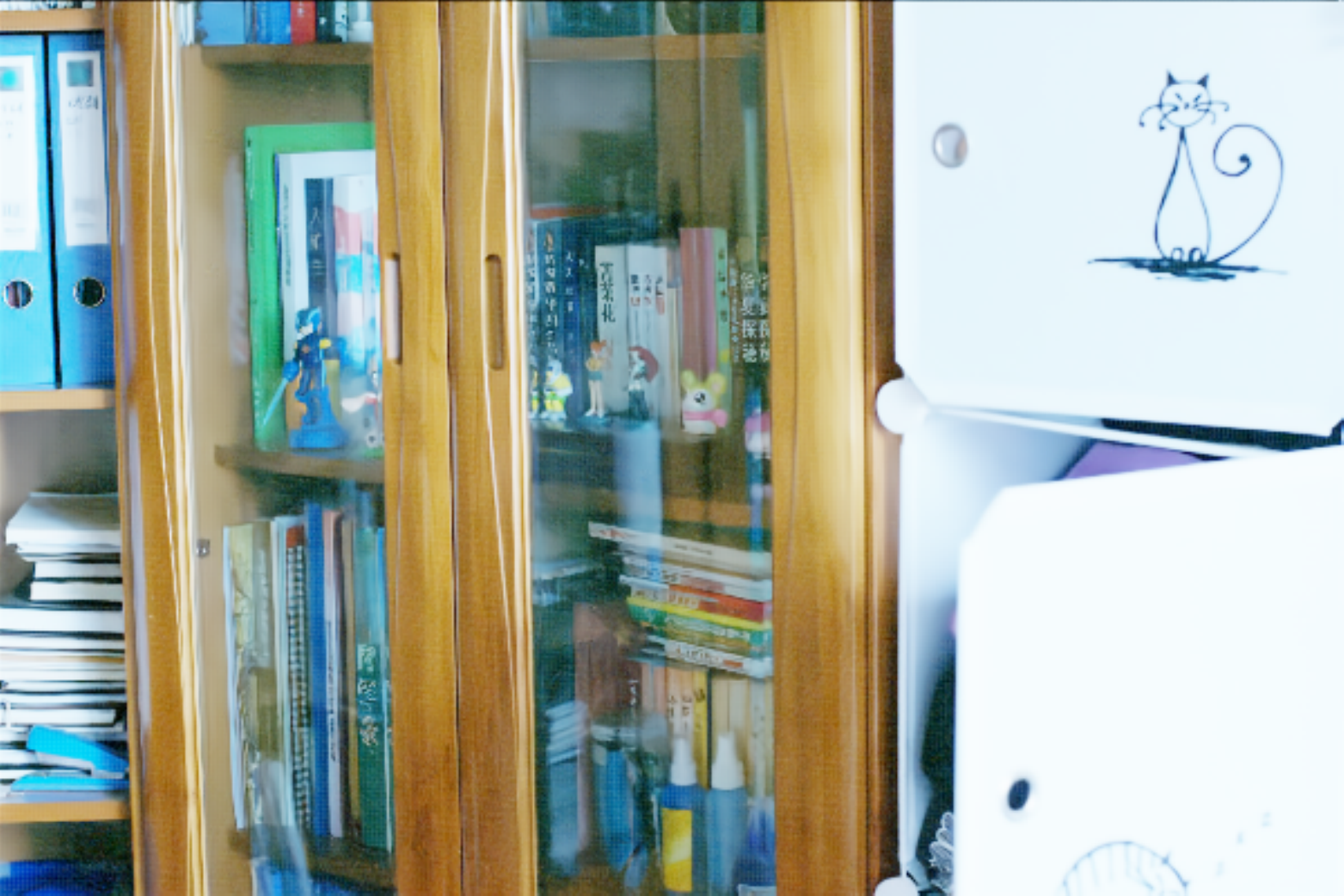}
}
\subfigure[Ours]{
\includegraphics[width=3.3cm]{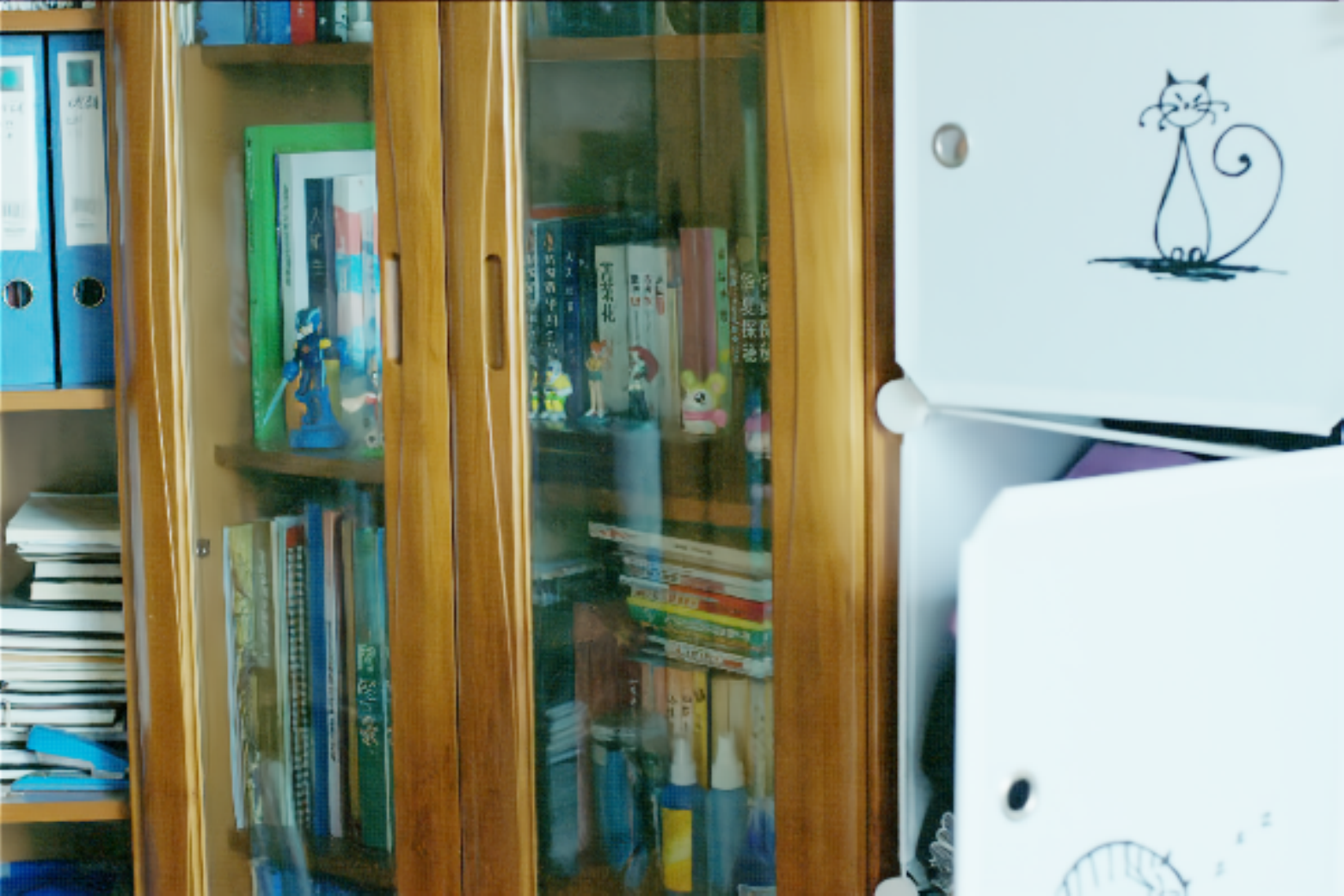}
}
\subfigure[Ground truth]{
\includegraphics[width=3.3cm]{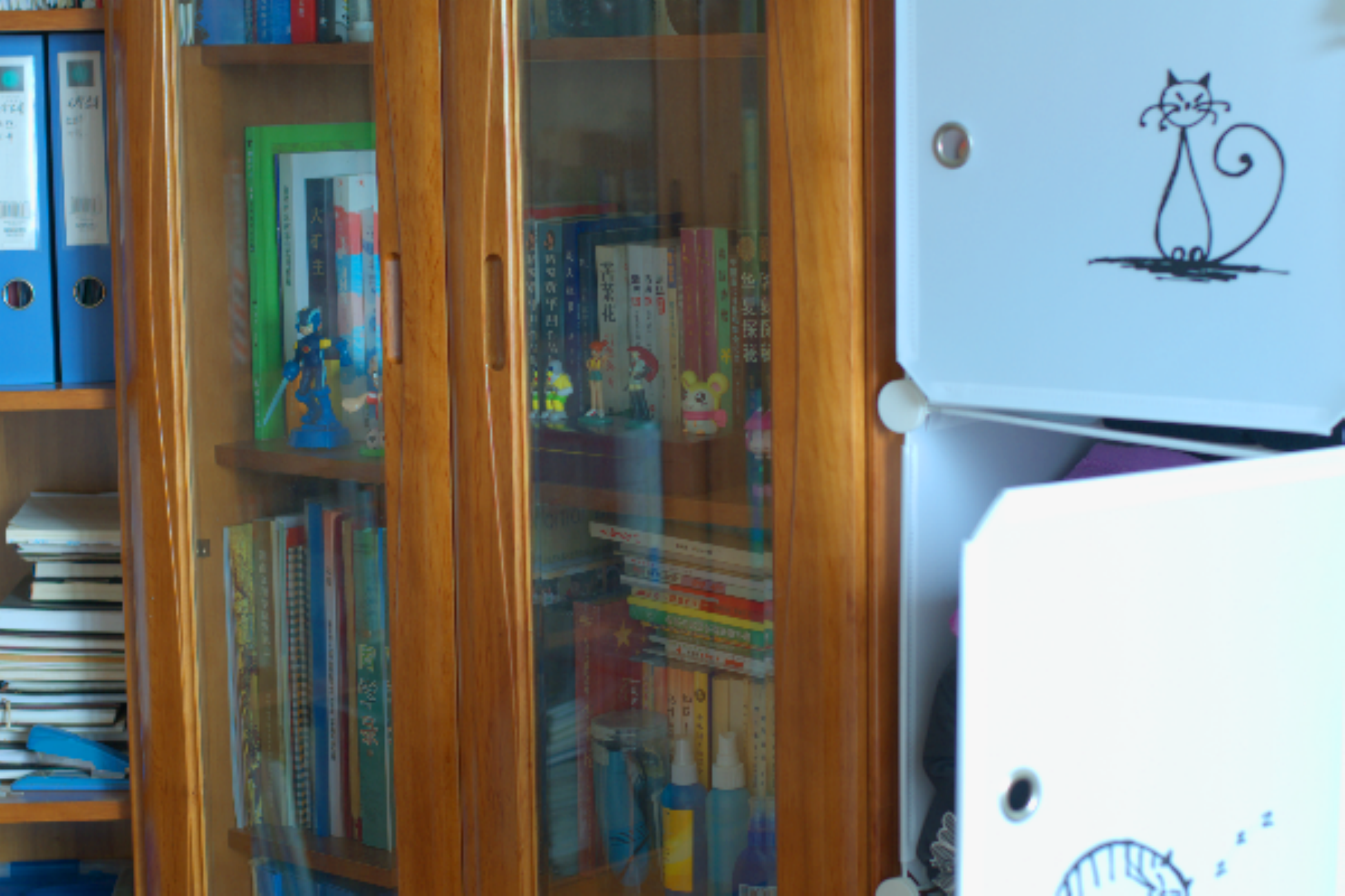}
}
\vspace{-0.2cm}
\caption{Ablation study on the contribution of each component.}
\vspace{-0.2cm}
\label{as1}
\end{figure*}

\begin{figure}[!htp]
\centering
\includegraphics[width=8.5cm]{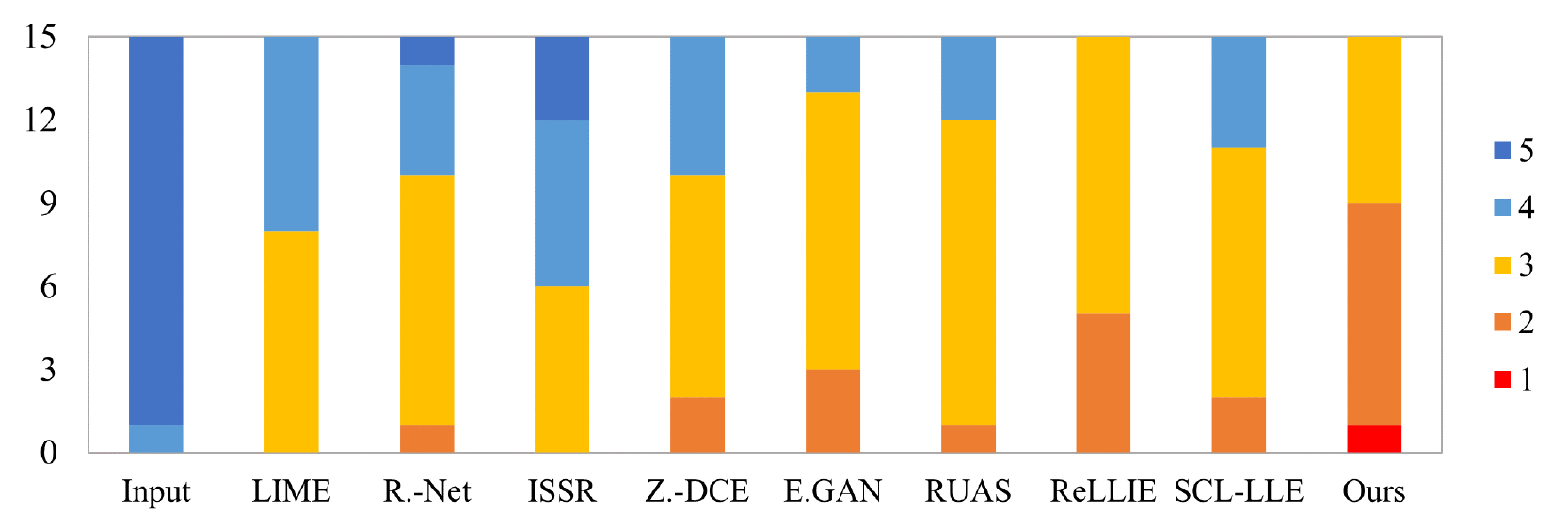}
\vspace{-0.5cm}
\caption{Results of the human subjective survey. The color-changing from warm to cool represents the deterioration of image quality; The y-axis represents the number of images in each rating index.}
\label{US}
\end{figure}

For quantitative comparison with existing methods, we have used two non-reference evaluation indicators, Natural Image Quality Evaluator (NIQE) \cite{Mittal2013MakingA}, and UNIQUE \cite{9369977}.
NIQE is a well-known no-reference image quality assessment for evaluating image restoration without ground truth and providing quantitative comparisons. Since NIQE is regarded as poorly correlated with subjective human opinion, we also adopt UNIQUE, a recently proposed metric for non-reference evaluation that is more rational and closer to subjective human opinion.

For full-reference image quality assessment, we employ the Peak Signal-to-Noise Ratio (PSNR, dB) and Structural Similarity (SSIM) metrics to compare the performance of various approaches quantitatively. 

Table~\ref{tab1} and \ref{tab2} summarizes the performances of our technique and thetechnique state-of-the-art methods on the test images of LOL test dataset \cite{Chen2018Retinex}, DICM \cite{lee2012contrast}, LIME \cite{guo2016lime}, MEF \cite{ma2015perceptual}, VV\footnote{https://sites.google.com/site/vonikakis/datasets}, and NPE \cite{wang2013naturalness}. DICM, LIME, MEF, VV, NPE are ad hoc test datasets, including 64, 10, 8, 24, 17 images, respectively. They are widely used in LLE testing: SCL-LLE \cite{liang2022semantically}, EnlightenGAN \cite{jiang2021enlightengan}, Zero-DCE \cite{Guo_2020_CVPR} \emph{et al.}.
Images in them are diverse and representative:  DICM is mainly landscaped with extreme darkness; LIME focuses on dark street landscapes; MEF focuses on dark indoor scenes and buildings; VV is mostly backlit and portraits; NPE mainly includes natural scenery in low light.
Note that only the PSNR of our method is not ranked first. We believe this is because the aesthetic reward focuses more on the global aesthetic evaluation and is insufficiently responsive to local noise, as detailed in the supplemental material.

\subsubsection{Human Subjective Survey}
We conduct a human subjective survey (user study) for comparisons. Each image in the LOL test dataset was enhanced by nine methods (LIME, Retinex-Net, ISSR, Zero-DCE, EnlightenGAN, RUAS, ReLLIE, SCL-LLE, and our approach), 25 human volunteers were asked to rank the enhanced images. These subjects are instructed to consider: 

1) Whether or not noticeable noise is present in the images; 

2) If the images contain over or underexposure artifacts; 

3) Whether or not the images display non-realistic color or texture distortion. 

We assign each image a score between 1 and 5; the lower the value, the higher the image quality. The final results are shown in Table~\ref{tab1} and Fig.~\ref{US}.
One can see that our method achieves the highest score.

\subsection{Ablation Study}
To demonstrate the effectiveness of the aesthetic rewards and the action space configuration proposed by our technique,  
we performed several ablation experiments.
Since $r^t_{fea}$ and $r^t_{exp}$ are demonstrated to be valid in the recent literature \cite{Guo_2020_CVPR,zhang2021rellie}, we consider them to be the baseline reward of the method and do not conduct ablation studies on them. 

\textbf{Action Space and Aesthetics Quality Reward.} Regarding the the action space ($A.S.$) settings, 
the baseline follows \cite{zhang2021rellie}: the range of $A.S. \in[-0.3, 1]$ with graduation as $0.05$. In comparison, our settings fall within the range of $A.S. \in[-0.5, 1]$ with a graduation of $1/18$.
%Specifically, we design two experiments by removing the components of image aesthetics loss and action space, respectively. 
Specifically, we design two experiments by removing the aesthetics quality reward component and keeping the baseline action space settings.

\textbf{Subjective Experience.} The visualization of the effects of action space and aesthetics quality reward $r^t_{aes}$ are shown in Fig.~\ref{as1}. The absence of $r^t_{aes}$ rendered the image gloomy and unappealing, while improper action space settings led to overexposure in certain portions of the enhanced image. 

\textbf{Objective Evaluation.} Table~\ref{tabas1} shows the NIQE, UNIQUE, PSNR, and SSIM scores under each experiment. 
Note that, from Table~\ref{tabas1}, the absence of $r^t_{aes}$ does not appear to have a significant impact on the overall outcomes; however, the absence of a suitable action space ($A.S.$) setting appears to have a   significant negative impact on performance.   
Since our experimental setup relies on the presence of different settings of the action space and image aesthetics quality reward, the absence of any component would result in subpar results. The aesthetics-guided action space setting can achieve the best results.

Our method can be seen as a compromise between aesthetics and Image Quality Assessment (IQA).
In some specific scenarios, there would be situations where a high aesthetic score with poor IQA.
The Aesthetics Quality Reward $r_{aes}$ considers general aesthetics, while
the Feature Preservation Reward $r_{fea}$ and the Exposure Control Reward $r_{exp}$ in our method ensure
relatively good IQA.

\begin{table}[!htb]
\centering
\small
%\resizebox{2.1\columnwidth}{!}
{
\begin{tabular}{cc|cccc}
\hline
Our $A.S.$ & $r^t_{aes}$ & NIQE $\downarrow$ & UNIQUE $\uparrow$ &PSNR $\uparrow$ &SSIM $\uparrow $  \\ \hline
& &4.822 &1.131  &13.920 &0.697 \\
\checkmark & &4.766 &1.221  &15.211 &0.723\\
 &\checkmark &4.131 &0.912  &14.526 &0.682\\
\hline    
\checkmark &\checkmark  &\textbf{3.774} &\textbf{1.227} &\textbf{18.216} &\textbf{0.763}\\ 
\hline
\end{tabular}}
\caption{Ablation study. NIQE $\downarrow$, UNIQUE $\uparrow$, PSNR $\uparrow$ and SSIM $\uparrow$ scores on LOL test dataset. }
\label{tabas1}

\end{table}

\section{Conclusion}
We have proposed an effective aesthetics-guided reinforcement learning method to solve the LLE problem. ALL-E illustrates how we can leverage aesthetics to balance the subjective and the output. 
Unlike most existing learning-based methods, using the aesthetic policy generation and aesthetic assessment modules, our method treats LLE as a Markov decision process to realize progressive learning. With aesthetic assessment scores as a reward, general human subjective preferences are introduced, which aids in producing aesthetically pleasing effects, \emph{i.e.} it interacts with the environment (aesthetic assessment module) to yield reward $r$ at step $t$ to mimic a human's image retouching process. “Is the image more pretty? Then continue to adjust it ($s^t$); otherwise, adjust the policy ($A^t$)". Experiments demonstrate that our ALL-E is superior to existing methods on six cross-scene datasets. 

\section*{Acknowledgments}
We thank Professor Jie Qin from NUAA and Dr.Tianyu Ding from Microsoft Seattle for their helpful comments. This work is partly supported by the National Natural Science Foundation of China under grant 62272229, and the Natural Science Foundation of Jiangsu Province under grant BK20222012.

\bibliographystyle{named}
\bibliography{ijcai23}

\begin{thebibliography}{}

\bibitem[\protect\citeauthoryear{Buchsbaum}{1980}]{buchsbaum1980spatial}
Gershon Buchsbaum.
\newblock A spatial processor model for object colour perception.
\newblock {\em Journal of the Franklin institute}, 1980.

\bibitem[\protect\citeauthoryear{Cai \bgroup \em et al.\egroup
  }{2018}]{Cai2018deep}
Jianrui Cai, Shuhang Gu, and Lei Zhang.
\newblock Learning a deep single image contrast enhancer from multi-exposure
  images.
\newblock {\em IEEE Transactions on Image Processing}, pages 2049--2062, 2018.

\bibitem[\protect\citeauthoryear{Cho \bgroup \em et al.\egroup
  }{2020}]{cho2020}
Se~Woon Cho, Na~Rae Baek, Ja~Hyung Koo, Muhammad Arsalan, and Kang~Ryoung Park.
\newblock Semantic segmentation with low light images by modified
  cyclegan-based image enhancement.
\newblock {\em IEEE Access}, 8:93561--93585, 2020.

\bibitem[\protect\citeauthoryear{Deng \bgroup \em et al.\egroup
  }{2018}]{deng2018aesthetic}
Yubin Deng, Chen~Change Loy, and Xiaoou Tang.
\newblock Aesthetic-driven image enhancement by adversarial learning.
\newblock In {\em ACM International Conference on Multimedia}, pages 870--878,
  2018.

\bibitem[\protect\citeauthoryear{Deng \bgroup \em et al.\egroup
  }{2020}]{Deng_2020_CVPR}
Jiankang Deng, Jia Guo, Evangelos Ververas, Irene Kotsia, and Stefanos
  Zafeiriou.
\newblock Retinaface: Single-shot multi-level face localisation in the wild.
\newblock In {\em Proceedings of the IEEE/CVF Conference on Computer Vision and
  Pattern Recognition (CVPR)}, June 2020.

\bibitem[\protect\citeauthoryear{Esfandarani and Milanfar}{2018}]{TalebiM18}
Hossein~Talebi Esfandarani and Peyman Milanfar.
\newblock Nima: Neural image assessment.
\newblock {\em IEEE Transactions on Image Processing}, 27(8):3998--4011, 2018.

\bibitem[\protect\citeauthoryear{Fan \bgroup \em et al.\egroup
  }{2020}]{FanWY020}
Minhao Fan, Wenjing Wang, Wenhan Yang, and Jiaying Liu.
\newblock Integrating semantic segmentation and retinex model for low-light
  image enhancement.
\newblock In {\em ACM International Conference on Multimedia, Virtual Event},
  pages 2317--2325, 2020.

\bibitem[\protect\citeauthoryear{Furuta \bgroup \em et al.\egroup
  }{2019}]{furuta2019fully}
Ryosuke Furuta, Naoto Inoue, and Toshihiko Yamasaki.
\newblock Fully convolutional network with multi-step reinforcement learning
  for image processing.
\newblock In {\em Proceedings of the AAAI Conference on Artificial
  Intelligence}, volume~33, pages 3598--3605, 2019.

\bibitem[\protect\citeauthoryear{Guo \bgroup \em et al.\egroup
  }{2016}]{guo2016lime}
Xiaojie Guo, Yu~Li, and Haibin Ling.
\newblock Lime: Low-light image enhancement via illumination map estimation.
\newblock {\em IEEE Transactions on Image Processing}, pages 982--993, 2016.

\bibitem[\protect\citeauthoryear{Guo \bgroup \em et al.\egroup
  }{2020}]{Guo_2020_CVPR}
Chunle Guo, Chongyi Li, Jichang Guo, Chen~Change Loy, Junhui Hou, Sam Kwong,
  and Runmin Cong.
\newblock Zero-reference deep curve estimation for low-light image enhancement.
\newblock In {\em IEEE Conference on Computer Vision and Pattern Recognition},
  pages 1780--1789, 2020.

\bibitem[\protect\citeauthoryear{Hosu \bgroup \em et al.\egroup
  }{2019}]{Pool-3FC}
Vlad Hosu, Bastian Goldlucke, and Dietmar Saupe.
\newblock Effective aesthetics prediction with multi-level spatially pooled
  features.
\newblock In {\em proceedings of the IEEE/CVF conference on computer vision and
  pattern recognition}, pages 9375--9383, 2019.

\bibitem[\protect\citeauthoryear{Jiang \bgroup \em et al.\egroup
  }{2021}]{jiang2021enlightengan}
Yifan Jiang, Xinyu Gong, Ding Liu, Yu~Cheng, Chen Fang, Xiaohui Shen, Jianchao
  Yang, Pan Zhou, and Zhangyang Wang.
\newblock Enlightengan: Deep light enhancement without paired supervision.
\newblock {\em IEEE Transactions on Image Processing}, pages 2340--2349, 2021.

\bibitem[\protect\citeauthoryear{Jobson \bgroup \em et al.\egroup
  }{1997}]{Jobson1997A}
Jobson, Daniel, J., Rahman, Zia-ur, Woodell, Glenn, and A.
\newblock A multiscale retinex for bridging the gap between color images and
  the human observation of scenes.
\newblock {\em IEEE Transactions on Image Processing}, 1997.

\bibitem[\protect\citeauthoryear{Kang \bgroup \em et al.\egroup
  }{2020}]{kang2020eva}
Chen Kang, Giuseppe Valenzise, and Fr{\'e}d{\'e}ric Dufaux.
\newblock Eva: An explainable visual aesthetics dataset.
\newblock In {\em Joint Workshop on Aesthetic and Technical Quality Assessment
  of Multimedia and Media Analytics for Societal Trends}, pages 5--13, 2020.

\bibitem[\protect\citeauthoryear{Kao \bgroup \em et al.\egroup }{2017}]{MTCNN}
Yueying Kao, Ran He, and Kaiqi Huang.
\newblock Deep aesthetic quality assessment with semantic information.
\newblock {\em IEEE Transactions on Image Processing}, 26(3):1482--1495, 2017.

\bibitem[\protect\citeauthoryear{Kong \bgroup \em et al.\egroup
  }{2016}]{kong2016photo}
Shu Kong, Xiaohui Shen, Zhe Lin, Radomir Mech, and Charless Fowlkes.
\newblock Photo aesthetics ranking network with attributes and content
  adaptation.
\newblock In {\em European Conference on Computer Vision}, pages 662--679.
  Springer, 2016.

\bibitem[\protect\citeauthoryear{Land}{1977}]{land1977retinex}
Edwin~H Land.
\newblock The retinex theory of color vision.
\newblock {\em Scientific american}, pages 108--129, 1977.

\bibitem[\protect\citeauthoryear{Lee \bgroup \em et al.\egroup
  }{2012}]{lee2012contrast}
Chulwoo Lee, Chul Lee, and Chang-Su Kim.
\newblock Contrast enhancement based on layered difference representation.
\newblock In {\em IEEE International Conference on Image Processing}, pages
  965--968, 2012.

\bibitem[\protect\citeauthoryear{Li \bgroup \em et al.\egroup }{2020}]{PA_IAA}
Leida Li, Hancheng Zhu, Sicheng Zhao, Guiguang Ding, and Weisi Lin.
\newblock Personality-assisted multi-task learning for generic and personalized
  image aesthetics assessment.
\newblock {\em IEEE Transactions on Image Processing}, page 3898–3910, Feb
  2020.

\bibitem[\protect\citeauthoryear{Liang \bgroup \em et al.\egroup
  }{2022}]{liang2022semantically}
Dong Liang, Ling Li, Mingqiang Wei, Shuo Yang, Liyan Zhang, Wenhan Yang, Yun
  Du, and Huiyu Zhou.
\newblock Semantically contrastive learning for low-light image enhancement.
\newblock In {\em Proceedings of the AAAI Conference on Artificial
  Intelligence}, volume~36, pages 1555--1563, 2022.

\bibitem[\protect\citeauthoryear{Liu \bgroup \em et al.\egroup
  }{2021}]{liu2021retinex}
Risheng Liu, Long Ma, Jiaao Zhang, Xin Fan, and Zhongxuan Luo.
\newblock Retinex-inspired unrolling with cooperative prior architecture search
  for low-light image enhancement.
\newblock In {\em IEEE Conference on Computer Vision and Pattern Recognition},
  pages 10561--10570, 2021.

\bibitem[\protect\citeauthoryear{Lore \bgroup \em et al.\egroup
  }{2017}]{lore2017llnet}
Kin~Gwn Lore, Adedotun Akintayo, and Soumik Sarkar.
\newblock Llnet: A deep autoencoder approach to natural low-light image
  enhancement.
\newblock {\em Pattern Recognition}, pages 650--662, 2017.

\bibitem[\protect\citeauthoryear{Lu \bgroup \em et al.\egroup
  }{2014}]{lu2014rapid}
Xin Lu, Zhe Lin, Hailin Jin, Jianchao Yang, and James~Z Wang.
\newblock Rapid: Rating pictorial aesthetics using deep learning.
\newblock In {\em ACM International Conference on Multimedia}, pages 457--466,
  2014.

\bibitem[\protect\citeauthoryear{Lu \bgroup \em et al.\egroup }{2015}]{DMA-Net}
Xin Lu, Zhe Lin, Xiaohui Shen, Radomir Mech, and James~Z Wang.
\newblock Deep multi-patch aggregation network for image style, aesthetics, and
  quality estimation.
\newblock In {\em Proceedings of the IEEE international conference on computer
  vision}, pages 990--998, 2015.

\bibitem[\protect\citeauthoryear{Ma \bgroup \em et al.\egroup
  }{2015}]{ma2015perceptual}
Kede Ma, Kai Zeng, and Zhou Wang.
\newblock Perceptual quality assessment for multi-exposure image fusion.
\newblock {\em IEEE Transactions on Image Processing}, pages 3345--3356, 2015.

\bibitem[\protect\citeauthoryear{Ma \bgroup \em et al.\egroup
  }{2017}]{2017Lamp}
Shuang Ma, Jing Liu, and Chang-Wen Chen.
\newblock A-lamp: Adaptive layout-aware multi-patch deep convolutional neural
  network for photo aesthetic assessment.
\newblock In {\em IEEE Conference on Computer Vision and Pattern Recognition},
  pages 722--731, 07 2017.

\bibitem[\protect\citeauthoryear{Ma \bgroup \em et al.\egroup
  }{2022}]{ma2022toward}
Long Ma, Tengyu Ma, Risheng Liu, Xin Fan, and Zhongxuan Luo.
\newblock Toward fast, flexible, and robust low-light image enhancement.
\newblock In {\em Proceedings of the IEEE/CVF Conference on Computer Vision and
  Pattern Recognition}, pages 5637--5646, 2022.

\bibitem[\protect\citeauthoryear{Mittal \bgroup \em et al.\egroup
  }{2013}]{Mittal2013MakingA}
Anish Mittal, R.~Soundararajan, and A.~Bovik.
\newblock Making a “completely blind” image quality analyzer.
\newblock {\em IEEE Signal Processing Letters}, pages 209--212, 2013.

\bibitem[\protect\citeauthoryear{Mnih \bgroup \em et al.\egroup
  }{2016}]{mnih2016asynchronous}
Volodymyr Mnih, Adria~Puigdomenech Badia, Mehdi Mirza, Alex Graves, Timothy
  Lillicrap, Tim Harley, David Silver, and Koray Kavukcuoglu.
\newblock Asynchronous methods for deep reinforcement learning.
\newblock In {\em International Conference on Machine Learning}, pages
  1928--1937. PMLR, 2016.

\bibitem[\protect\citeauthoryear{Murray \bgroup \em et al.\egroup
  }{2012}]{murray2012ava}
Naila Murray, Luca Marchesotti, and Florent Perronnin.
\newblock Ava: A large-scale database for aesthetic visual analysis.
\newblock In {\em IEEE Conference on Computer Vision and Pattern Recognition},
  pages 2408--2415. IEEE, 2012.

\bibitem[\protect\citeauthoryear{Park \bgroup \em et al.\egroup
  }{2018}]{park2018distort}
Jongchan Park, Joon-Young Lee, Donggeun Yoo, and In~So Kweon.
\newblock Distort-and-recover: Color enhancement using deep reinforcement
  learning.
\newblock In {\em IEEE Conference on Computer Vision and Pattern Recognition},
  pages 5928--5936, 2018.

\bibitem[\protect\citeauthoryear{Pizer \bgroup \em et al.\egroup
  }{1990}]{Pizer1990ContrastlimitedAH}
S.~Pizer, R.~Johnston, J.~P. Ericksen, B.~Yankaskas, and K.~Muller.
\newblock Contrast-limited adaptive histogram equalization: speed and
  effectiveness.
\newblock {\em Conference on Visualization in Biomedical Computing}, pages
  337--345, 1990.

\bibitem[\protect\citeauthoryear{Ren \bgroup \em et al.\egroup }{2017}]{PAM}
Jian Ren, Xiaohui Shen, Zhe Lin, Radomir Mech, and David~J. Foran.
\newblock Personalized image aesthetics.
\newblock In {\em 2017 IEEE International Conference on Computer Vision
  (ICCV)}, Dec 2017.

\bibitem[\protect\citeauthoryear{Ren \bgroup \em et al.\egroup
  }{2019}]{8692732}
Wenqi Ren, Sifei Liu, Lin Ma, Qianqian Xu, Xiangyu Xu, Xiaochun Cao, Junping
  Du, and Ming-Hsuan Yang.
\newblock Low-light image enhancement via a deep hybrid network.
\newblock {\em IEEE Transactions on Image Processing}, pages 4364--4375, 2019.

\bibitem[\protect\citeauthoryear{Wang \bgroup \em et al.\egroup
  }{2013}]{wang2013naturalness}
Shuhang Wang, Jin Zheng, Hai-Miao Hu, and Bo~Li.
\newblock Naturalness preserved enhancement algorithm for non-uniform
  illumination images.
\newblock {\em IEEE Transactions on Image Processing}, pages 3538--3548, 2013.

\bibitem[\protect\citeauthoryear{Wang \bgroup \em et al.\egroup }{2018}]{CFAN}
Guolong Wang, Junchi Yan, and Zheng Qin.
\newblock Collaborative and attentive learning for personalized image aesthetic
  assessment.
\newblock In {\em Proceedings of the Twenty-Seventh International Joint
  Conference on Artificial Intelligence}, Jul 2018.

\bibitem[\protect\citeauthoryear{Wei \bgroup \em et al.\egroup
  }{2018}]{Chen2018Retinex}
Chen Wei, Wenjing Wang, Wenhan Yang, and Jiaying Liu.
\newblock Deep retinex decomposition for low-light enhancement.
\newblock In {\em British Machine Vision Conference}, 2018.

\bibitem[\protect\citeauthoryear{Wu \bgroup \em et al.\egroup
  }{2022}]{wu2022uretinex}
Wenhui Wu, Jian Weng, Pingping Zhang, Xu~Wang, Wenhan Yang, and Jianmin Jiang.
\newblock Uretinex-net: Retinex-based deep unfolding network for low-light
  image enhancement.
\newblock In {\em Proceedings of the IEEE/CVF Conference on Computer Vision and
  Pattern Recognition}, pages 5901--5910, 2022.

\bibitem[\protect\citeauthoryear{Xu \bgroup \em et al.\egroup
  }{2014}]{xu2014novel}
Qing Xu, Hailin Jiang, Riccardo Scopigno, and Mateu Sbert.
\newblock A novel approach for enhancing very dark image sequences.
\newblock {\em Signal processing}, pages 309--330, 2014.

\bibitem[\protect\citeauthoryear{Xu \bgroup \em et al.\egroup
  }{2020}]{Xu_2020_CVPR}
Ke~Xu, Xin Yang, Baocai Yin, and Rynson~W.H. Lau.
\newblock Learning to restore low-light images via
  decomposition-and-enhancement.
\newblock In {\em IEEE Conference on Computer Vision and Pattern Recognition},
  2020.

\bibitem[\protect\citeauthoryear{Yang \bgroup \em et al.\egroup
  }{2016}]{Yang_2016_CVPR}
Shuo Yang, Ping Luo, Chen-Change Loy, and Xiaoou Tang.
\newblock Wider face: A face detection benchmark.
\newblock In {\em Proceedings of the IEEE Conference on Computer Vision and
  Pattern Recognition (CVPR)}, June 2016.

\bibitem[\protect\citeauthoryear{Yang \bgroup \em et al.\egroup }{2022}]{PIAA}
Yuzhe Yang, Liwu Xu, Leida Li, Nan Qie, Yaqian Li, Peng Zhang, and Yandong Guo.
\newblock Personalized image aesthetics assessment with rich attributes.
\newblock In {\em IEEE Conference on Computer Vision and Pattern Recognition},
  pages 19861--19869, 2022.

\bibitem[\protect\citeauthoryear{Yu and Moon}{2004}]{yu2004exploratory}
So-Young Yu and Sung-Been Moon.
\newblock An exploratory study of image retrieval using aesthetic impressions.
\newblock {\em Journal of the Korean Society for information Management},
  21(4):187--208, 2004.

\bibitem[\protect\citeauthoryear{Yu \bgroup \em et al.\egroup
  }{2018}]{yu2018crafting}
Ke~Yu, Chao Dong, Liang Lin, and Chen~Change Loy.
\newblock Crafting a toolchain for image restoration by deep reinforcement
  learning.
\newblock In {\em IEEE Conference on Computer Vision and Pattern Recognition},
  pages 2443--2452, 2018.

\bibitem[\protect\citeauthoryear{Yu \bgroup \em et al.\egroup
  }{2021a}]{yu2021novel}
Jing Yu, Lulu Zhao, et~al.
\newblock A novel deep cnn method based on aesthetic rule for user preferential
  images recommendation.
\newblock {\em Journal of Applied Science and Engineering}, 24(1):49--55, 2021.

\bibitem[\protect\citeauthoryear{Yu \bgroup \em et al.\egroup
  }{2021b}]{Yu_2021_ICCV}
Jun Yu, Xinlong Hao, and Peng He.
\newblock Single-stage face detection under extremely low-light conditions.
\newblock In {\em Proceedings of the IEEE/CVF International Conference on
  Computer Vision (ICCV) Workshops}, pages 3523--3532, October 2021.

\bibitem[\protect\citeauthoryear{Yuan \bgroup \em et al.\egroup
  }{2019}]{yuan2019ug}
Ye~Yuan, Wenhan Yang, Wenqi Ren, Jiaying Liu, Walter~J Scheirer, and Zhangyang
  Wang.
\newblock Ug2+ track 2: A collective benchmark effort for evaluating and
  advancing image understanding in poor visibility environments.
\newblock {\em arXiv preprint arXiv:1904.04474}, 2019.

\bibitem[\protect\citeauthoryear{Zhang \bgroup \em et al.\egroup
  }{2019}]{zhang2019kindling}
Yonghua Zhang, Jiawan Zhang, and Xiaojie Guo.
\newblock Kindling the darkness: A practical low-light image enhancer.
\newblock In {\em ACM International Conference on Multimedia}, page
  1632–1640, 2019.

\bibitem[\protect\citeauthoryear{Zhang \bgroup \em et al.\egroup
  }{2021a}]{zhang2021rellie}
Rongkai Zhang, Lanqing Guo, Siyu Huang, and Bihan Wen.
\newblock Rellie: Deep reinforcement learning for customized low-light image
  enhancement.
\newblock In {\em ACM International Conference on Multimedia}, pages
  2429--2437, 2021.

\bibitem[\protect\citeauthoryear{Zhang \bgroup \em et al.\egroup
  }{2021b}]{9369977}
Weixia Zhang, Kede Ma, Guangtao Zhai, and Xiaokang Yang.
\newblock Uncertainty-aware blind image quality assessment in the laboratory
  and wild.
\newblock {\em IEEE Transactions on Image Processing}, pages 3474--3486, 2021.

\bibitem[\protect\citeauthoryear{Zhao \bgroup \em et al.\egroup }{2020}]{ReLIC}
Lin Zhao, Meimei Shang, Fei Gao, Rongsheng Li, Fei Huang, and Jun Yu.
\newblock Representation learning of image composition for aesthetic
  prediction.
\newblock {\em Computer Vision and Image Understanding}, 199:103024, 2020.

\bibitem[\protect\citeauthoryear{Zhao \bgroup \em et al.\egroup
  }{2021}]{zhao2021deep}
Lin Zhao, Shao-Ping Lu, Tao Chen, Zhenglu Yang, and Ariel Shamir.
\newblock Deep symmetric network for underexposed image enhancement with
  recurrent attentional learning.
\newblock In {\em Proceedings of the IEEE/CVF international conference on
  computer vision}, pages 12075--12084, 2021.

\bibitem[\protect\citeauthoryear{Zhu \bgroup \em et al.\egroup
  }{2020}]{BLG-PIAA}
Hancheng Zhu, Leida Li, Jinjian Wu, Sicheng Zhao, Guiguang Ding, and Guangming
  Shi.
\newblock Personalized image aesthetics assessment via meta-learning with
  bilevel gradient optimization.
\newblock {\em IEEE Transactions on Cybernetics}, 52(3):1798–1811, Jun 2020.

\end{thebibliography}

%\newpage
\appendix
\section{Actor-Critic Reinforcement Learning and A3C Algorithm}
%In this section, we will introduce the Actor-Critic algorithm, since the A3C~\cite{mnih2016asynchronous} algorithm is based on the Actor-Critic algorithm. 
There are two main types of Reinforcement Learning methods out there:\\
(1) Value-Based: 

Value-Based RL algorithms try to find or approximate the optimal value function, which is a mapping between an action and a value. The higher the value, the better the action. The most famous algorithm is Q learning and all its enhancements like Deep Q Networks, Double Dueling Q Networks, etc.\\
(2) Policy-Based: 

Policy-Based RL algorithms like Policy Gradients and REINFORCE try to find the optimal policy directly without the Q-value as a middleman.

Each method has its advantages. For example, policy-based methods are better for continuous and stochastic environments, with faster convergence, while value-based methods are more efficient and steady. 
The actor-critic algorithm is proposed by integrating the advantages of these two approaches. The architecture of the Actor-Critic consists of two parts:
\begin{itemize}
    \item [$\bullet$] Actor is a policy function $\pi _{\theta }(s)$ where the input is the current state, and the output is an action. The training goal of this network is to maximize the expectation of cumulative returns. 
    \item [$\bullet$] Critic is the value function $V^{\pi}(s)$, which is a network that can estimate the value function of the current strategy; that is, it can evaluate the goodness of the Actor ($i.e.$ strategy function).
\end{itemize}

The asynchronous advantage actor-critic (A3C) algorithm uses the Actor-Critic framework and introduces the idea of asynchronous training, which speeds up training while improving performance. This process is in Algorithm 1.

\begin{algorithm}[!h]
	%\textsl{}\setstretch{1.8}
	\renewcommand{\algorithmicrequire}{\textbf{Input:}}
	\renewcommand{\algorithmicensure}{\textbf{Output:}}
	\caption{Asynchronous advantage actor-critic - pseudocode for each actor-learner thread.}
	\label{alg1}
	\begin{algorithmic}[1]
	    \STATE \textit{// Assume global shared parameter vectors $\theta$ and $\theta _{v}$ and global shared counter T = 0}
	    \STATE \textit{// Assume thread-specific parameter vectors $\theta'$ and $\theta'_{v}$}
	    \STATE Initialize thread step counter $t\leftarrow 1$
	    \REPEAT
		\STATE Reset gradients: $d\theta \leftarrow 0$ and $d\theta_{v} \leftarrow 0$
		\STATE Synchronize thread-specific parameters $\theta' = \theta$ and $\theta'_{v} = \theta_{v}$
		\STATE $t^{start} = t$
		\STATE Get state $s^t$
		\REPEAT
		\STATE Perform $A^t$ according to policy $\pi(A^t\mid s^t;\theta') $
		\STATE Receive reward $r^t$ and new state $s^{t+1}$
		\STATE $t \leftarrow t + 1$
		\STATE $T \leftarrow T + 1$
		\UNTIL terminal $s^{t}$ \OR $t - t_{start} == t_{max}$
		\STATE $R = \left\{\begin{matrix}
0 & $for terminal $s^t\\ 
V(s^t, \theta'_v) &$for non-terminal $s^t
\end{matrix}\right.$
        \FOR{$i \in \{t-1, \dots, t^{start} \}$}
        \STATE $R \leftarrow r^i + \gamma R$
        \STATE Accumulate gradients wrt $\theta'$:$d\theta \leftarrow d\theta + \triangledown _{\theta'}\log \pi(A^i\mid s^i;\theta')(R-V(s^i;\theta'_v)) $ \STATE Accumulate gradients wrt $\theta'_v$: $d\theta _v \leftarrow  d\theta _v+\partial (R-V(s^i;\theta '_v))^2/\partial \theta '_v$
        \ENDFOR
        \STATE Perform asynchronous update of $\theta$ using $d\theta$ and of $\theta_v$ using $d\theta _v$
		\UNTIL $T > T_{max}$
	\end{algorithmic}  
\end{algorithm}

The A3C algorithm uses the idea of asynchronous training in order to improve the training speed using multiple threads. Each thread is equivalent to one agent exploring randomly and multiple agents exploring together, computing the policy gradient in parallel and updating the parameters.

\section{Effect of Action Space}
Regarding the settings of the action space,
the baseline uses the settings in \cite{zhang2021rellie}: the range $A.S. \in[-0.3, 1]$ with graduation as $0.05$. Compared to this setting, our settings are the range $A.S. \in[-0.5, 1]$ with graduation as $1/18$.

\begin{figure}[!htb]
\centering
\includegraphics[width=9.5cm]{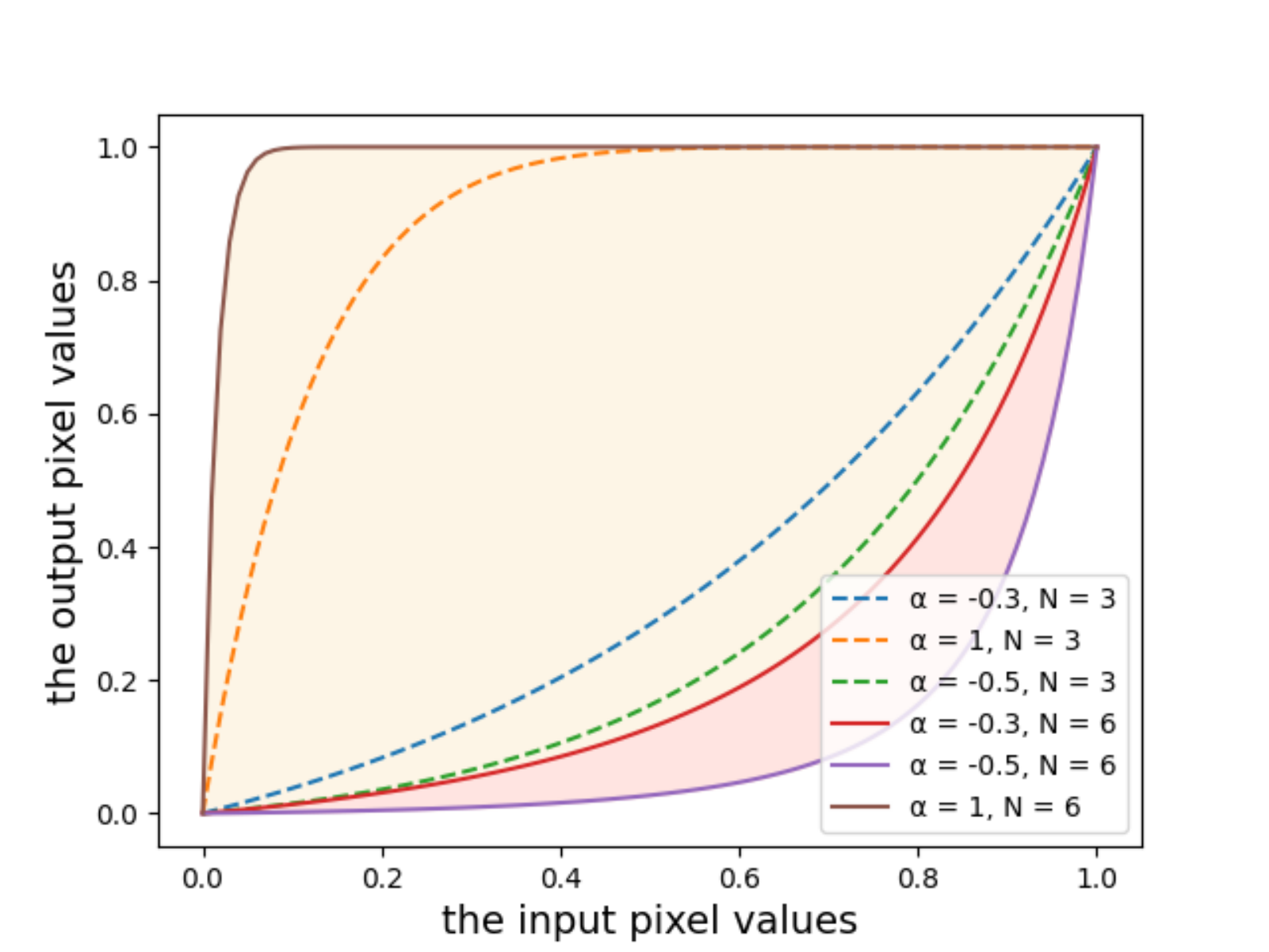}

\caption{Illustration of how the adjustment ranges with different $N$. $N$ denotes the number of steps.
}

\label{AS}
\end{figure}

Figure~\ref{AS} and Table~\ref{LEC} show that the pixel-wise adjustment curve PAC can effectively cover the pixel value space under the proposed action space settings, with a more extensive adjustment range than ReLLIE \cite{zhang2021rellie}. The mapping space covered by multiple enhancement steps ($N$ = 6) is much larger than that of a single step ($N$ = 1), allowing ALL-E processing fine-grained brightness. With the action space setting of our method, the covered pixel space is a concatenation of the yellow and pink areas.

\begin{figure*}[t]
\centering
\subfigure[Baseline (ACMMM’21)]{
\includegraphics[width=5.65cm]{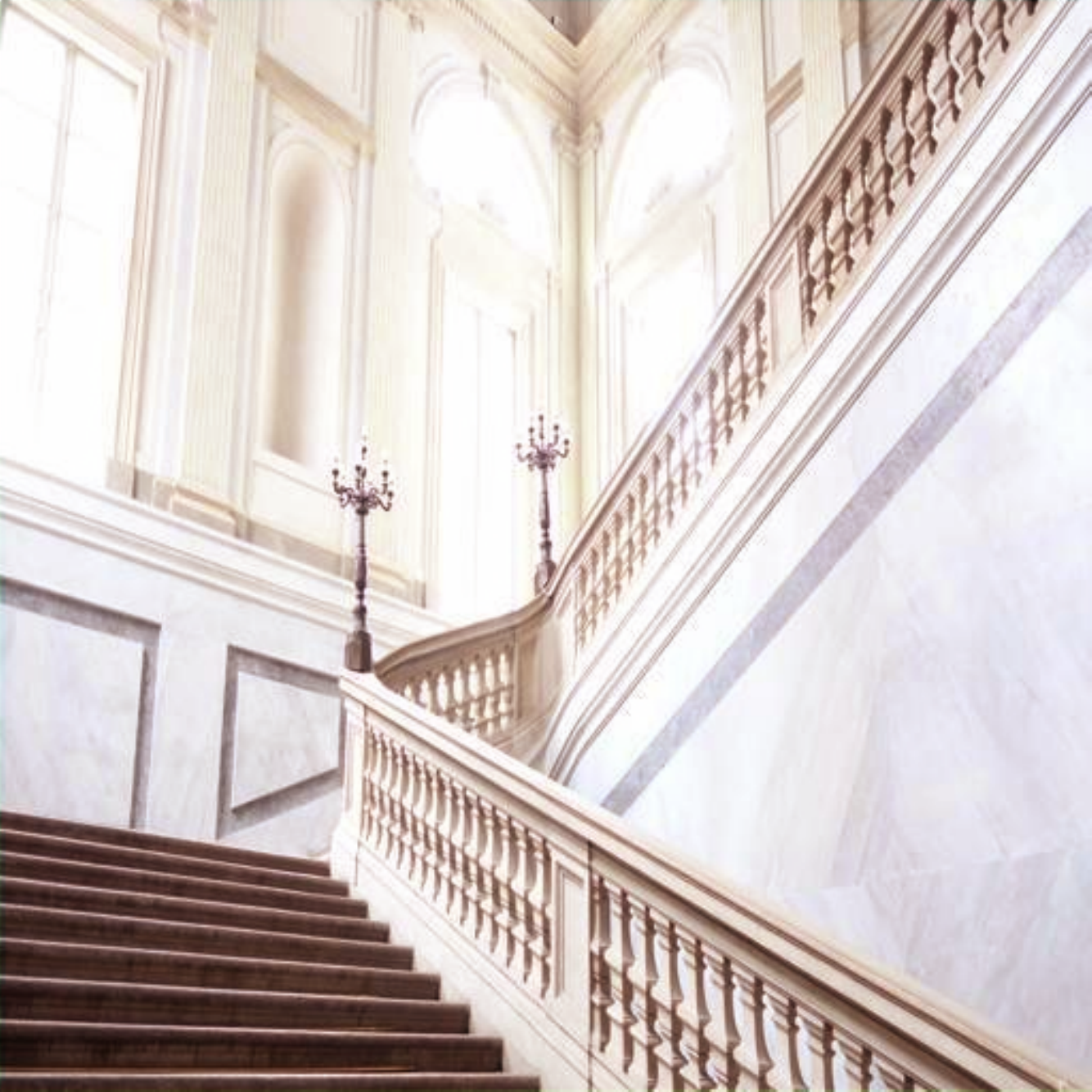}} 
\subfigure[Ours]{
\includegraphics[width=5.65cm]{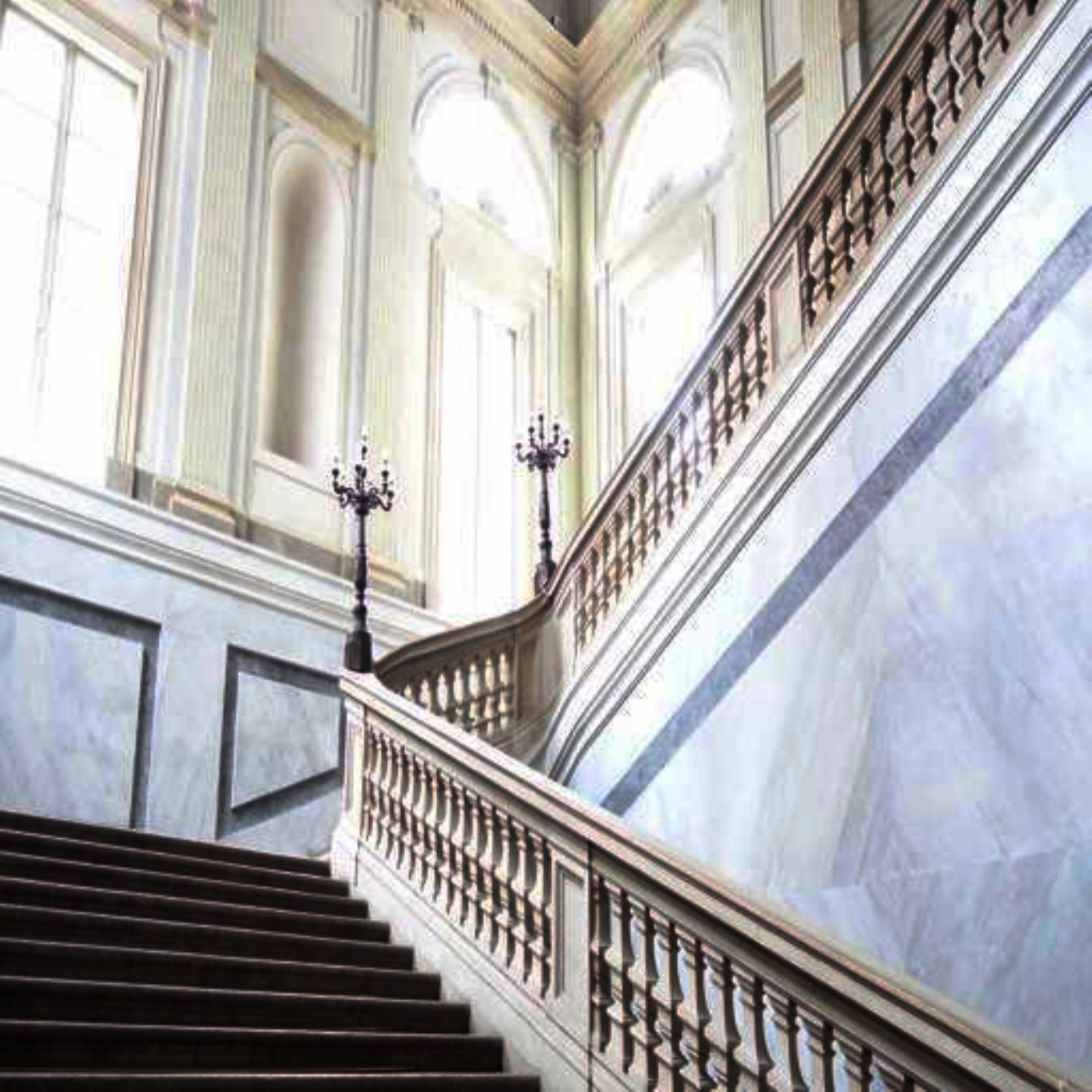}}
\subfigure[Ground Truth]{
\includegraphics[width=5.65cm]{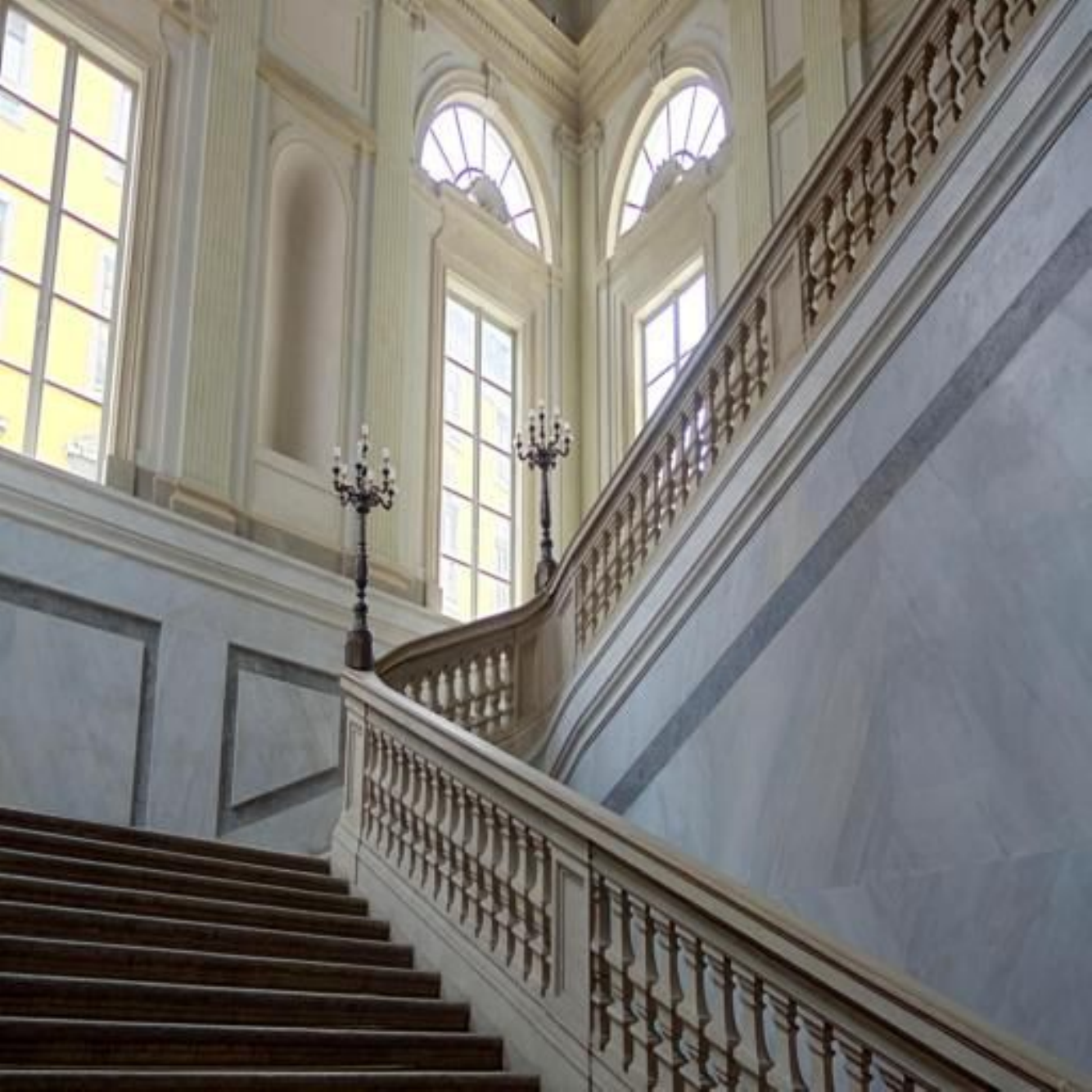}}
\caption{Visual comparison of different action spaces.}
\label{as2}
\end{figure*}

To further validate the impact of the action space settings, we use images from the SICE dataset~\cite{Cai2018deep} to retrain the network under different settings.
As shown in Fig.~\ref{as2}, trained using the same dataset, the model based on our action space settings performs better than ReLLIE.

\begin{table}[!htb]
\centering
{
\begin{tabular}{c|ccc}
\hline
&N = 1 &N = 3 &N = 6 \\\hline
Baseline& 0.2167& 0.5353& 0.7560\\
Ours& 0.2500& 0.6229& 0.8721\\
\hline
\end{tabular}}
\caption{Describes the adjustment range for different motion ranges, N denotes the number of steps.}
\label{LEC}
\end{table}

\section{Supplementary Experiment of Cost Comparisons}
It takes 1200s for an epoch on LOL training dataset with 485 images with batch size 2. Training 1000 epoch cost about 1 day using a 1080Ti GPU. Our test module has 278,816 trainable parameters and 113.85 GFlops.
With image 1200×900×3, at 6 steps in the test phase setting, each step costs 0.1506s.
As shown in Table~\ref{cost}, we list the Flops and the test runtime comparisons on a GTX1080Ti GPU. The model efficiency of the proposed ALL-E is at the middle level.
% Ours 1.0545s, 113.85G. ReLLIE 1.480s, 125.13G. LLNet 36.270s, 4124.17G. MBLLEN 13.995s, 108.53G. KinD++ 1.068s, 12238.02G. Zero-DCE 0.003s 84.99G.
% EnligtenGAN 0.008s, 273.24G. SCL-LLE 0.004s 95.21G.

\begin{table}[!htb]
\centering
{
\begin{tabular}{c|cc}
\hline
Method & GFLOPs $\downarrow$ &runtime (s) $\downarrow$\\\hline
(PR'17) LLNet & 4124.17 &36.270 \\
(BMCV'18) MBLLEN& 301.12 & 13.995\\
(IJCV'21) KinD++& 12238.02 & 1.068\\
(CVPR'20) Zero-DCE &84.99 & \textbf{0.003}\\
(TIP'21) EnlightenGAN &273.24 &0.008\\
(ACMMM'21) ReLLIE &125.13 &1.480\\
(AAAI'22) SCL-LLE &95.21 &0.004\\ 
(TIP'17) LIME& (on CPU) & 21.530\\
(BMVC'18) Retinex-Net& 587.470 &0.120 \\
(ACMMM'20) ISSR&  (unavailable) & 9.645\\
(CVPR'21) RUAS &1.069 & 0.006\\
(CVPR'22) Uretinex-net &938.235 &0.400\\
(ICCV'21) Zhao \emph{et al.} &12438.282 &4.216\\
(CVPR'22) Ma \emph{et al.}&\textbf{0.580} &0.010\\
\hline
Ours&113.85 &1.055\\
\hline
\end{tabular}}
\caption{Experimental results of model efficiency.}
\label{cost}
\end{table}

\section{Supplementary Experimental of Downstream Face Detection Task}
We test face detection after LLE on DARK FACE dataset \cite{yuan2019ug}. RetinaFace \cite{Deng_2020_CVPR} trained on WIDER FACE \cite{Yang_2016_CVPR} dataset is chosen as the detector in Table~\ref{FACE_D}. Our method improves face detection, but not on the top. We believe this result is reasonable, as our method compromises the human aesthetic feeling and the image quality. Note that the Top-2 are LIME (a Retinex-based none-deep-learning method) and SCL-LLE (uses a downstream task via multi-task learning), providing solutions for further improving the gain in downstream tasks. 
%We will supplement this part in the manuscript.
\begin{table}[!htb]
\centering
{\begin{tabular}{c|ccc}
\hline
 \multirow{2}*{Methods}&\multicolumn{3}{c}{IOU thresholds}\\ 
 \cline{2-4}
& 0.5& 0.7& 0.9\\ \hline
 Input&0.2820 &0.0693 &0.0002 \\
 (TIP'17) LIME&\textbf{0.4221} &0.1068 &\textbf{0.0004} \\
 (BMVC'18) RetinexNet&0.3874 &0.1065 &0.0002 \\
 (ACMMM'20) ISSR&0.2825 &0.0674 &0.0001 \\
 (CVPR'20) Zero-DCE&0.4130 &0.1067 &0.0002 \\
 (TIP'21) EnlightenGAN&0.3762 &0.1009 &{0.0003} \\
 (CVPR'21) RUAS&0.2782 &0.0659 &0.0002 \\
 (ACMMM'21) ReLLIE&0.3583 &0.0958 &0.0001 \\
 (AAAI'22) SCL-LLE&0.4199 &\textbf{0.1082} &{0.0003} \\
 \hline
 Ours &0.3985 &0.1032 &0.0003 \\
\hline
\end{tabular}}
\caption{ The average precision (AP) for face detection in the dark under different IoU thresholds (0.5, 0.7, 0,9).}
\label{FACE_D}
\end{table}

\section{Compare the Aesthetics Assessment Metric}
The most commonly used metrics of aesthetics, SRCC and LCC, require subjective human scores, which are not offered in any LLE datasets. Alternatively, we chose ReLIC \cite{ReLIC}, the most accurate in all available aesthetic models (with 82.35$\%$ accuracy report above), as a third-party aesthetic evaluator to test the LLE outputs shown in Table \ref{AA}. Our method is runner-up, which is comparable with EnlightenGAN \cite{jiang2021enlightengan}. 

\begin{table}[!htb]
\centering
{\begin{tabular}{c|c}
\hline
{Methods}&aesthetic scores\\ \hline
 Input&0.2820 \\
 (TIP'17) LIME&{6.038}  \\
 (BMVC'18) RetinexNet&5.800 \\
 (ACMMM'20) ISSR&6.015\\
 (CVPR'20) Zero-DCE&6.078\\
 (TIP'21) EnlightenGAN&\textbf{6.106} \\
 (CVPR'21) RUAS&5.500\\
 (ACMMM'21) ReLLIE&6.020\\
 (AAAI'22) SCL-LLE&6.063 \\
 (CVPR'22) Uretinex-net & 6.040\\
 (ICCV'21) Zhao \emph{et al.}&6.011\\
 (CVPR'22) Ma \emph{et al.} &6.052\\
 \hline
 Ours &6.094\\
\hline
\end{tabular}}
\caption{Compare the aesthetics assessment with peer methods.}
\label{AA}
\end{table}
\begin{figure*}[h!]
\centering
\subfigure[Input]{
\includegraphics[height=7.7cm, width=7.4cm]{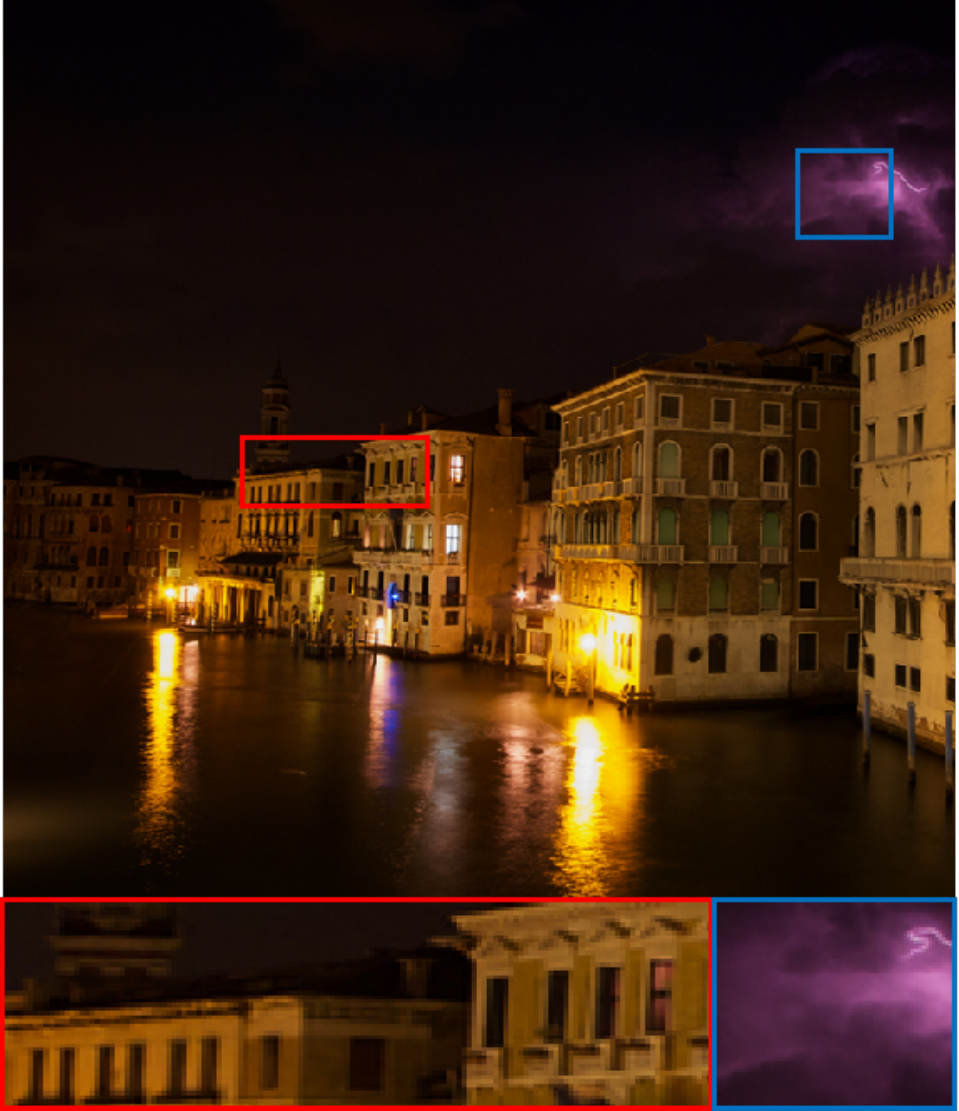}
}
\subfigure[ReLLIE (ACMMM’21)]{
\includegraphics[height=7.7cm, width=7.4cm]{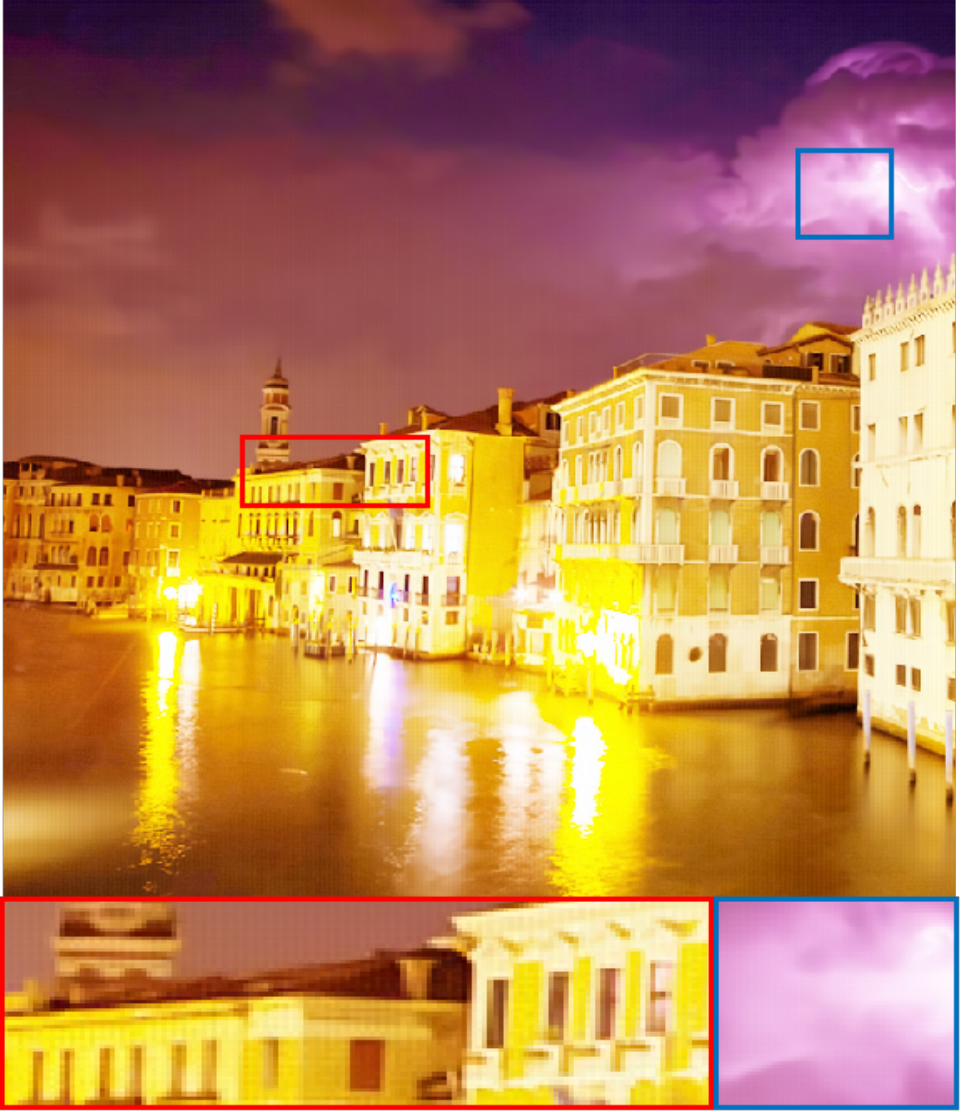}
}
\subfigure[Ours]{
\includegraphics[height=7.7cm, width=7.4cm]{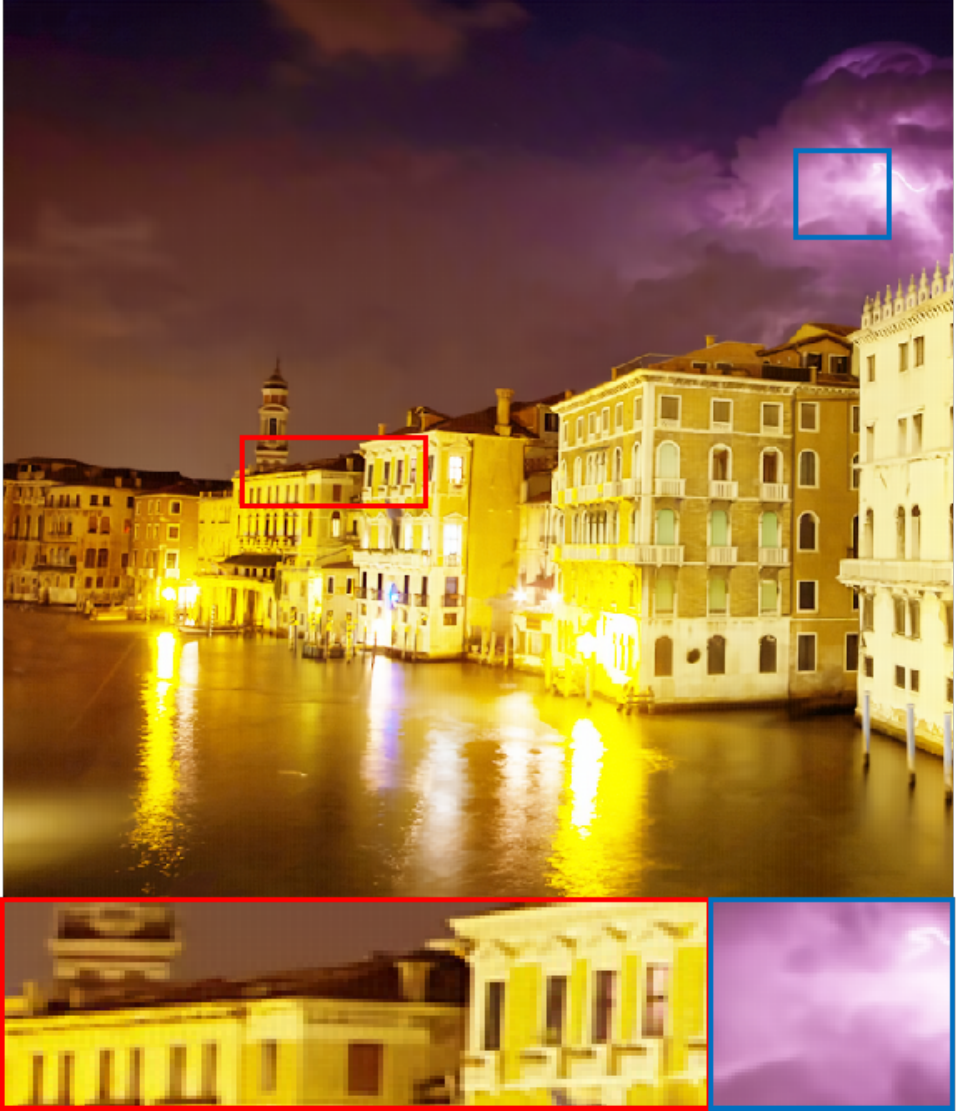}
}
\caption{The failure case when the scene's theme is nightscape.
}
\label{fc}
\end{figure*}

\section{Failure Case}
At last, we show a controversial situation when we conduct the user study, maybe a type of failure case or limitation of ALL-E method, in Fig.\ref{fc}. When the scene's theme is nightscape, such as buildings and water under the moonlight shown in Fig.\ref{fc}, the photographer is often unwilling to fully expose the whole scene to approximate the visual effect in the daytime. However, in this case, the proposed ALL-E will also produce large-scale exposure adjustment; the enhanced image using ALL-E appears to be over-exposure and over-smoothed in the detail region, which is controversial in the user study. In our future work, we will consider introducing more high-level semantic theme guidance to alleviate this excessive exposure adjustment.

\section{Supplementary Visual Quality Comparison}
\begin{figure*}[!h]
\centering
\subfigure[Input]{
\includegraphics[height=3.7cm,width=5.4cm]{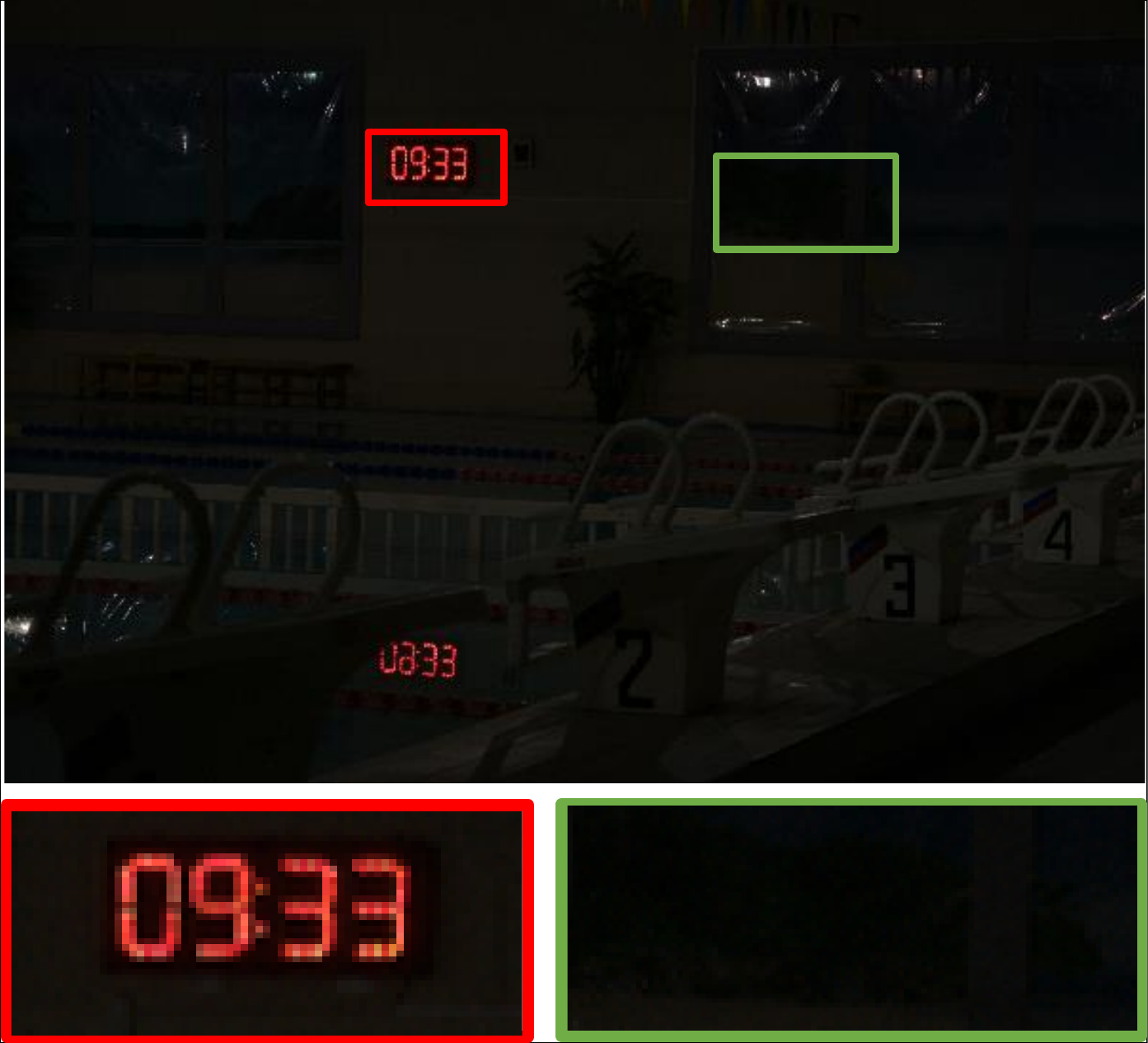}
}
\hspace{.10in}
\subfigure[LIME]{
\includegraphics[height=3.7cm,width=5.4cm]{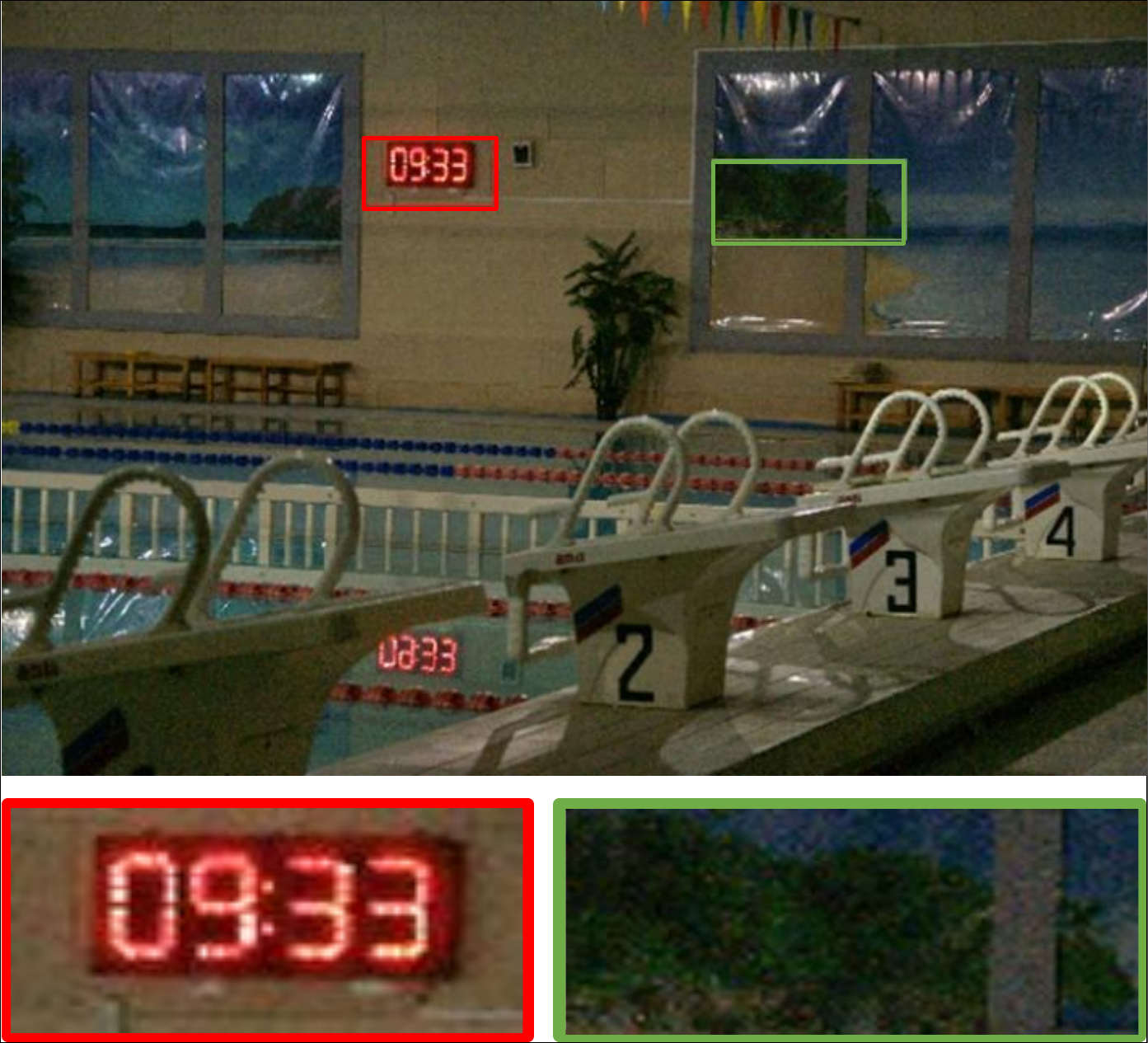}
}
\hspace{.10in}
\subfigure[Retinex-Net]{
\includegraphics[height=3.7cm,width=5.4cm]{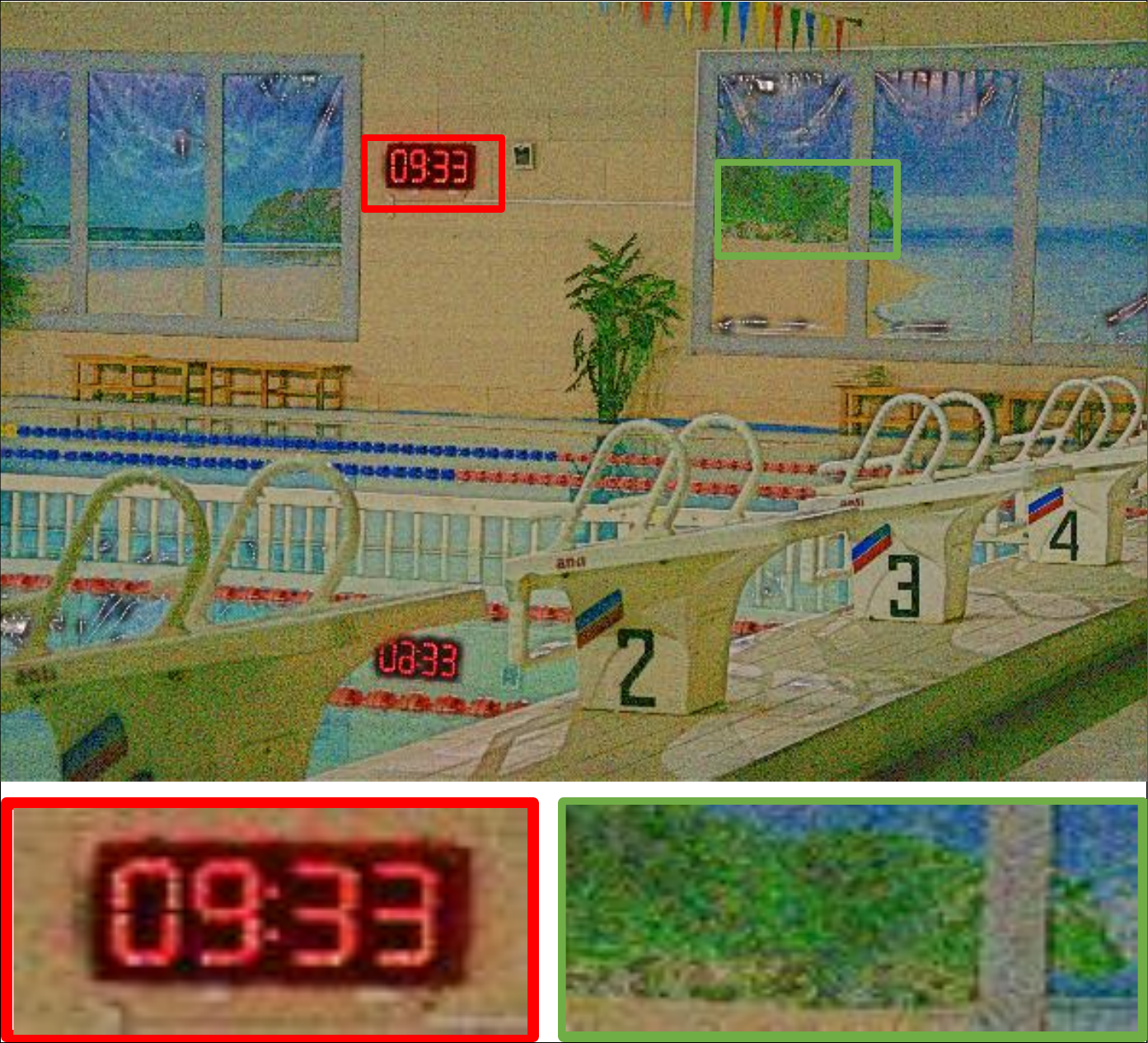}
}
\hspace{.10in}
\subfigure[ISSR]{
\includegraphics[height=3.7cm,width=5.4cm]{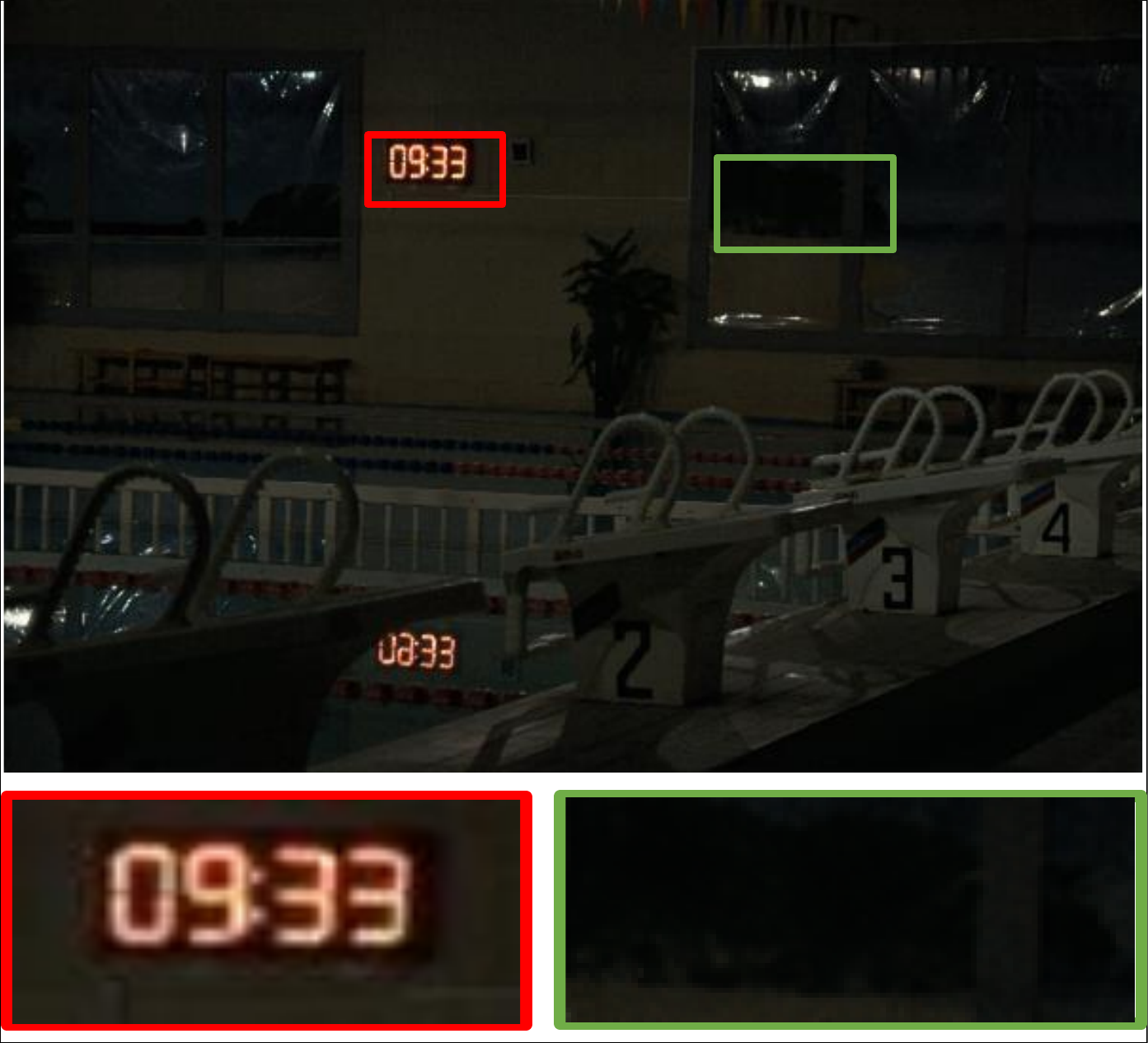}
}
\hspace{.10in}
\subfigure[Zero-DCE]{
\includegraphics[height=3.7cm,width=5.4cm]{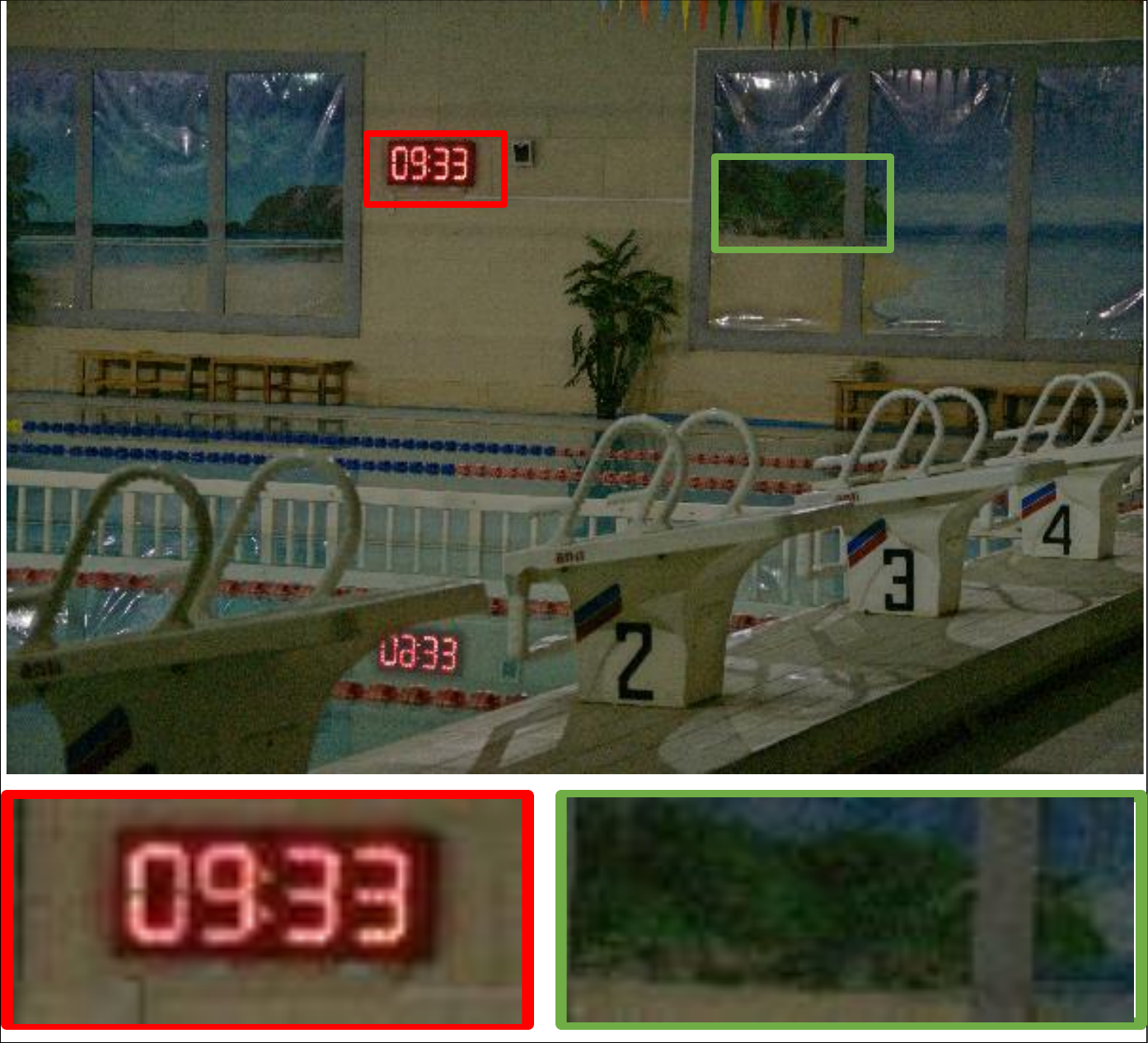}
}
\hspace{.10in}
\subfigure[EnlightenGAN]{
\includegraphics[height=3.7cm,width=5.4cm]{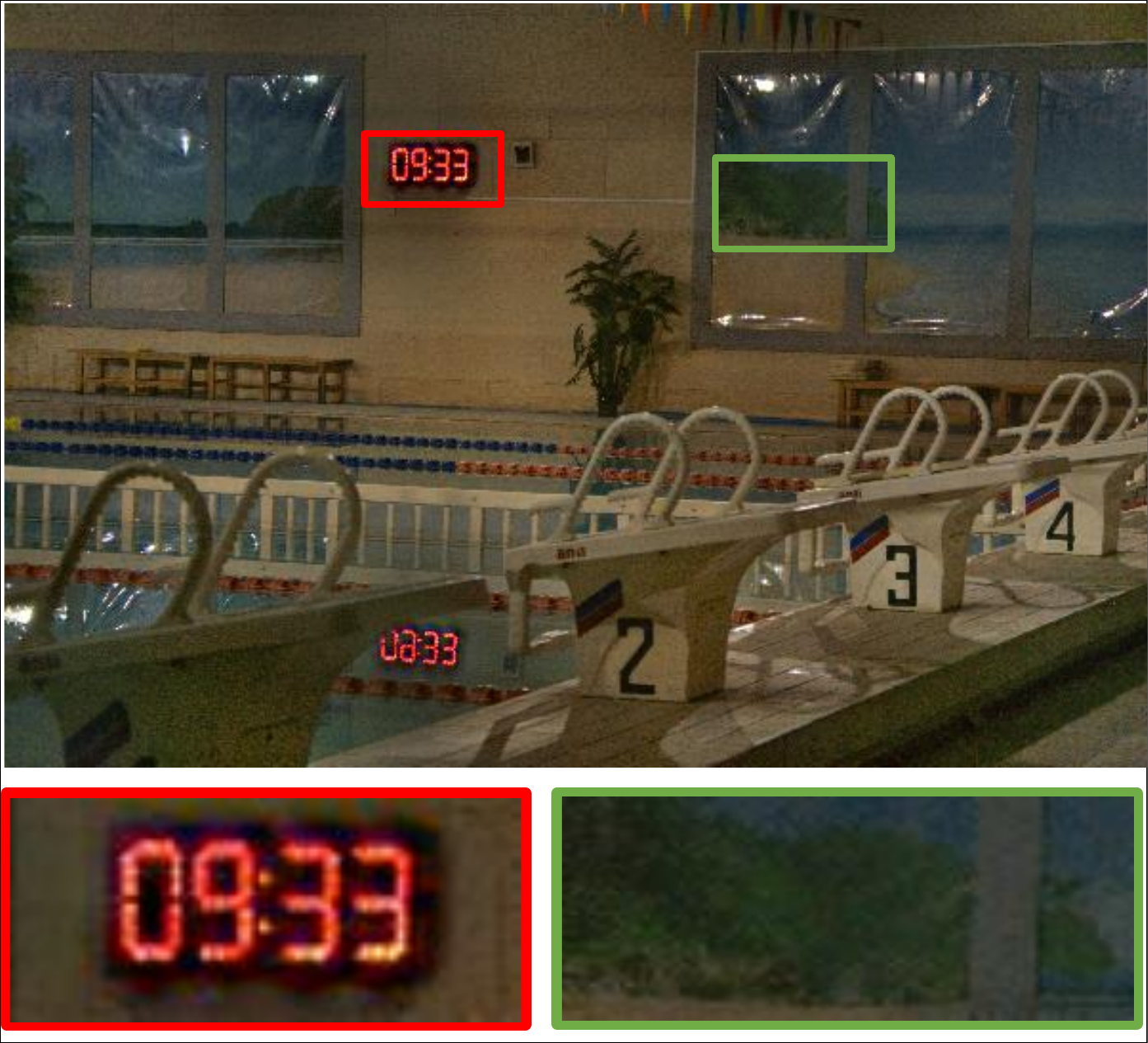}
}
\hspace{.10in}
\subfigure[RUAS]{
\includegraphics[height=3.7cm,width=5.4cm]{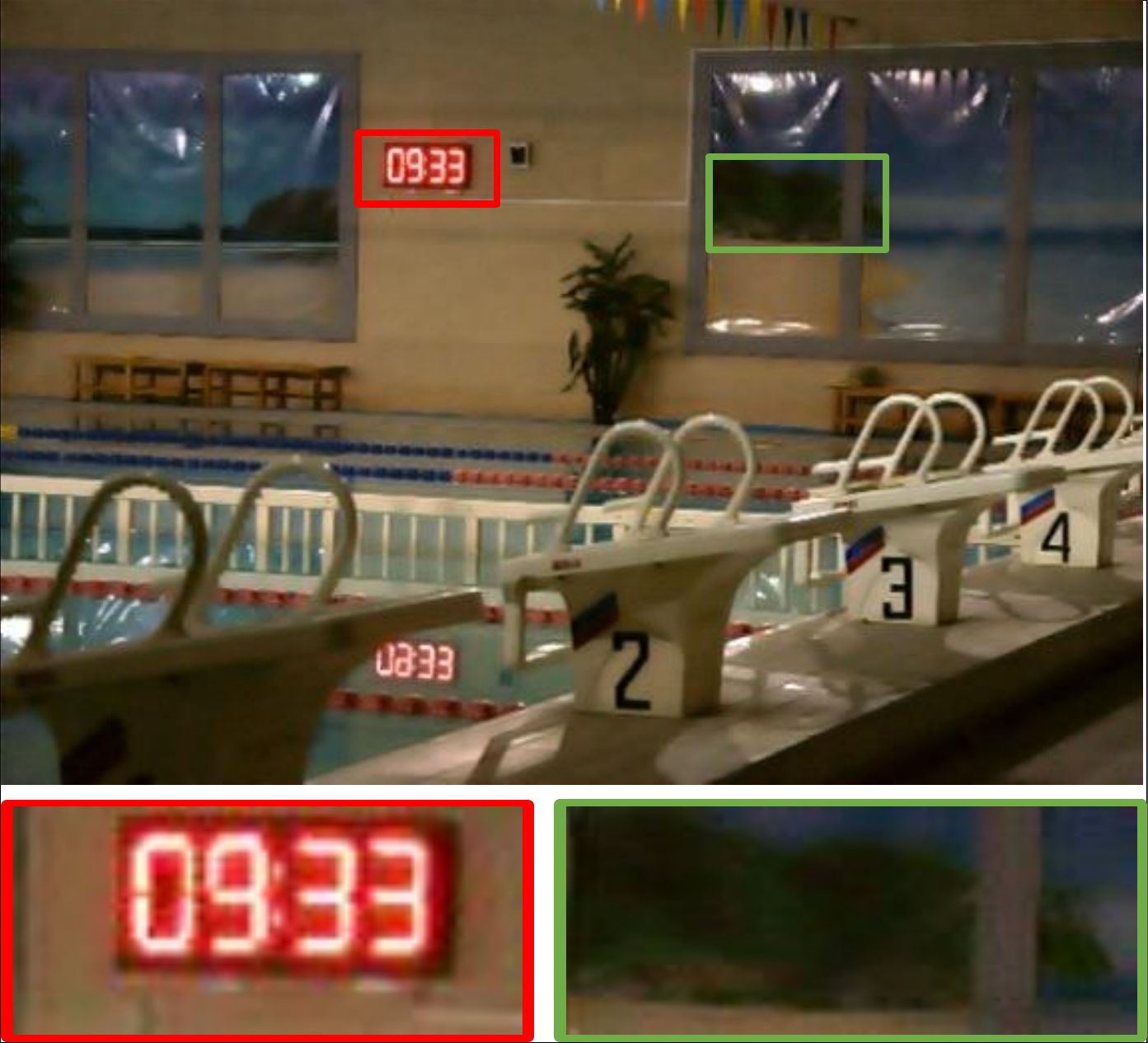}
}
\hspace{.10in}
\subfigure[ReLLIE]{
\includegraphics[height=3.7cm,width=5.4cm]{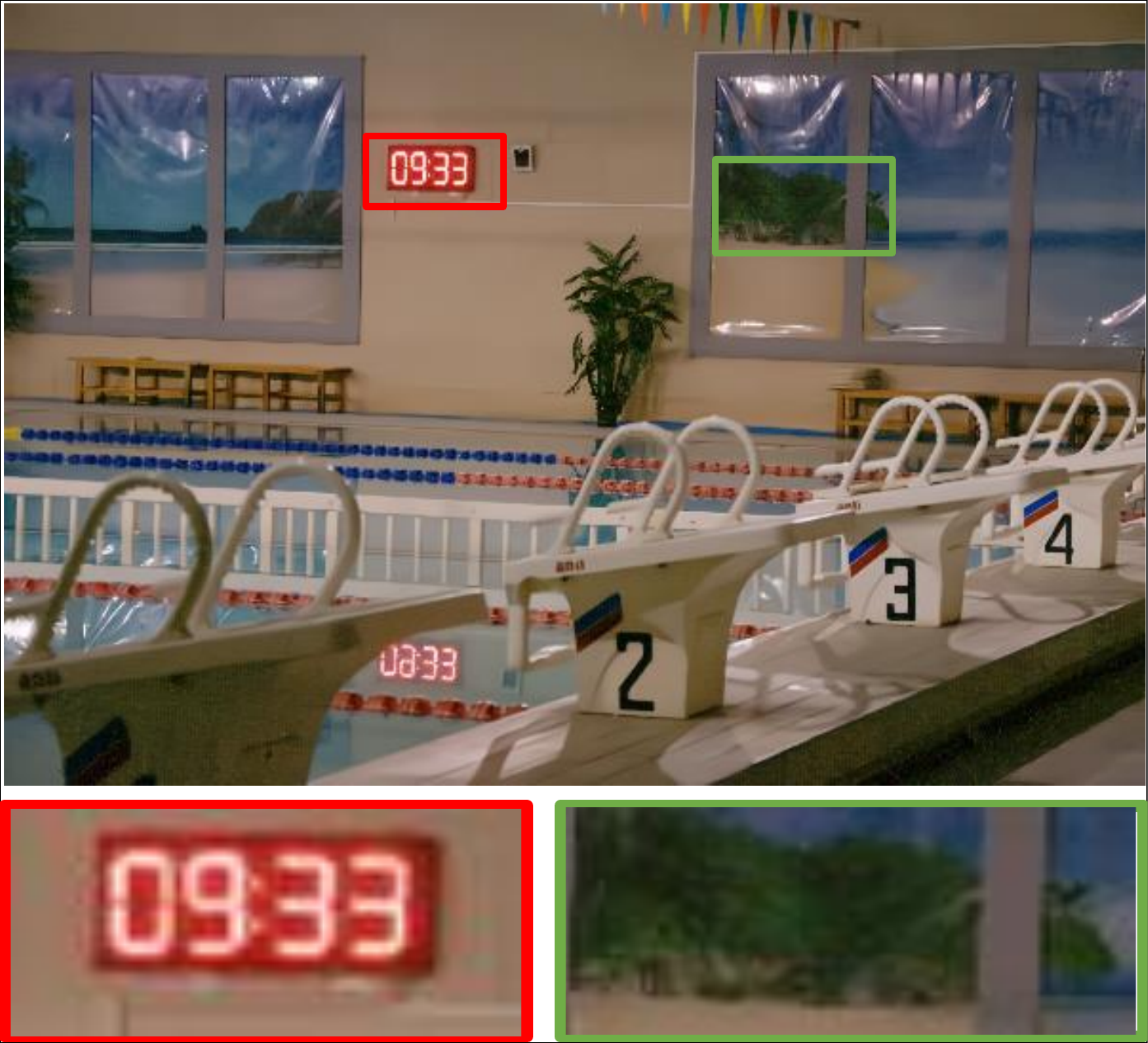}
}
\hspace{.10in}
\subfigure[SCL-LLE]{
\includegraphics[height=3.7cm,width=5.4cm]{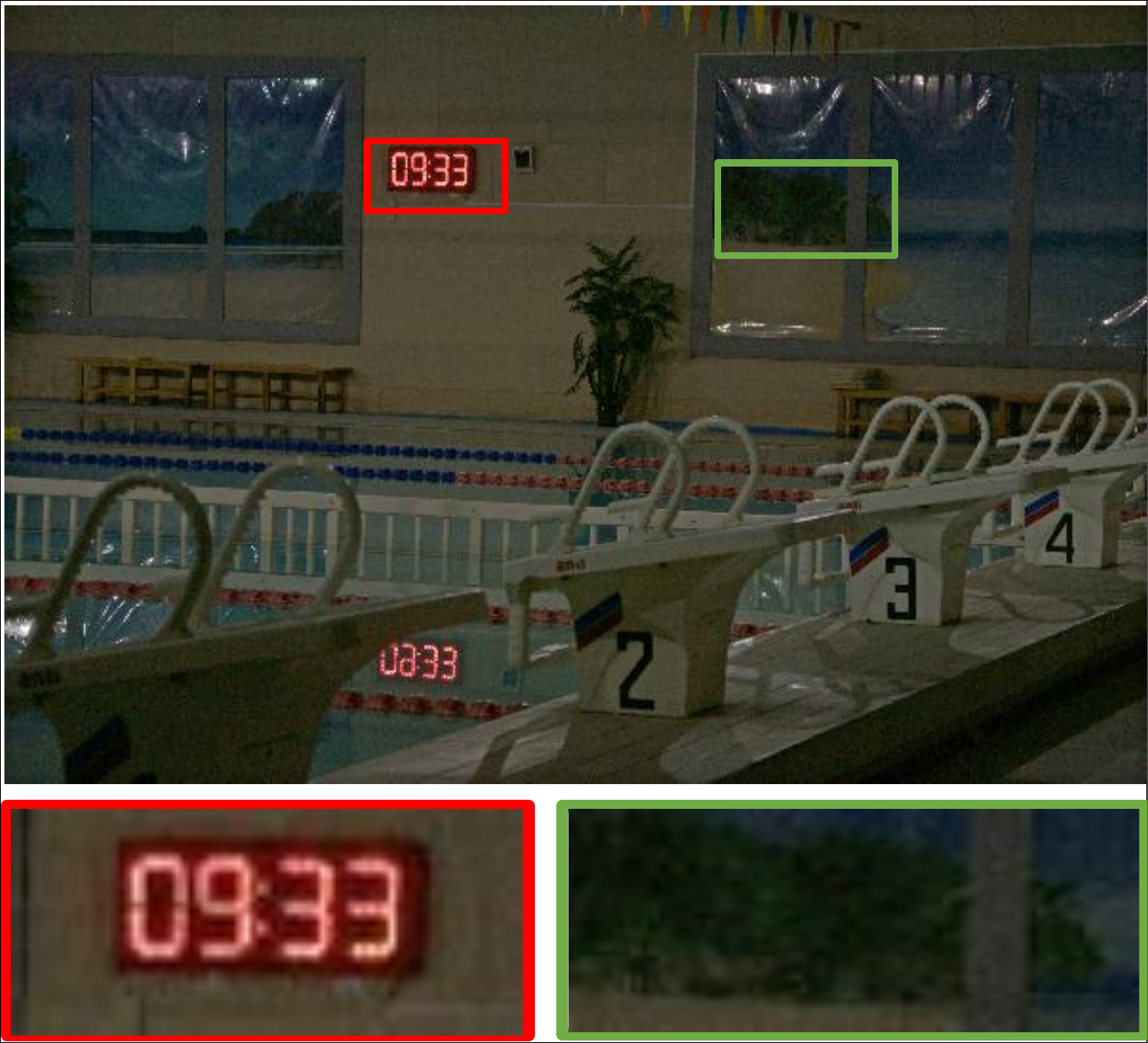}
}
\hspace{.10in}
\subfigure[Ours]{
\includegraphics[height=3.7cm,width=5.4cm]{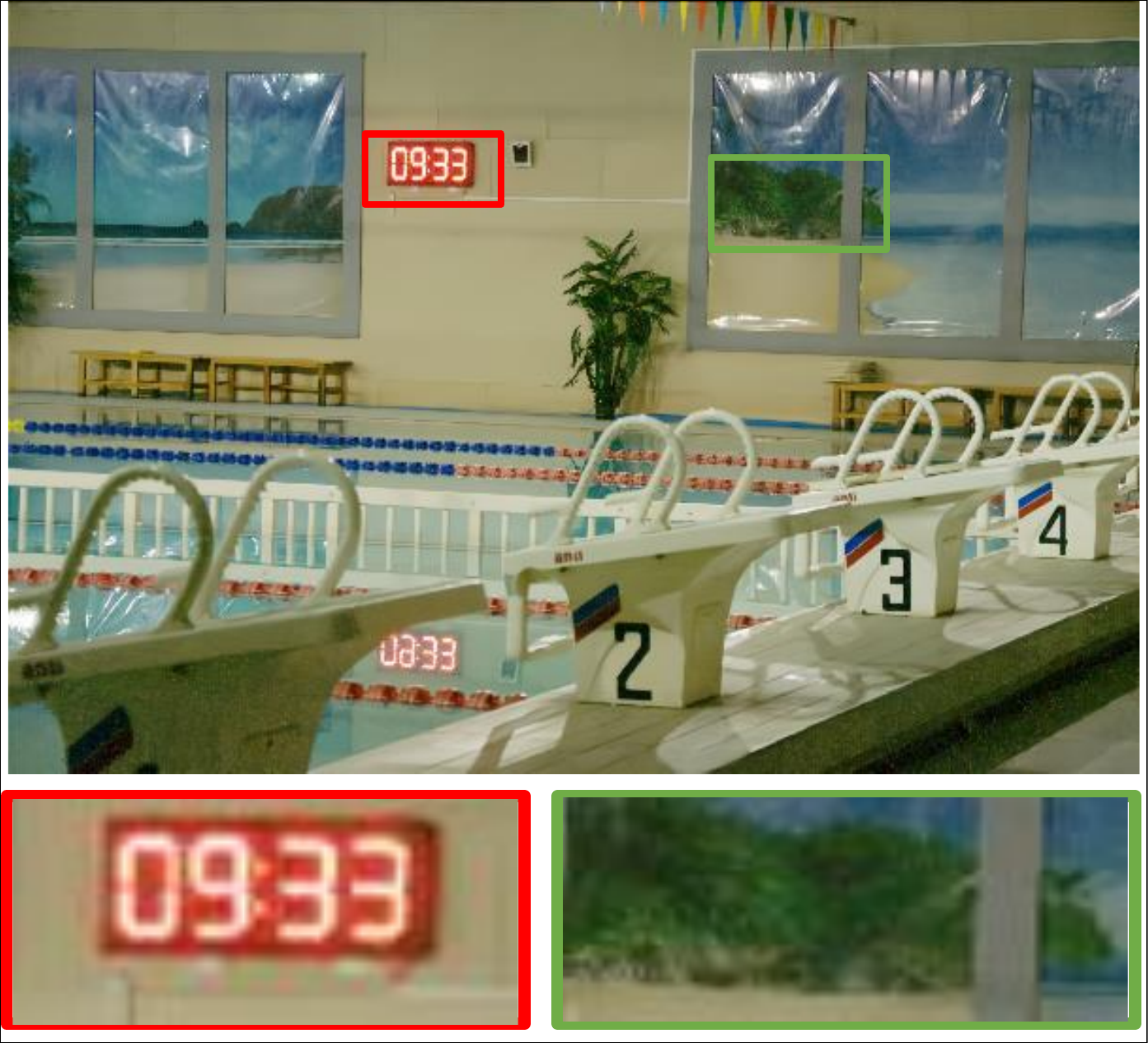}
}
\caption{Compared to the state-of-the-art methods, ALL-E produces fewer artifacts, and the overall visual experience is superior.}
\label{Vq1}
% \vspace{-1cm}
\end{figure*}

\begin{figure*}[h]
\centering
\subfigure[Input]{
\includegraphics[height=4cm,width=5cm]{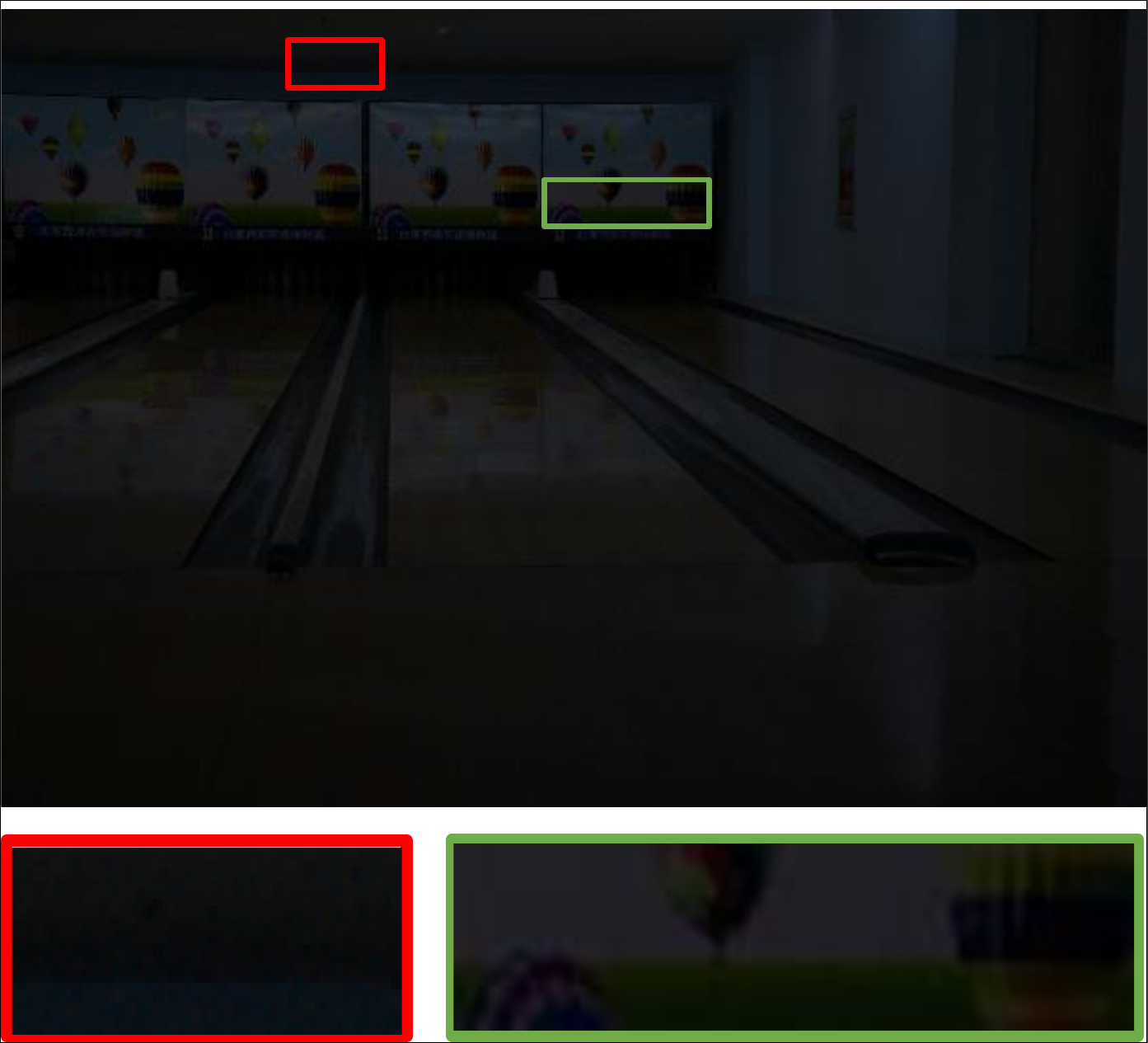}
}
\subfigure[LIME]{
\includegraphics[height=4cm,width=5cm]{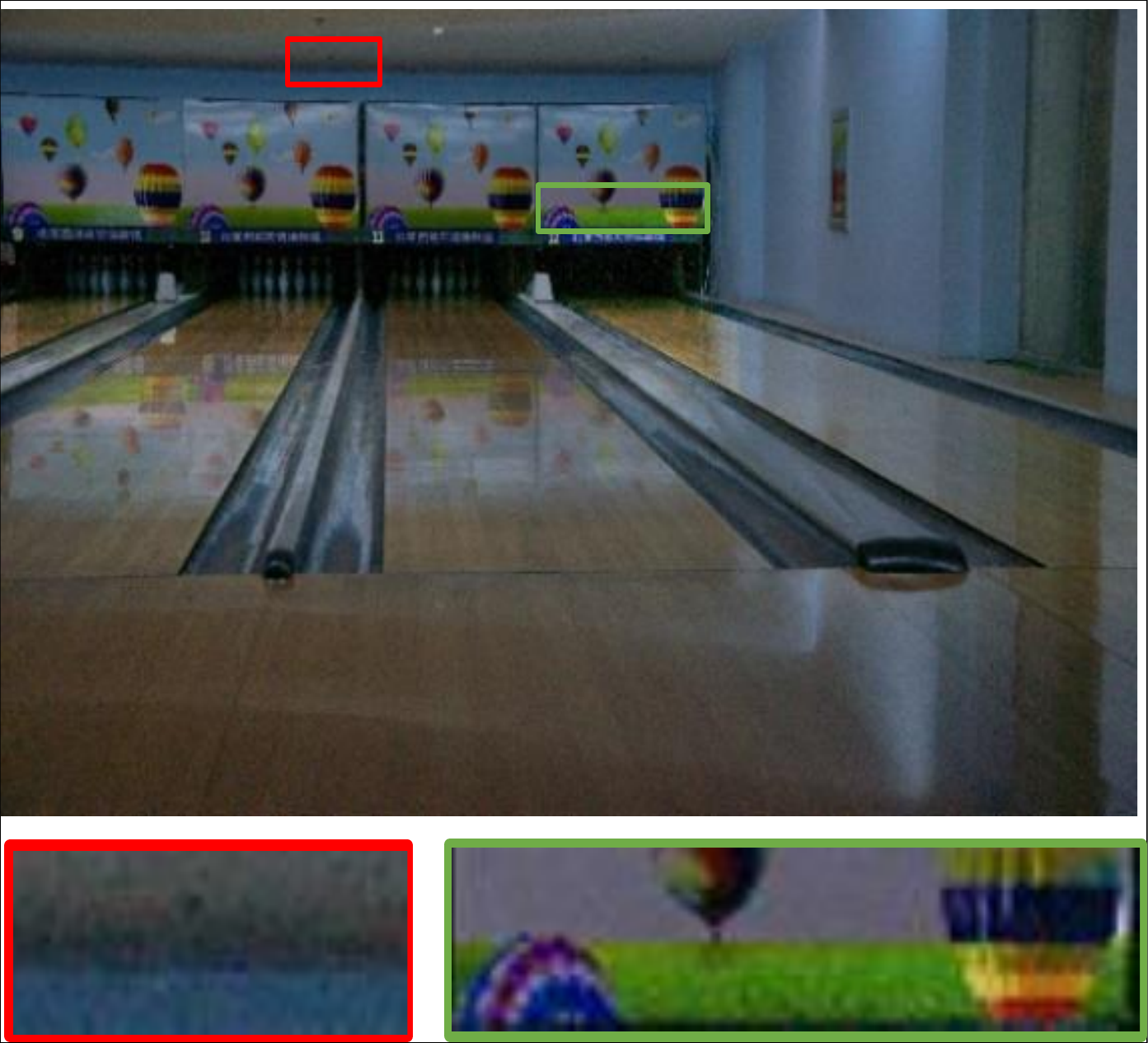}
}
% \end{figure*}
% \begin{figure*}[h]
% \centering
\subfigure[Retinex-Net]{
\includegraphics[height=4cm,width=5cm]{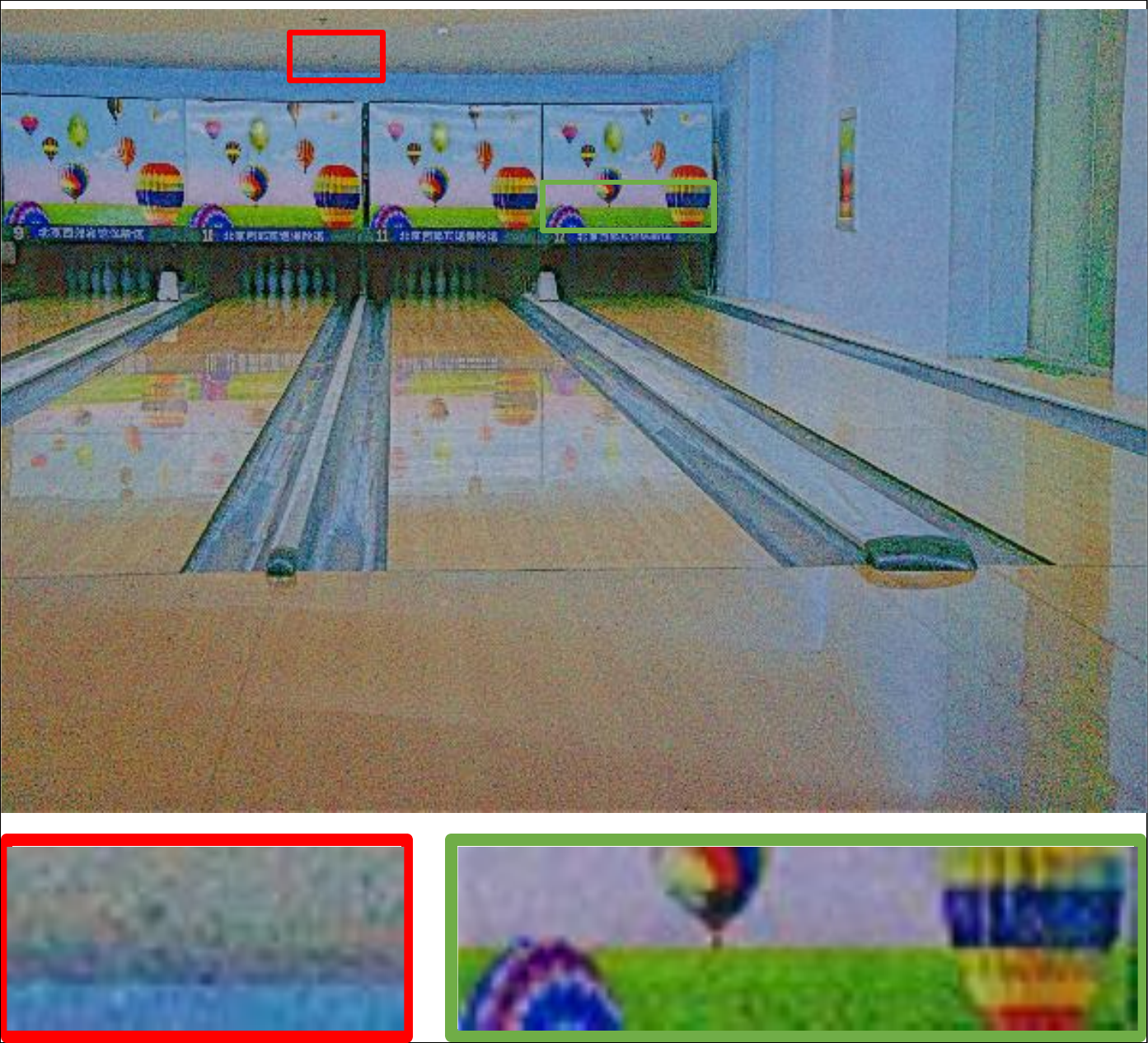}
}
\subfigure[ISSR]{
\includegraphics[height=4cm,width=5cm]{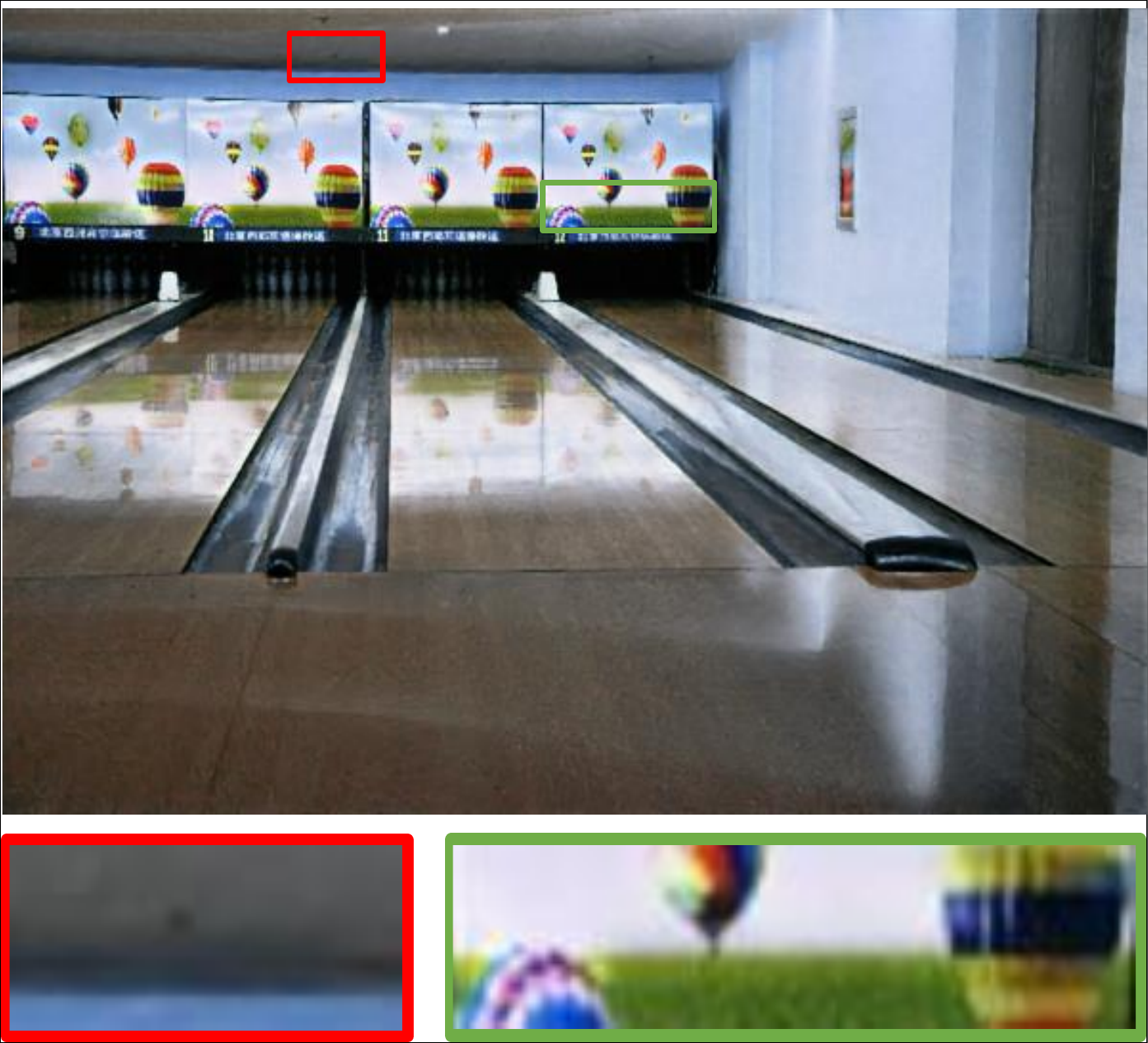}
}
\subfigure[Zero-DCE]{
\includegraphics[height=4cm,width=5cm]{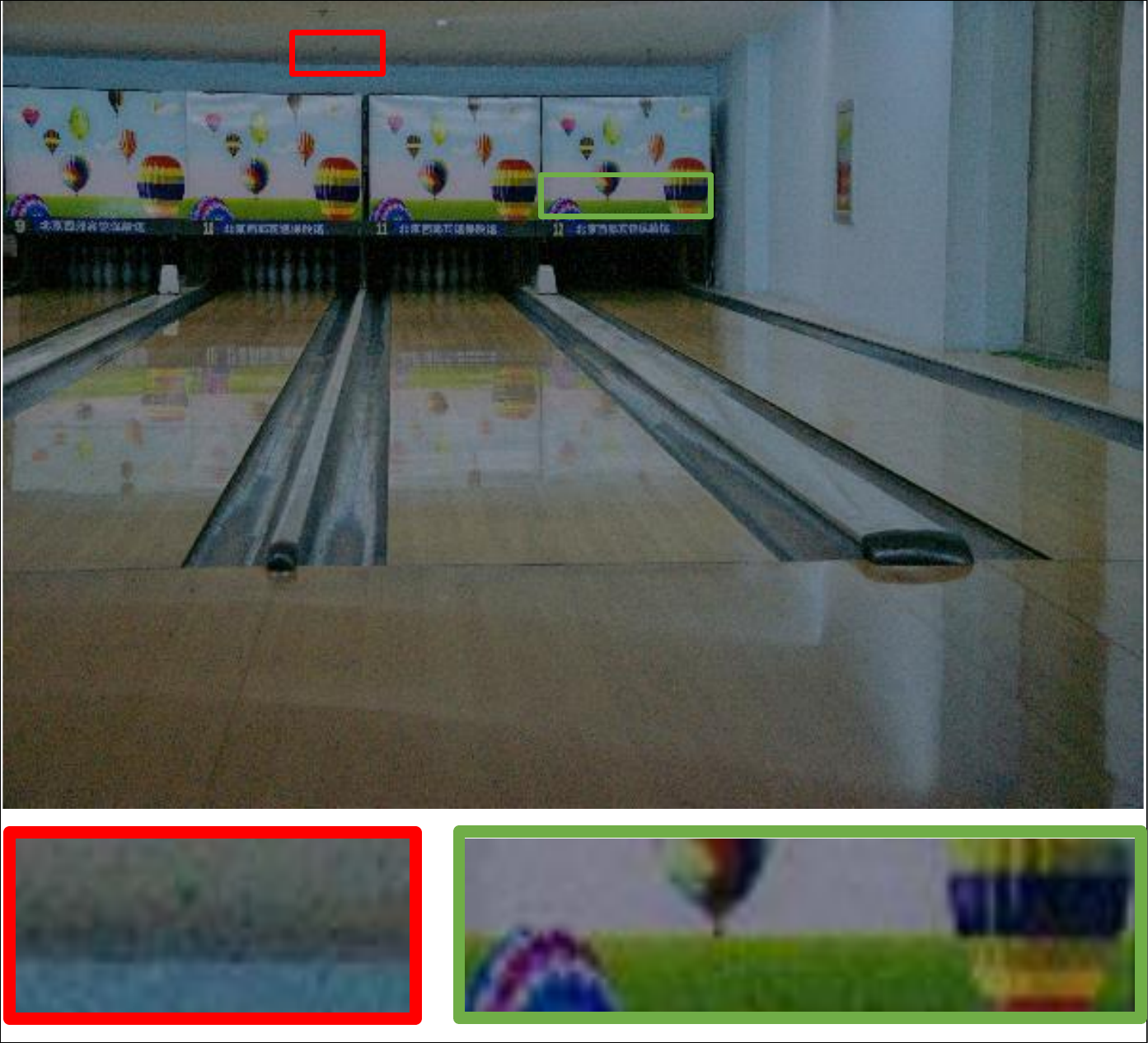}
}
\subfigure[EnlightenGAN]{
\includegraphics[height=4cm,width=5cm]{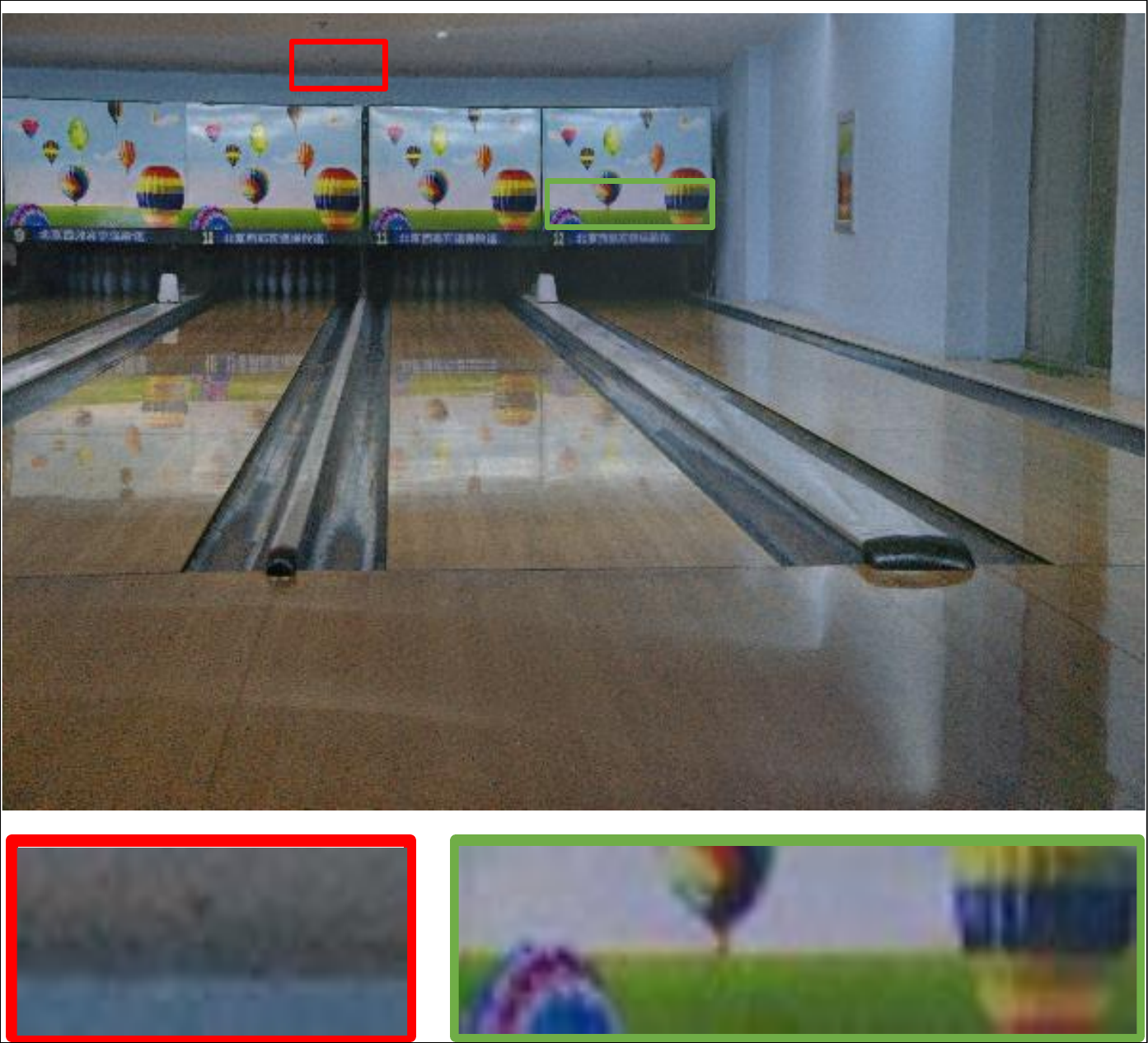}
}
% \end{figure*}
% \begin{figure*}[h]
% \centering
\subfigure[RUAS]{
\includegraphics[height=4cm,width=5cm]{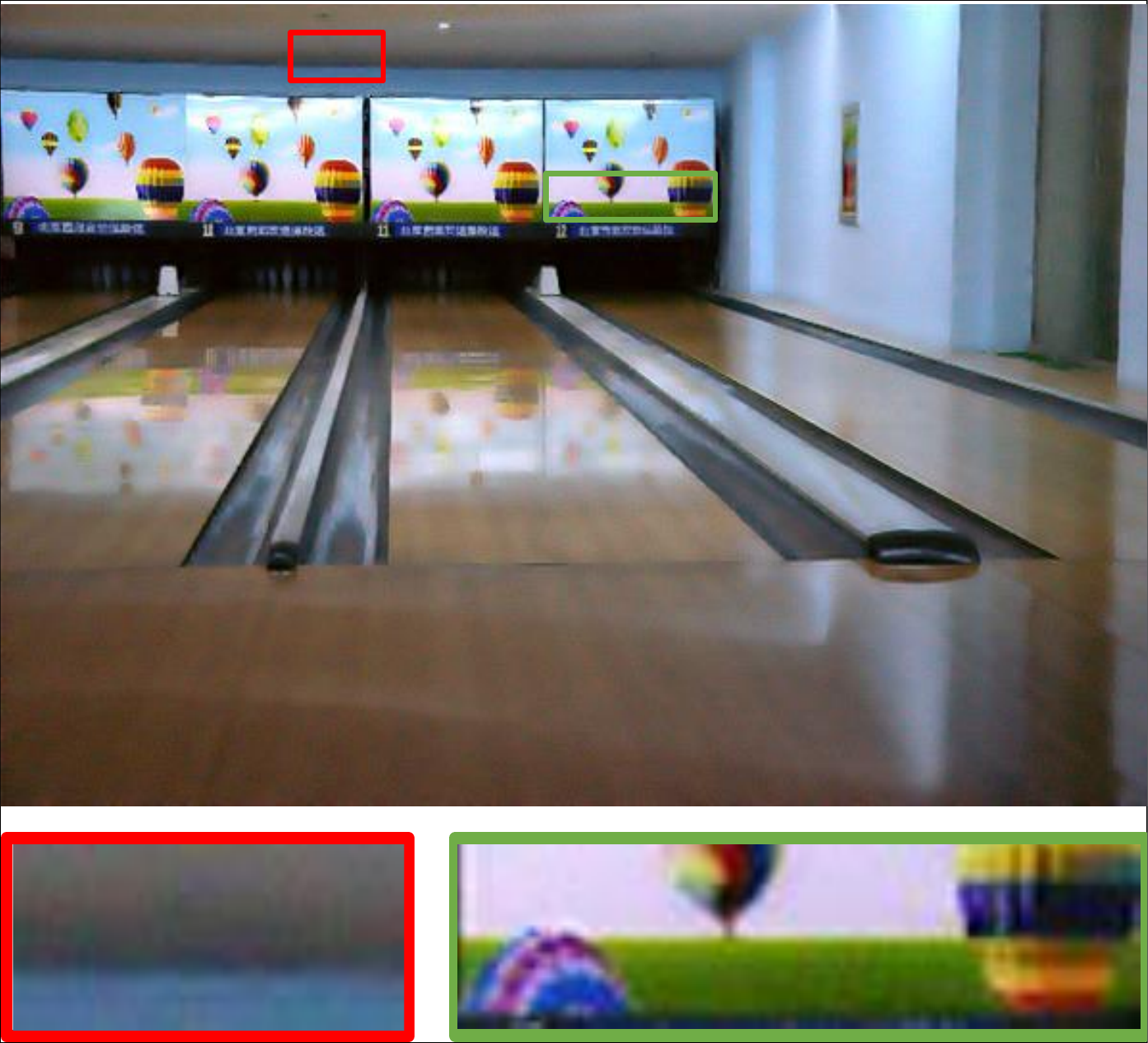}
}
\subfigure[ReLLIE]{
\includegraphics[height=4cm,width=5cm]{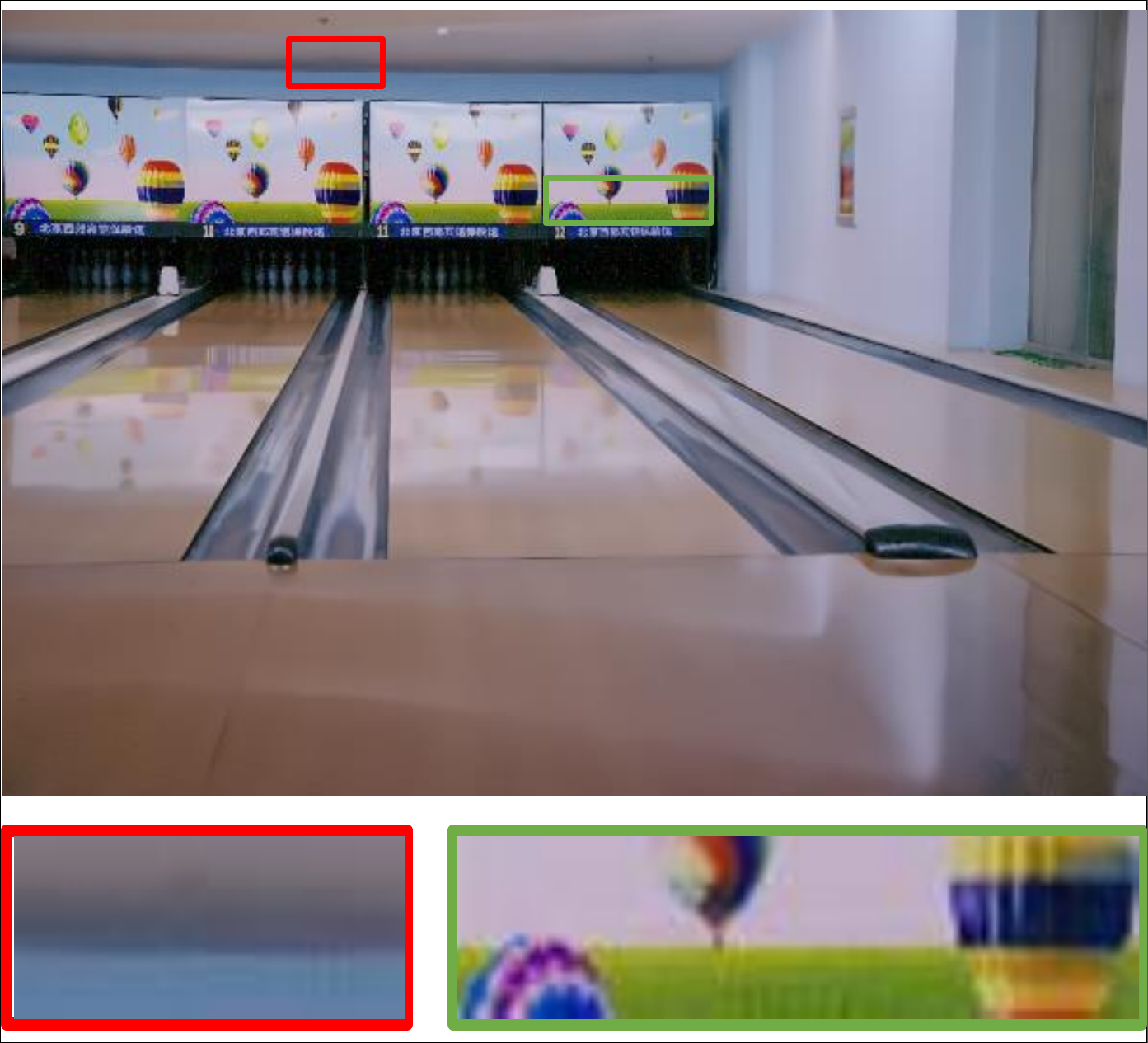}
}
% \end{figure*}
% \begin{figure*}[h]
% \centering
\subfigure[SCL-LLE]{
\includegraphics[height=4cm,width=5cm]{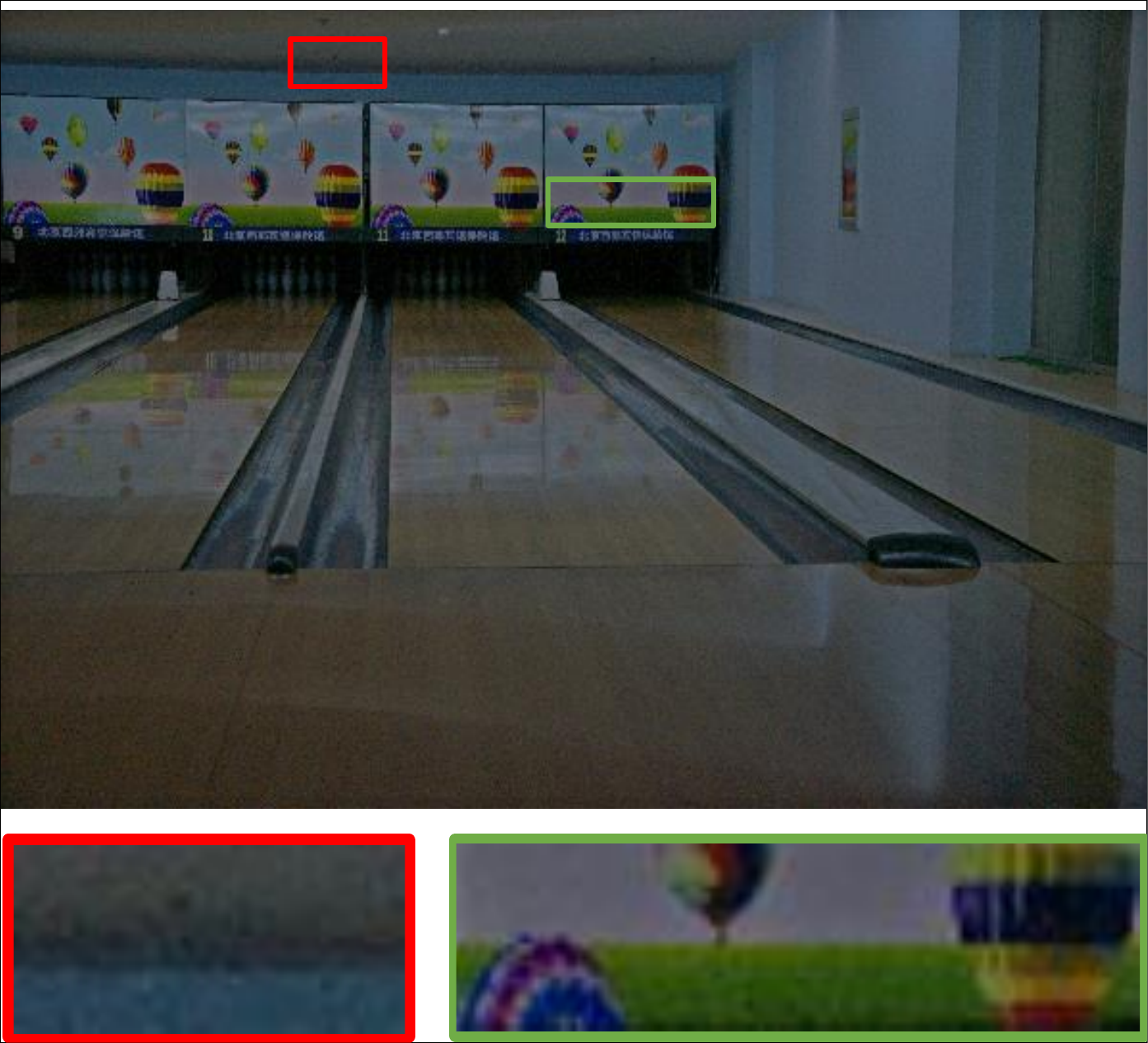}
}
\subfigure[Ours]{
\includegraphics[height=4cm,width=5cm]{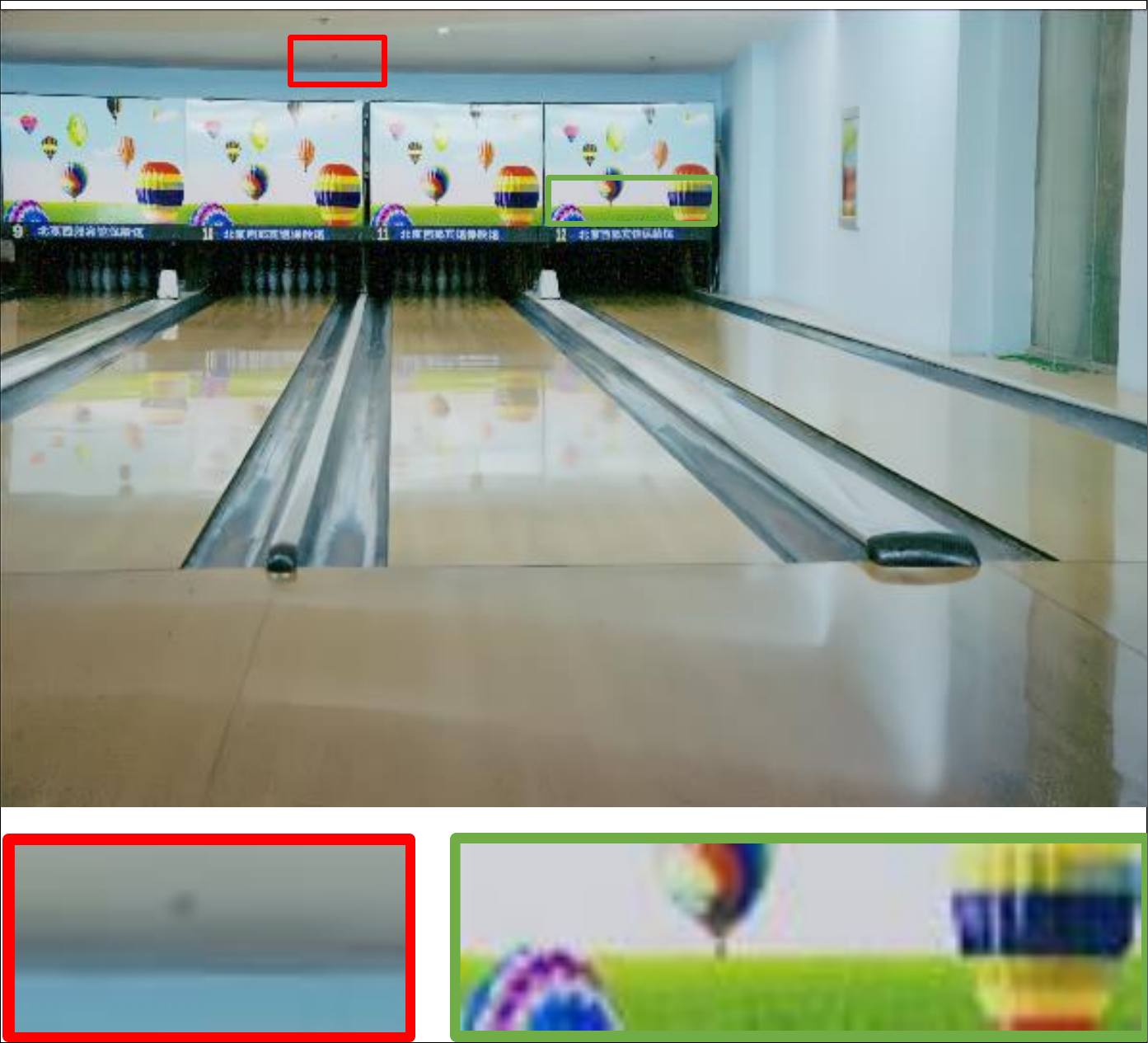}
}
\caption{Compared to the state-of-the-art methods, our method preserves more details and colors while improving the image's overall brightness, making the enhancement results more natural.}
\label{Vq2}
\end{figure*}

\begin{figure*}[h]
\centering
\subfigure[Input]{
\includegraphics[height=4cm,width=7cm]{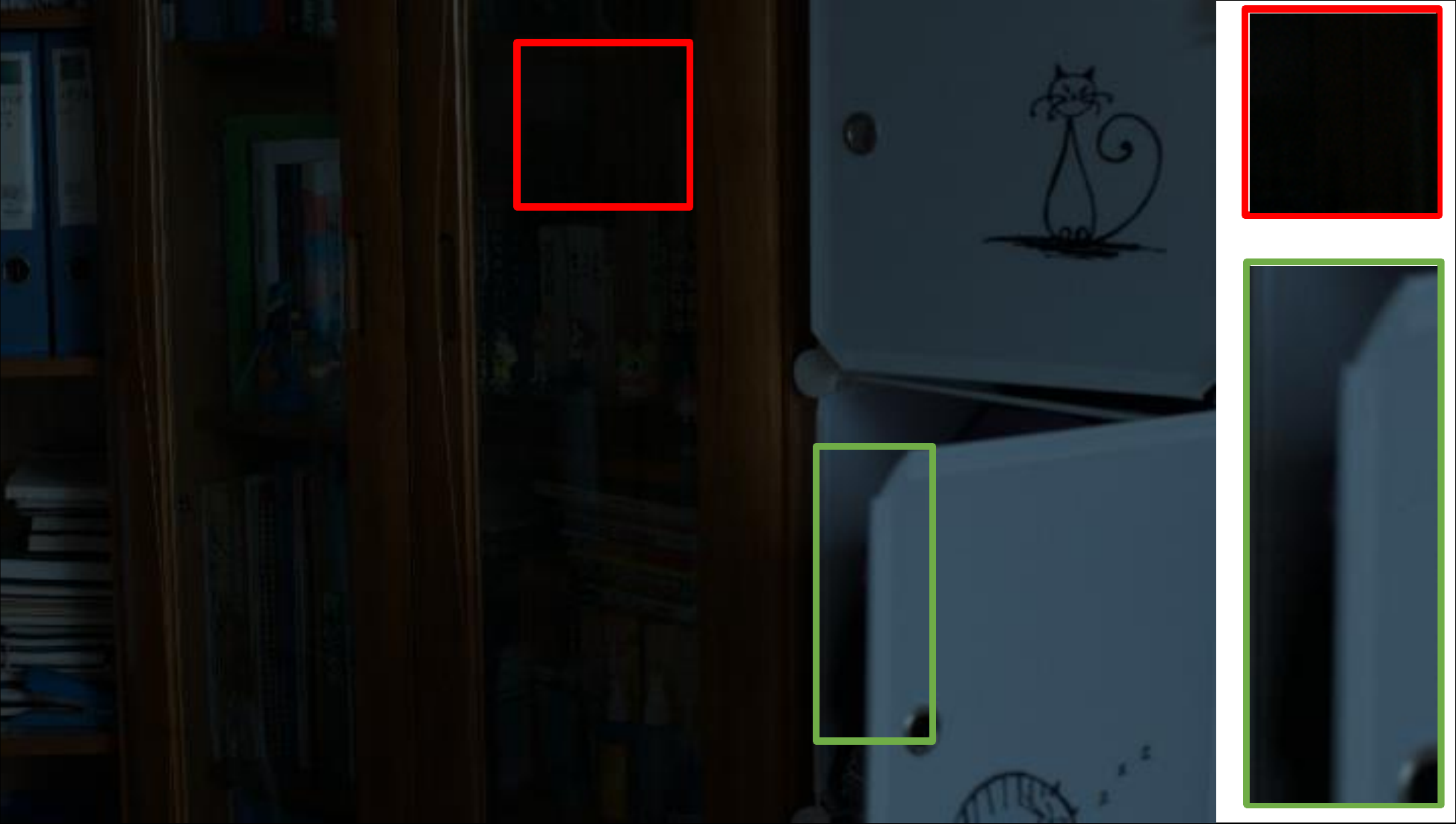}
}
\hspace{.15in}
\subfigure[LIME]{
\includegraphics[height=4cm,width=7cm]{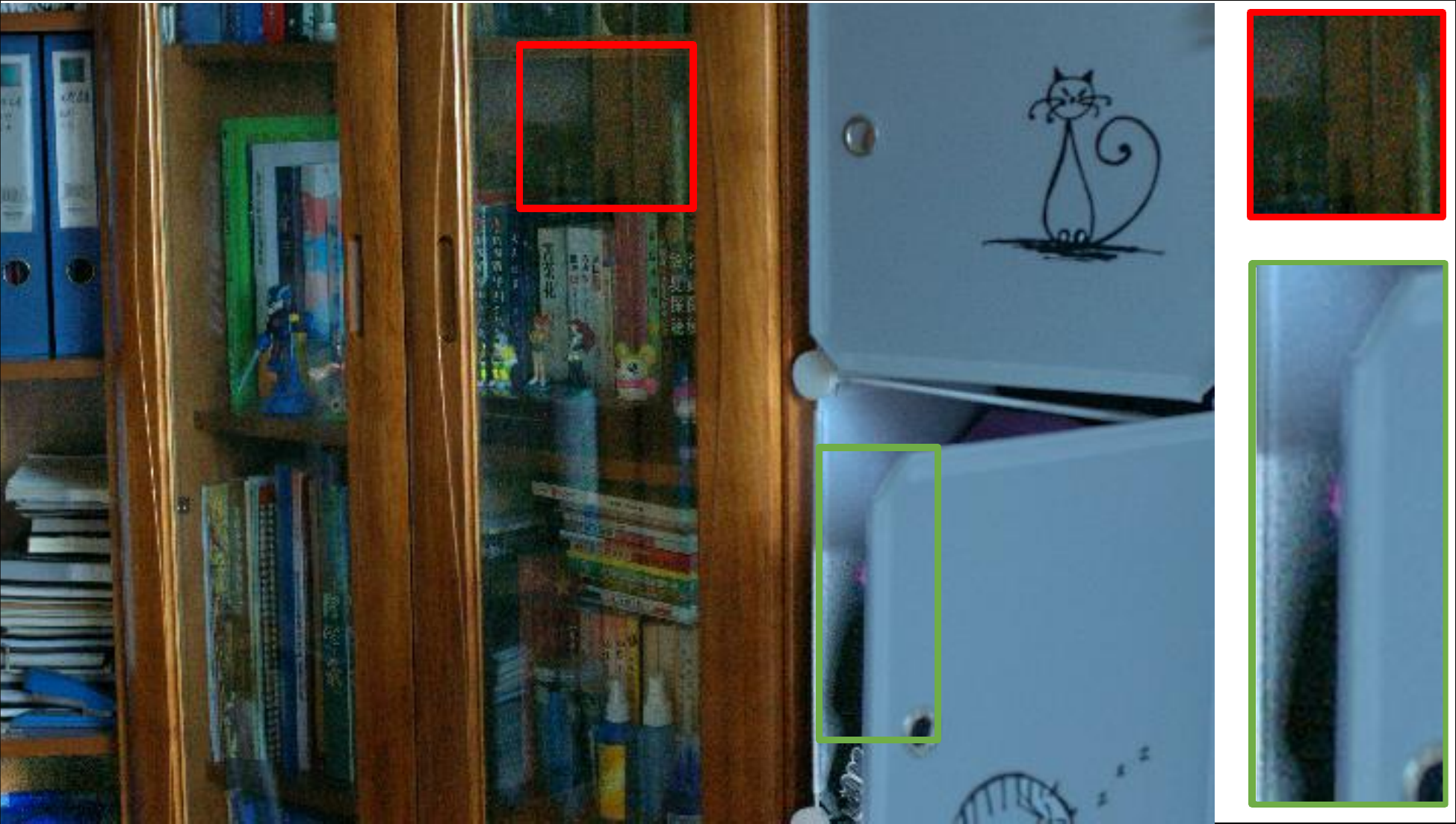}
}
\hspace{.15in}
% \end{figure*}
% \begin{figure*}[h]
% \centering
\subfigure[Retinex-Net]{
\includegraphics[height=4cm,width=7cm]{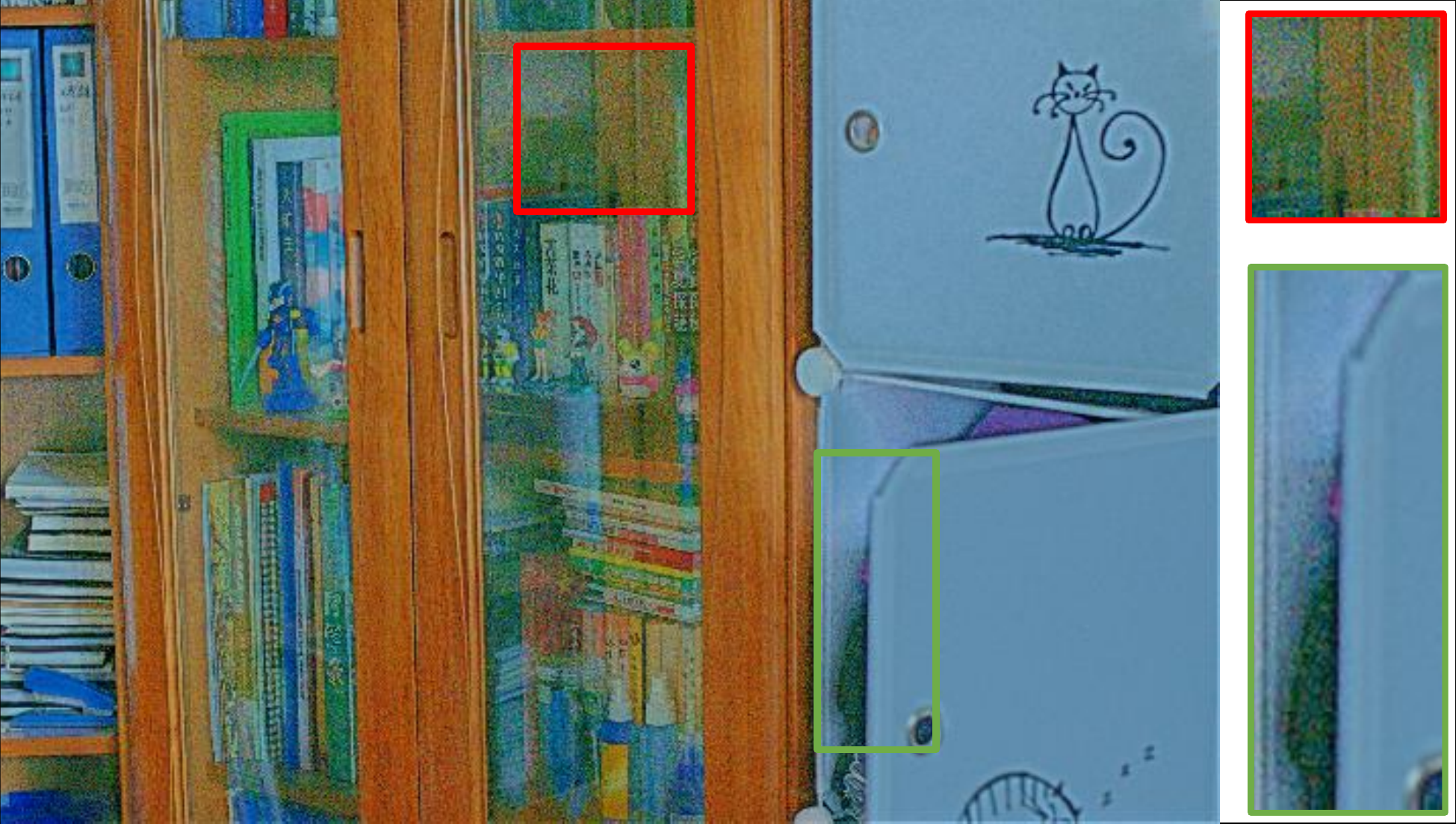}
}
\hspace{.15in}
\subfigure[ISSR]{
\includegraphics[height=4cm,width=7cm]{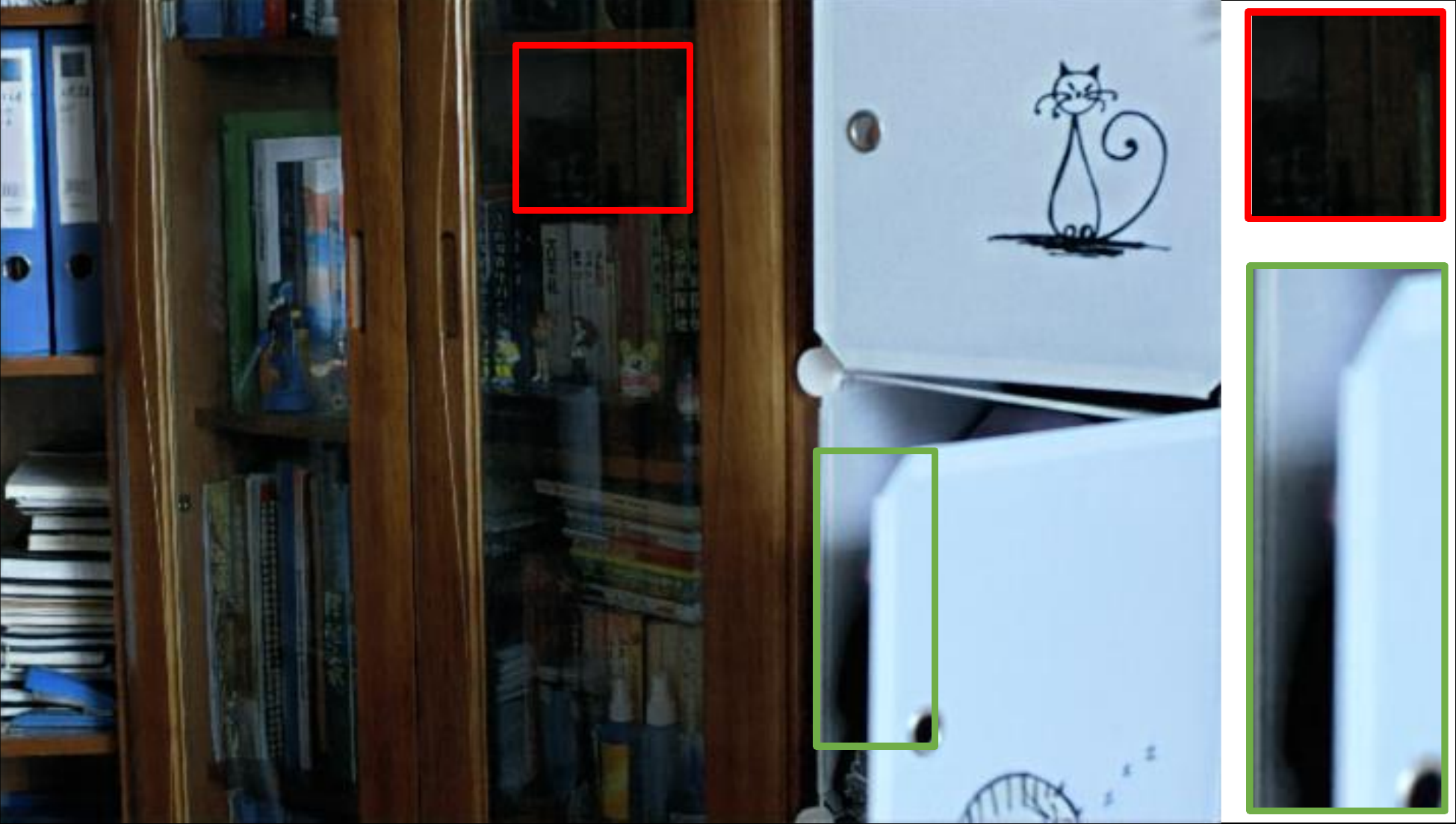}
}
\hspace{.15in}
\subfigure[Zero-DCE]{
\includegraphics[height=4cm,width=7cm]{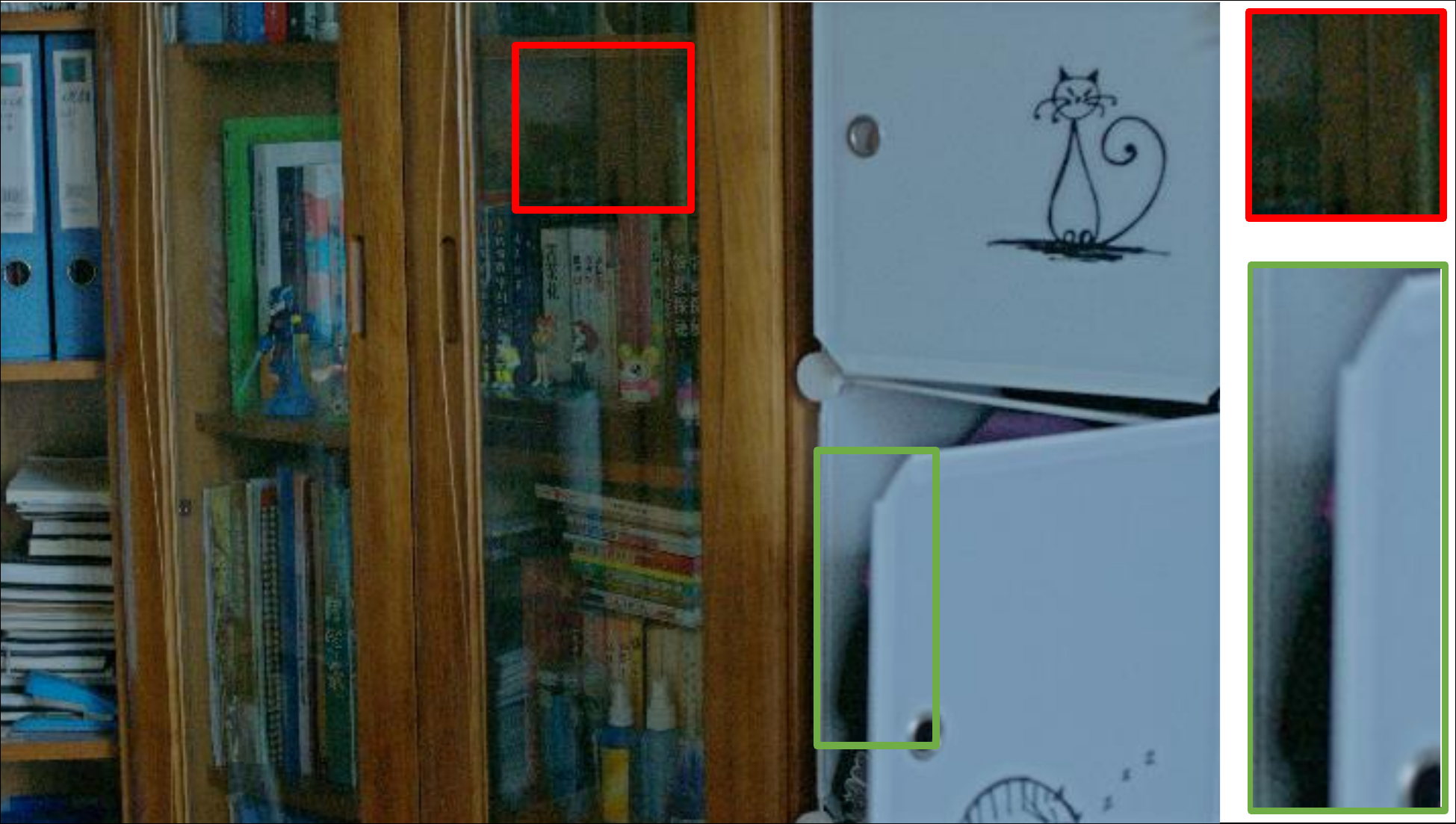}
}
\hspace{.15in}
\subfigure[EnlightenGAN]{
\includegraphics[height=4cm,width=7cm]{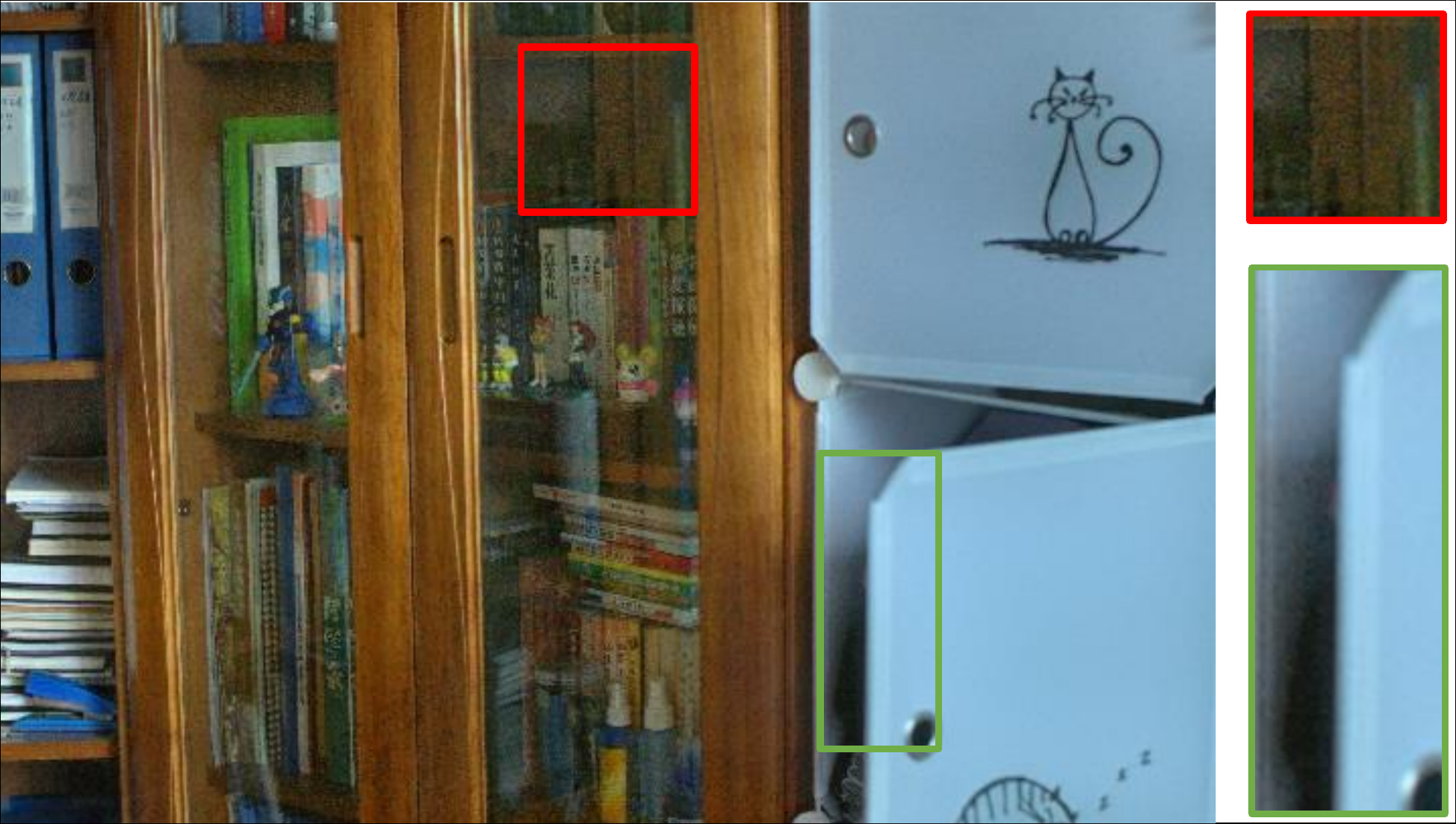}
}
\hspace{.15in}
\subfigure[RUAS]{
\includegraphics[height=4cm,width=7cm]{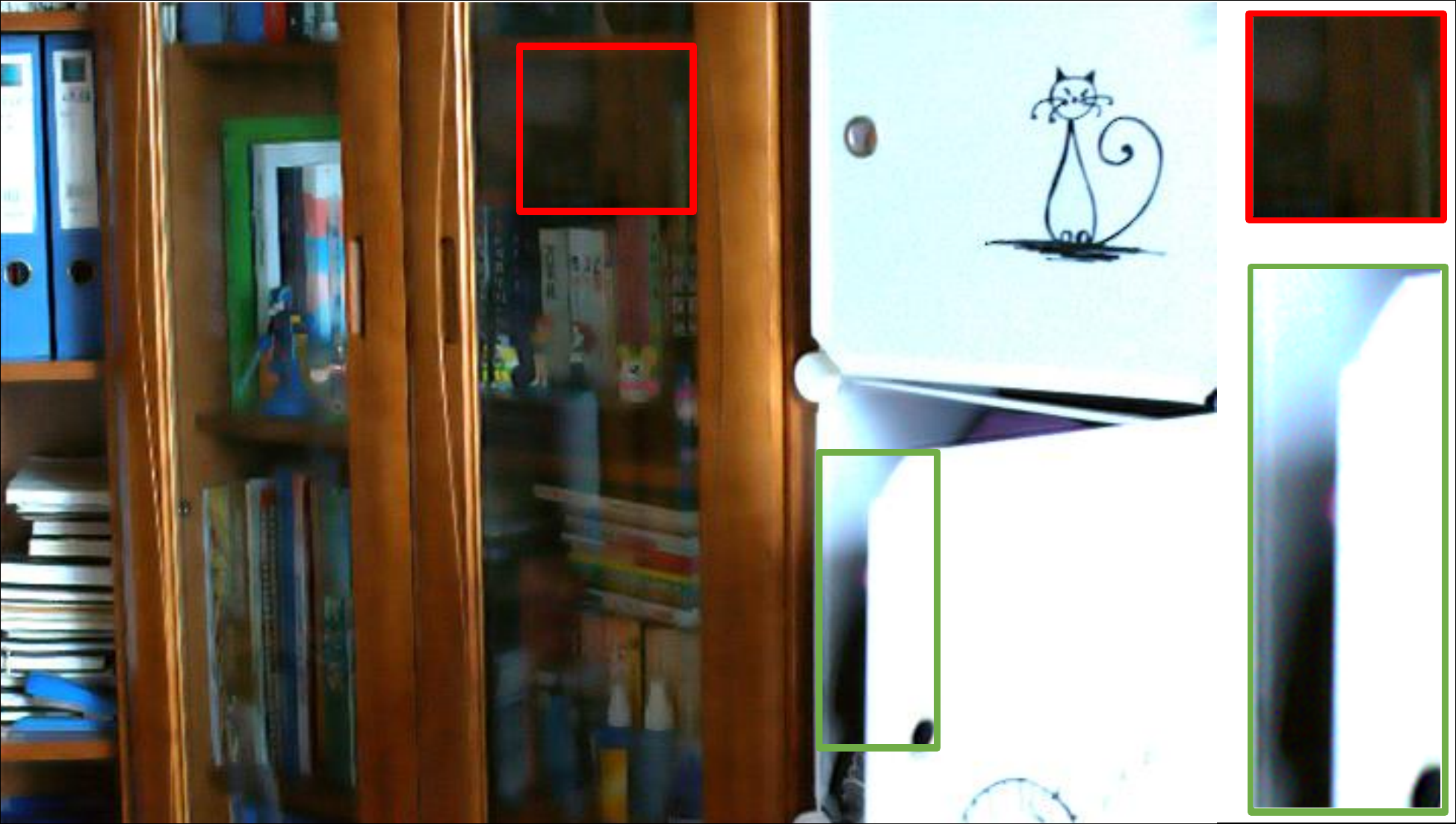}
}
\hspace{.15in}
\subfigure[ReLLIE]{
\includegraphics[height=4cm,width=7cm]{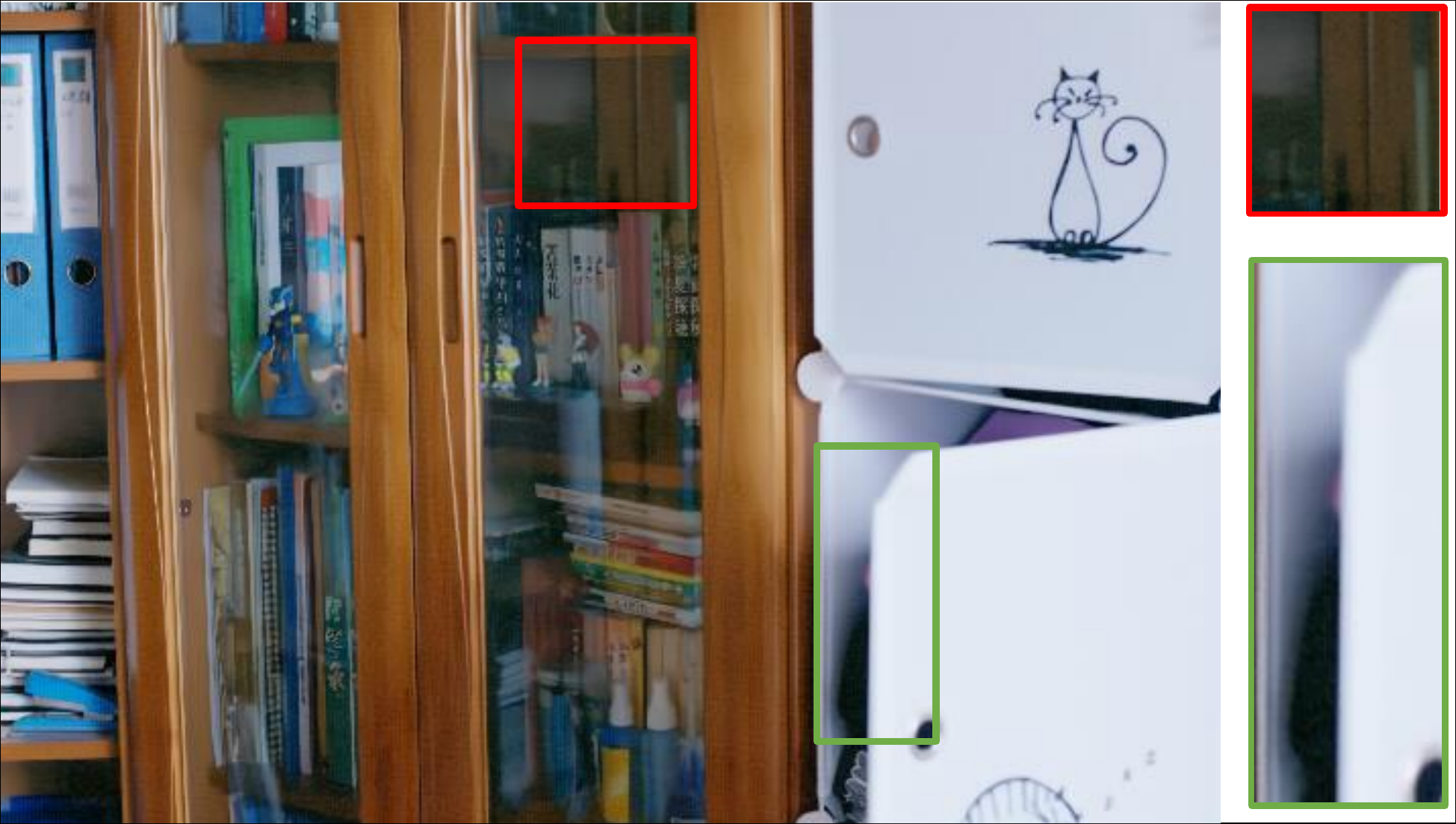}
}
\hspace{.15in}
\subfigure[SCL-LLE]{
\includegraphics[height=4cm,width=7cm]{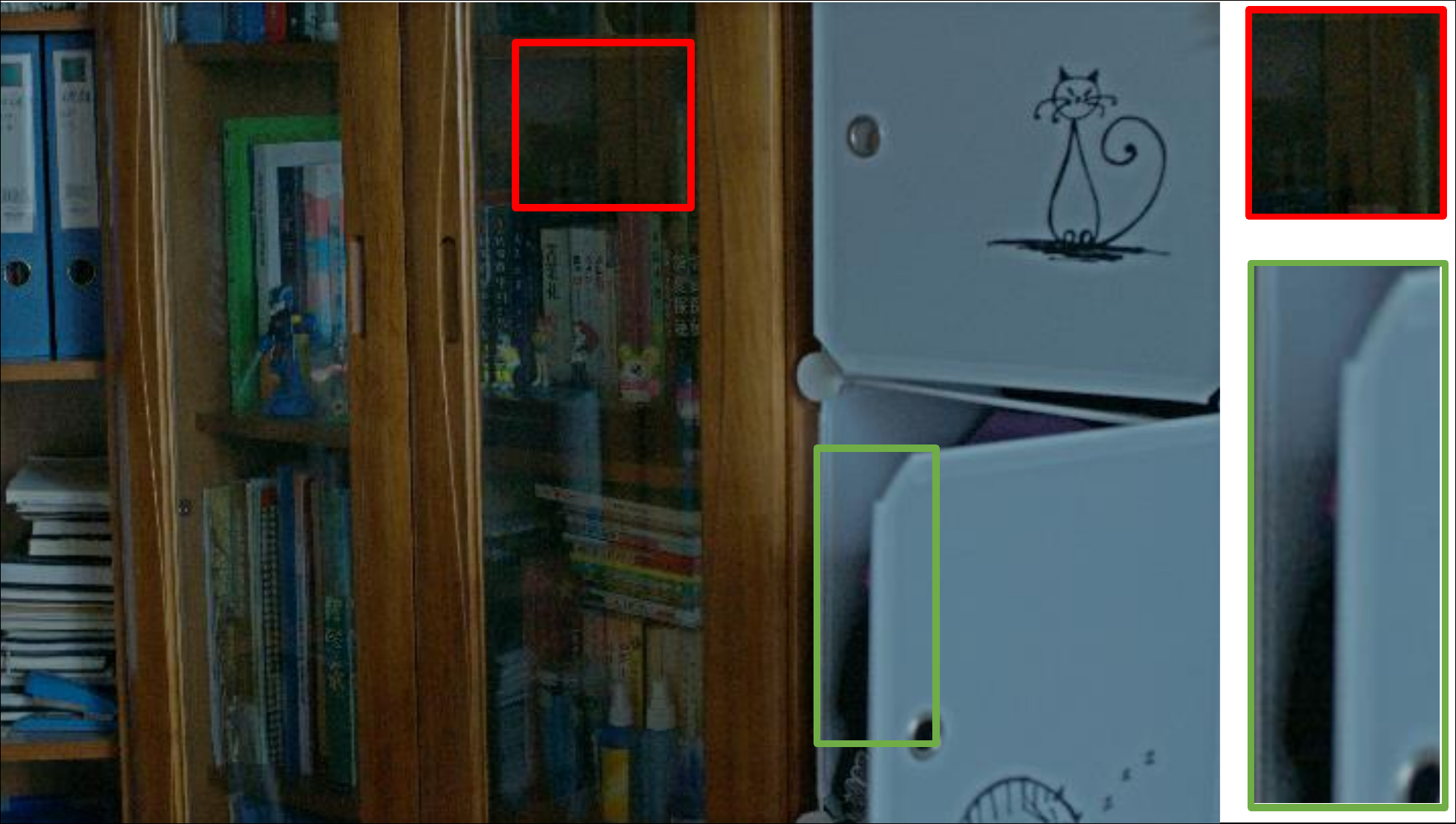}
}
\hspace{.15in}
\subfigure[Ours]{
\includegraphics[height=4cm,width=7cm]{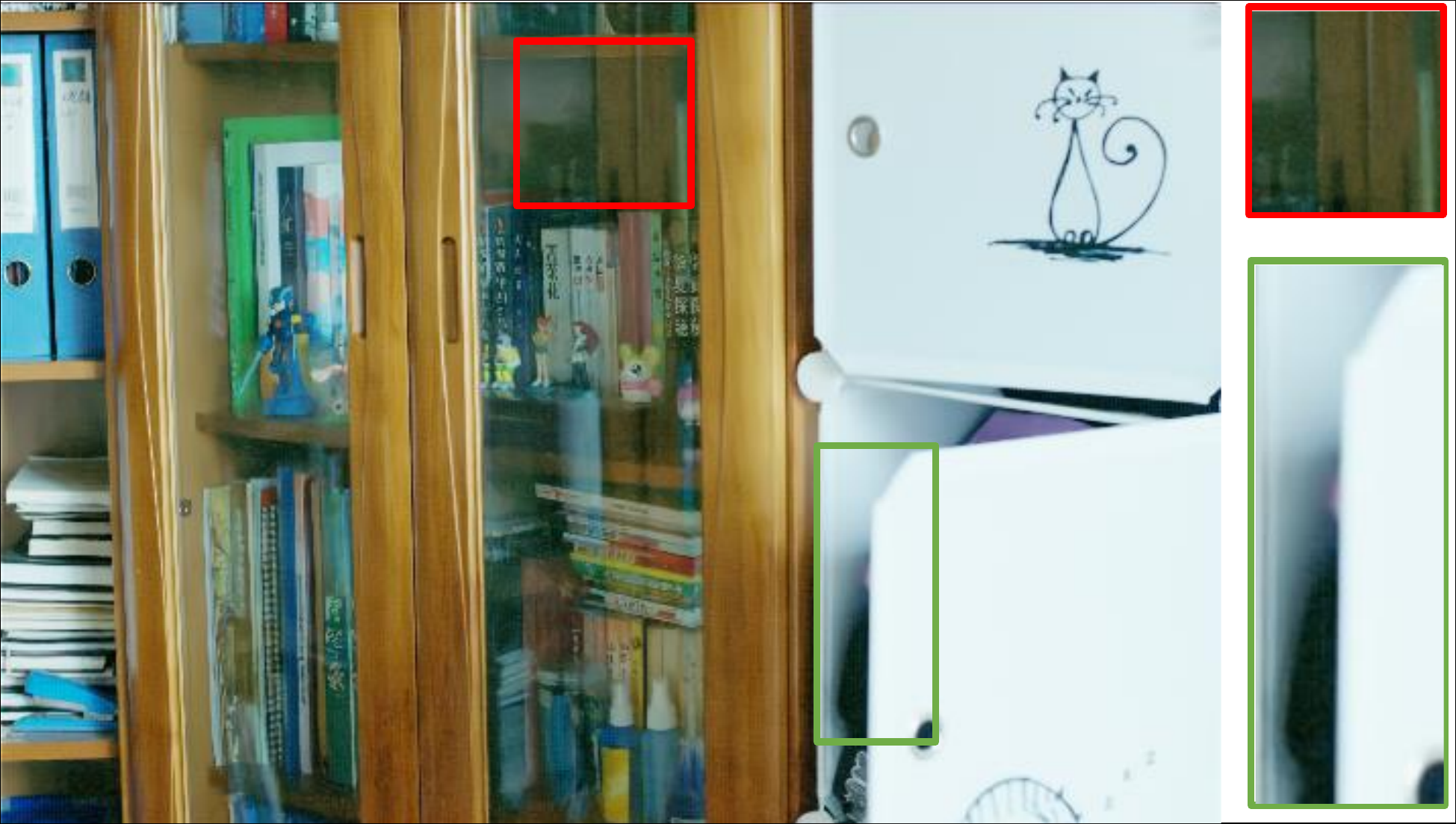}
}
\caption{Compared to the state-of-the-art methods, our method does not over-enhance low-light areas, exhibiting sharper details and more natural colors.}
\label{Vq3}
\end{figure*}

In this section, we present more visual comparison results against eight state-of-the-art methods
on LOL dataset as shown in Fig.~\ref{Vq1}-\ref{Vq3}. The eight methods are LIME \cite{guo2016lime}, Retinex-Net \cite{Chen2018Retinex}, ISSR \cite{FanWY020}, Zero-DCE \cite{Guo_2020_CVPR}, EnlightenGAN \cite{jiang2021enlightengan}, RUAS \cite{liu2021retinex}, ReLLIE \cite{zhang2021rellie}, and SCL-LLE \cite{liang2022semantically}. 
Except that LIME \cite{guo2016lime} is a non-deep learning method, other methods are based on deep learning. Our ALL-E produces no noticeable artifacts and is more visually pleasing.

\clearpage

\clearpage

\end{document}